\documentclass[lettersize,journal]{IEEEtran}
\usepackage{amsmath,amsfonts}
\usepackage{algorithm}
\usepackage{array}
\usepackage[caption=false,font=normalsize,labelfont=sf,textfont=sf]{subfig}
\usepackage{textcomp}
\usepackage{stfloats}
\usepackage{url}
\usepackage{verbatim}
\usepackage{graphicx}
\usepackage{cite}
\usepackage{algpseudocode}
\usepackage[accsupp]{axessibility} % Improves PDF readability for those with disabilities.
\usepackage{multirow}
\usepackage[normalem]{ulem}
\usepackage{amssymb}
\usepackage{hhline}
\usepackage[caption=false,font=normalsize,labelfont=sf,textfont=sf]{subfig}

\usepackage{graphicx}
\usepackage{colortbl}
\usepackage{multirow}
\usepackage{arydshln}

\usepackage{hyperref}
\hypersetup{
  colorlinks=true,
  linkcolor=blue, 
  citecolor=blue, 
  urlcolor=blue,    
  pdfborder={0 0 0}, 
}
\usepackage[capitalise]{cleveref}
\Crefname{figure}{Fig.}{Figs.}
\crefname{figure}{fig.}{figs.}
\Crefname{equation}{Eq.}{Eqs.}
\crefname{equation}{Eq.}{Eqs.}

\usepackage{xspace}
\newcommand{\ie}{i.e.\@\xspace}
\newcommand{\eg}{e.g.\@\xspace}
\newcommand{\wrt}{w.r.t.\@\xspace}

\usepackage{fontawesome5}
\definecolor{MScolor}{RGB}{158,131,253}
\definecolor{MIMcolor}{RGB}{255,168,60}
\definecolor{CLcolor}{RGB}{115,174,66}
\definecolor{RMcolor}{RGB}{0,32,96}
\definecolor{CRFR-Pcolor}{RGB}{255,0,0}

\begin{document}

% \title{Scalable Face Security Vision Foundation Models for Generalizable Deepfake Detection, Diffusion Facial Forensic, and Face Anti-Spoofing}
% \title{Scalable Face Security Vision Foundation Models for Deepfake Detection and Face Anti-Spoofing}
% \title{Face Security Vision Foundation Models against Deepfake, Diffusion, and Spoofing Faces}
\title{Scalable Face Security Vision Foundation Model for Deepfake, Diffusion, and Spoofing Detection}


\author{Gaojian Wang,
Feng Lin*,~\IEEEmembership{Senior Member,~IEEE,}
Tong Wu,
Zhisheng Yan,~\IEEEmembership{Member,~IEEE,}
Kui Ren,~\IEEEmembership{Fellow,~IEEE}

\thanks{* Corresponding author.}
\thanks{Gaojian Wang, Feng Lin, Tong Wu, and Kui Ren are with the State Key Laboratory of Blockchain and Data Security, Zhejiang University, and also with the Hangzhou High-Tech Zone (Binjiang) Institute of Blockchain and Data Security, Hangzhou 310000, China. (e-mail: \{wolo, flin, cocotwu, kuiren\}@zju.edu.cn); Zhisheng Yan is with the George Mason University, Fairfax, VA 22030 USA (e-mail: zyan4@gmu.edu)}
}

\markboth{Journal of \LaTeX\ Class Files,~Vol.~, No.~, September~2025}%
{Shell \MakeLowercase{\textit{et al.}}: Scalable Face Security Vision Foundation Models for Deepfake Detection and Face Anti-Spoofing}

\IEEEpubid{0000--0000/00\$00.00~\copyright~2025 IEEE}

\maketitle

\begin{abstract}
With abundant, unlabeled real faces, how can we learn robust and transferable facial representations to boost generalization across various face security tasks? We make the first attempt and propose \textbf{FS-VFM}, a scalable self-supervised pre-training framework, to learn fundamental representations of real face images. We introduce three learning objectives, namely \textbf{3C}, that synergize masked image modeling (MIM) and instance discrimination (ID), empowering FS-VFM to encode both local patterns and global semantics of real faces. Specifically, we formulate various facial masking strategies for MIM and devise a simple yet effective CRFR-P masking, which explicitly prompts the model to pursue meaningful intra-region \textbf{C}onsistency and challenging inter-region \textbf{C}oherency. We present a reliable self-distillation mechanism that seamlessly couples MIM with ID to establish underlying local-to-global \textbf{C}orrespondence. After pre-training, vanilla vision transformers (ViTs) serve as universal \textbf{V}ision \textbf{F}oundation \textbf{M}odels for downstream \textbf{F}ace \textbf{S}ecurity tasks: cross-dataset deepfake detection, cross-domain face anti-spoofing, and unseen diffusion facial forensics. To efficiently transfer the pre-trained FS-VFM, we further propose \textbf{FS-Adapter}, a lightweight plug-and-play bottleneck atop the frozen backbone with a novel real-anchor contrastive objective. Extensive experiments on 11 public benchmarks demonstrate that our FS-VFM consistently generalizes better than diverse VFMs, spanning natural and facial domains, fully, weakly, and self-supervised paradigms, small, base, and large ViT scales, and even outperforms SOTA task-specific methods, while FS-Adapter offers an excellent efficiency-performance trade-off. The code and models are available on \url{https://fsfm-3c.github.io/fsvfm.html}.
% Extensive experiments on 11 public benchmarks demonstrate that our FS-VFM consistently generalizes better than diverse VFMs span pre-training domains (natural/facial), paradigms (fully/weakly/self-supervised), and scales (small/base/large ViT), and even outperforms SOTA task-specific methods, while FS-Adapter offers an excellent efficiency–performance trade-off. The Code and models are available on \url{https://fsfm-3c.github.io/fsvfm.html}.
\end{abstract}


\section{Introduction}

\IEEEPARstart{F}{aces} sit at the nexus of daily interactions and information systems. This dual role makes the face security landscape suffer from escalating digital forgery and physical presentation attacks. Face forgery alters digital content while preserving a realistic appearance. With advanced generative models~\cite{kingma2013auto, goodfellow2014generative, ho2020denoising}, the evolving technologies, a.k.a., deepfakes, have sparked severe trust crises. Presentation attacks employ physical materials, \eg, printed photos, video replays, or 3D masks, to impersonate live faces and spoof face recognition, compromising real-life applications like face unlock and payment~\cite{yu2022deep}. Thus, both academia and industry strive to secure facial authenticity against forgeries and presentation attacks, via dedicated tasks: Deepfake Detection (\texttt{DFD}), Face Anti-Spoofing (\texttt{FAS}), and the emerging Diffusion Facial Forgery Detection (\texttt{DiFF}). Despite progress, most methods still struggle with novel or training-unseen manipulations, \textbf{raising generalizability as the common and primary challenge.}
% Human faces sit at the nexus of daily lives and biometric systems, underpin social interactions, and also gate high-stakes applications
% Despite notable progress, existing methods often struggle with generalizing to novel or training-unseen manipulations.
% Despite notable progress, generalization to novel or training-unseen manipulations remains the common and primary challenge

{
\setlength{\abovecaptionskip}{0pt}
\begin{figure}[t!]
\centering
\includegraphics[width=\linewidth]{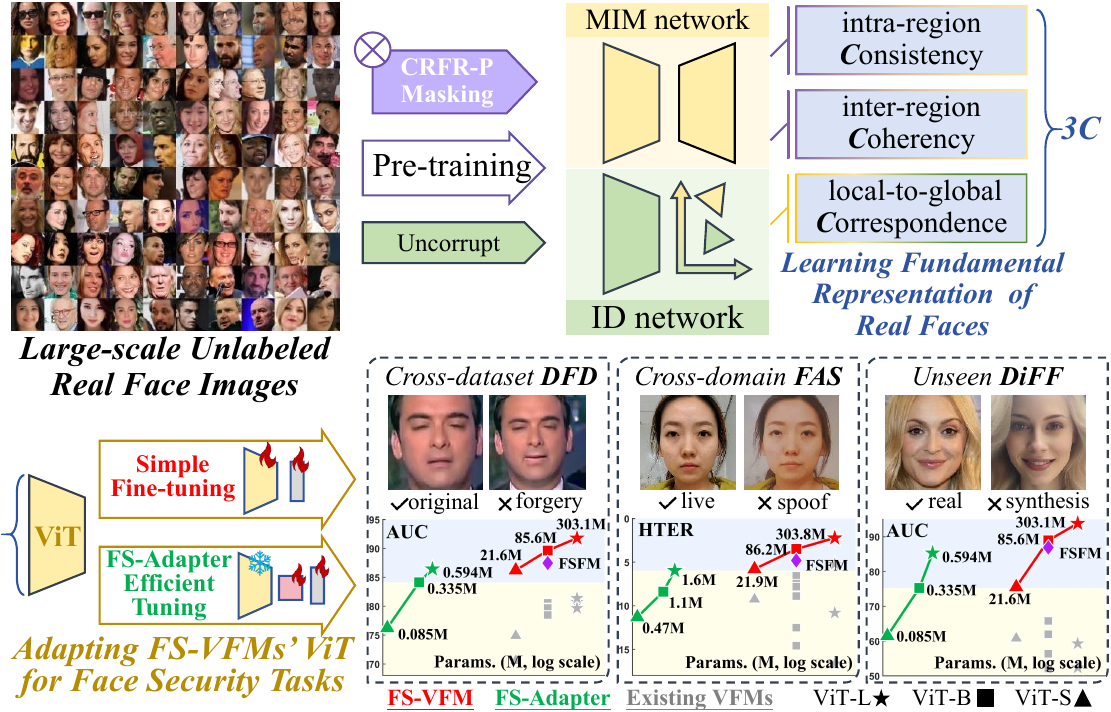}
% \vspace{-10pt}
\caption{\label{fig:teaser} A transferable, generalizable, and scalable Face Security Vision Foundation Model (FS-VFM). Simple fine-tuning of the vanilla ViT pre-trained from FS-VFM sets a new generalization bar across various downstream face security tasks, while the FS-Adapter enables ultra-efficient tuning. Results in the line sub-chart are average metrics from~\Cref{tab:dfd_vfm}, ~\Cref{tab:fas_vfm}, and ~\Cref{tab:DiFF}.}
\end{figure}
}

% Current DfD and FAS methods aim to improve generalization to unseen datasets within their respective tasks. DfD methods mainly focus on digital forgery patterns, such as spatiotemporal inconsistency~\cite{han2025towards, nguyen2025vulnerability, bai2023aunet}, generative augmentations~\cite{xia2024advancing, nguyen2024laa, bai2023aunet, shiohara2022detecting}, and region artifacts~\cite{nguyen2024laa, hong2024contrastive, xia2024advancing}. FAS methods~\cite{hu2024rethinking, zhou2024test, le2024gradient, zhou2023instance, sun2023rethinking} incorporate domain generalization techniques to discern physical spoof cues, \eg, paper textures and screen moiré patterns. Given these incompatible digital and physical features, most studies treat DfD and FAS as separate face security tasks, \ie, \textbf{task-specific} and \textbf{hard-to-transfer}. Besides, until now, most DfD and FAS methods still follow fully supervised learning with diverse backbones~\cite{he2016deep, chollet2017xception, tan2019efficientnet, dosovitskiy2020image} that are initialized from scratch or ImageNet (natural images)~\cite{deng2009imagenet} supervised pre-training. However, full supervision requires large-scale annotations or generative augmentations, which incur expensive costs and \textbf{limit scalability}. Moreover, initial weights that lack facial representations may \textbf{impede} \textbf{capability} and \textbf{generality} for face-related tasks~\cite{bulat2022pre, zheng2022general}.

Accordingly, current works on face security aim to improve cross-domain generalization within each task. \texttt{DFD} methods focus on generation or manipulation artifacts, \eg, spatial-temporal inconsistency~\cite{han2025towards, nguyen2025vulnerability, bai2023aunet}, blending traces~\cite{sun2025towards, yan2025generalizing, cheng2024can}, region anomalies~\cite{nguyen2024laa, hong2024contrastive, xia2024advancing}, whereas \texttt{FAS} methods employ domain adaptation~\cite{liu2024source, li2025optimal} or generalization~\cite{cai2025rehearsal, zhou2024test, hu2024rethinking, le2024gradient} techniques to capture presentation and attack clues like material textures and screen moiré patterns. Given these distinct signatures of digital forgeries versus physical spoofs, most studies tackle \texttt{DFD} and \texttt{FAS} independently, with separate models and training regimes—\textbf{remaining task-specific and lacking a universal representation for various face security tasks}.

\IEEEpubidadjcol
Further, the backbones driving face security frameworks are typically generic vision foundation models (VFMs) pre-trained on natural data, with preferred networks varies in \texttt{DFD}~\cite{chollet2017xception, tan2019efficientnet, dosovitskiy2020image} and \texttt{FAS}~\cite{he2016deep, shao2019multi, dosovitskiy2020image} tasks, as marked in~\Cref{tab:dfd_sota} and~\Cref{tab:fas_sota}. ImageNet fully supervised pre-training remains the de facto initialization standard, and recent works shift to weakly supervised vision-language models like CLIP~\cite{radford2021learning}. However, these generic VFMs lack facial domain focus, leaving a representation gap that impedes capability and generality in face-related tasks~\cite{bulat2022pre, zheng2022general}. Moreover, fully supervised learning requires extensive human annotations or data generation; vision–language pre-training demands web-scale image-text pairs plus heavy textual computation, where web-crawled captions are noisy, seldom aligned to facial content, and rarely describe the fine-grained cues critical to face security. These learning paradigms incur substantial costs or limit scalability, \textbf{posing challenges to pre-training a face security VFM}.

% In contrast to supervised learning, self-supervised learning (SSL) takes pretext tasks for pre-training on unlabeled data, where masked image modeling (MIM)~\cite{he2022masked, xie2022simmim, bao2021beit} and instance discrimination (ID, including contrastive learning~\cite{chen2020simple, he2020momentum, chen2021empirical} and distillation-based~\cite{grill2020bootstrap, chen2021exploring, caron2021emerging}) have proven superior performance in various vision tasks. As recent studies~\cite{DBLP:conf/iclr/ParkKH0Y23, zhu2023understanding, ozbulak2023know} suggest that MIM and ID are complementary, SOTA SSL methods~\cite{assran2023self, li2023mage, zhao2024asymmetric, chen2024context, wei2024towards, zhou2021ibot, DBLP:conf/iclr/YiGLYLWSQ23, huang2023contrastive, tao2023siamese, eymael2024efficient, jamal2025multi} combine them to enhance visual representations for natural images. However, SSL progress for facial pre-training, especially security tasks, remains limited, raising \textbf{Q1: how can face security tasks benefit from self-supervised pre-training to learn scalable and generic representations?}

In contrast to fully and weakly supervised paradigms, self-supervised learning (SSL) eliminates annotations or other metadata paired with images, and unlocks scalable pre-training on unlabeled data via pretext tasks, notably masked image modeling (MIM)~\cite{bao2021beit, wei2022masked, xie2022simmim, he2022masked}, which masks parts of an image then reconstructs the masked content, and instance discrimination (ID), which distinguishes each instance from others (including contrastive learning~\cite{chen2020simple, he2020momentum, chen2021empirical} and distillation~\cite{grill2020bootstrap, chen2021exploring, caron2021emerging}), have delivered superior downstream performance. As multiple studies~\cite{DBLP:conf/iclr/ParkKH0Y23, zhu2023understanding, ozbulak2023know} suggest that MIM and ID complement each other, recent SSL methods~\cite{assran2023self, li2023mage, zhao2024asymmetric, chen2024context, wei2024towards, zhou2021ibot, DBLP:conf/iclr/YiGLYLWSQ23, huang2023contrastive, tao2023siamese, eymael2024efficient, jamal2025multi} integrate them within joint embedding architectures (JEA), to improve representation quality for general vision tasks. Yet, the potential of these SSL advances for facial representation learning, particularly security tasks, remains untapped, motivating \textbf{Q1: how can face security tasks benefit from self-supervised pre-training to learn universal and scalable representations?}

While existing works explore SSL for face security, most remain tied to specific forgery or spoof patterns and fall short of transferable representations. Some \texttt{DFD} methods~\cite{shiohara2022detecting, yan2025generalizing, cheng2024can} synthesize pseudo-fakes from real faces to simulate artifacts like blending, vulnerable to unknown manipulations or spoofing. Others~\cite{feng2023self, haliassos2022leveraging} rely on paired multimodal data, \eg, audio-video, limiting scalability. Recent efforts introduce JEA-based~\cite{zhang2024learning} and MIM-based~\cite{nguyen2025vulnerability} SSL to learn the temporal consistency of real videos, but fail on image forgeries and real video replays. In \texttt{FAS}, SSL has been used to exploit spoofing cues via domain positives~\cite{liu2023towards}, domain-invariant semantics~\cite{zheng2024mfae}, or domain alignment~\cite{liu2022source}, yet these \texttt{FAS} domain knowledge contribute little to digital forgeries detection.

Meanwhile, recent progress in facial pre-training~\cite{bulat2022pre, zheng2022general, cai2023marlin, wang2023toward, liu2023pose, gao2024self} seeks task-agnostic facial representations that transfer across diverse face analysis tasks, \eg, attribute recognition and AU detection. However, these methods focus primarily on salient appearances that deepfake, diffusion, or spoofing faces can also mimic well, rather than modeling facial ``realness'' representations \wrt authenticity, and thus struggle to extrapolate to face security tasks. Moreover, these facial VFMs are typically optimized for downstream intra-dataset evaluations, whereas face security demands cross-dataset generalization. These issues raise: \textbf{Q2: How can we learn fundamental representations of real faces that transfer well to diverse face security tasks and improve downstream generalization?} 

To bridge the above gaps, we propose to learn the intrinsic properties of unlabeled real face images, and present \textbf{FS-VFM}, a scalable self-supervised pre-training framework that contributes universal, transferable, and generalizable \textbf{V}ision \textbf{F}oundation \textbf{M}odels for various \textbf{F}ace \textbf{S}ecurity tasks. As shown in~\Cref{fig:teaser}, FS-VFM synergizes masked image modeling (MIM) and instance discrimination (ID) within a joint architecture to pursue three pre-training objectives. Specifically, we introduce a novel CRFR-P facial masking strategy, Covering a Random Facial Region (\eg, nose, eyes) and Proportionally masking other regions, into a masked autoencoder~\cite{he2022masked}, which not only yields a meaningful and challenging facial MIM task but also focuses the model’s attention on \textit{inter-region \textbf{C}oherency} and \textit{intra-region \textbf{C}onsistency}. For reliable facial semantics alignment, we formulated an ID network coupled with MIM via elaborate self-distillation: the CRFR-P masked online view induces spatial variances, the uncorrupted target view retains complete semantics, Siamese representation decoders build a disentangled space, and no data augmentation preserves intact information, linking \textit{local-to-global \textbf{C}orrespondence}. Together, these \textit{\textbf{3C}} objectives enrich facial representations with pixel-level context perceptiveness, region-level relation awareness, and instance-level face invariance. Thus, FS-VFM empowers both local and global facial perception, learns fundamental representations of real faces, transfers well to diverse face security tasks, and boosts downstream generalization.

We adopt vanilla ViTs~\cite{dosovitskiy2020image} as the FS-VFM encoder, providing a universal backbone that scales across model sizes. Simple fine-tuning of FS-VFM even outperforms many task-specific SOTA methods, and scaling up the model consistently improves downstream generalization. However, larger models accentuate the cost of per-task adaptation. In fact, given mismatched pre-trained domains and disparate backbones, existing face security methods necessitate either full fine-tuning~\cite{cheng2024can, sun2025towards, li2025optimal} or bespoke efficient tuning~\cite{fu2025exploring, han2025towards, hu2024fine} with nontrivial designs, undermining cross-task modularity and reusability. This motivates \textbf{Q3: How can we efficiently adapt the off-the-shelf facial representations from FS-VFM to various face security tasks?} As a promising solution, the adapter~\cite{houlsby2019parameter} appends and updates lightweight modules across layers of the fixed ViT backbone, but tuning vanilla adapters, \ie, multiple linear layers, often overfits to specific manipulations, overlooks generalizable patterns, and still backpropagates through the backbone. Hence, we propose \textbf{FS-Adapter}, a plug-and-play bottleneck attached only atop the frozen encoder. To harness our strong facial representations, we sustain our pre-training philosophy \ie, modeling realness, and introduce RACL for the FS-Adapter, which takes only Real faces as Anchors for Contrastive Learning in a compact bottleneck space. This not only retains most generalizability but also further reduces trainable parameters. As a result in~\Cref{fig:teaser}, built upon FS-VFM ViT-L/16 ($\sim$$303$M), our FS-Adapter ($\sim$$0.59$M) only occupies $<$$0.2\%$ backbone parameters and $<$$4.7\%$ of vanilla adapters, yet even generalizes better than fully fine-tuning other VFMs—enabling ultra-efficient adaptation to downstream face security tasks.

This paper is a substantial extension of our prior CVPR 2025 work~\cite{wang2025fsfm} on FSFM. In this version, we further enrich our framework as a full-stack, versatile solution that spans pre-training, fine-tuning, and adaptation stages, delivering not only a transferable and generalizable but also a scalable and deployable face security vision foundation model, as follows: 1) From a single FSFM ViT-B/16 to FS-VFM ViT-\{S/16, B/16, L/16\} families, we explore and scale the model capacity, recast pre-training recipe, and demonstrate consistent scalability \wrt generality across downstream face security tasks, see~\Cref{fig:teaser}. 2) We introduce a lightweight plug-and-play FS-Adapter with a novel real-anchor contrastive objective, ‌which efficiently transfers pre-trained FS-VFMs to downstream tasks readily. With a frozen FS-VFM ViT-L/16, FS-Adapter updates only a small bottleneck ($<$$0.2\%$ parameters), yet generalizes better than fully fine-tuning other VFMs. 3) We go all out to broaden evaluations: we benchmark FS-VFMs against a wider spectrum of VFMs covering pre-training domains (facial and natural) and paradigms (full, self, and vision-language supervised), plus different backbone sizes, across 11 face security benchmarks (adding Celeb-DF++~\cite{li2025celeb}), to thoroughly position our advantages, and we also update recent task-specialized methods in comparisons. 4) New results show that simple fine-tuning of our FS-VFM sets a new generalization groundwork for \texttt{DFD}, \texttt{FAS}, and \texttt{DiFF} tasks, while FS-Adaper offers a compelling efficiency-performance trade-off. 5) We provide more in-depth analysis of pre-training and scaling FS-VFM, and qualitative visualizations, to shed light on our framework. 

The main contributions of this paper are:

% To sum up, the main contributions of this paper are listed:
% The main contributions of this work are as follows:
% The contributions of this paper are summarized as below.
% Main contributions of our paper:
% The contribution of this work is summarized below:
% To sum up, the contributions of this paper are listed:

$\bullet$ We propose FS-VFM, a scalable self-supervised pre-training framework, which synergizes facial masked image modeling and instance discrimination for both local context perception and global semantic alignment, to pursue fundamental and transferable representations of real faces, serving as the first unified face security vision foundation model.

$\bullet$ We formulate \textit{\textbf{3C}} learning objectives, introduce a simple yet effective CRFR-P facial masking that directs MIM to prompt meaningful intra-region \textit{\textbf{C}onsistency} and reinforce challenging inter-region \textit{\textbf{C}oherency}, and elaborate a reliable joint self-distillation that couples MIM with ID to establish underlying local-to-global \textit{\textbf{C}orrespondence}.

% $\bullet$ We introduce FS-Adapter, a lightweight bottleneck appended only atop the frozen encoder, with a novel real-anchor contrastive learning. This plug-and-play module flexibly transfers our strong facial representations to various downstream face security tasks with minimal computational overhead, while retaining superior generalization.
$\bullet$ We introduce the FS-Adapter, a lightweight bottleneck atop the frozen encoder, featuring novel real-anchor contrastive learning. This plug-and-play module flexibly transfers our facial representations to various downstream face security tasks with minimal overhead, while retaining strong generalization.
% $\bullet$ We introduce FS-Adapter, a lightweight plug-and-play bottleneck atop the frozen encoder, with a novel real-anchor contrastive learning, which enables flexible transfer of our facial representations to various downstream face security tasks with minimal overhead and strong generalization.

$\bullet$ We conduct extensive experiments across 11 benchmarks on prevalent face security tasks: cross-dataset deepfake detection (\texttt{DFD}), cross-domain face anti-spoofing (\texttt{FAS}), and unseen diffusion facial forgery detection (\texttt{DiFF}), which demonstrate our FS-VFMs consistently generalize better than diverse VFMs that span natural and facial domains, full, self, and vision-language supervised paradigms, across small, base, and large ViT sizes. Simple fine-tuning of FS-VFM even outperforms SOTA task-specific methods and establishes a new generalization baseline, while FS-Adapter achieves an excellent efficiency–performance solution.

\section{Related Work}
\subsection{Visual Representation Learning} 
% In recent years, visual representation learning has transitioned from ImageNet-supervised~\cite{deng2009imagenet} to self-supervised~\cite{caron2021emerging, he2022masked} and vision–language~\cite{radford2021learning} pre-training families, with network architectures shifted from convolutional neural networks (CNNs)~\cite{he2016deep, chollet2017xception, tan2019efficientnet} to vision transformers (ViTs)~\cite{dosovitskiy2020image}. Self-supervised learning (SSL) has gained prominence by eliminating costly annotations while achieving superior downstream performance. Two powerful pretext tasks, masked image modeling (MIM) and instance discrimination (ID), have dominated generative and discriminative SSL paradigms, respectively.

Recently, visual representation learning has shifted from ImageNet-supervised~\cite{deng2009imagenet} to self-supervised~\cite{caron2021emerging, he2022masked} and vision–language~\cite{radford2021learning} pre-training, with vision transformers (ViTs)~\cite{dosovitskiy2020image} over traditional CNNs~\cite{he2016deep, chollet2017xception, tan2019efficientnet}. Self-supervised learning (SSL) has gained prominence by eliminating costly annotations while achieving strong downstream performance. Two powerful pretext tasks, masked image modeling (MIM) and instance discrimination (ID), have dominated generative and discriminative SSL paradigms, respectively.

% \noindent\textbf{Masked‐image modeling (MIM)} formulates a reconstruction task that masks portions of an image and takes visible parts to recover the masked contents, such as abstract tokens~\cite{bao2021beit}, auxiliary features~\cite{wei2022masked}, or pixel values~\cite{xie2022simmim, he2022masked}. The pioneering work BEiT~\cite{bao2021beit} employs a visual tokenizer to map image patches into ``visual words'' and predicts the discrete token index. To simplify tokenization, MaskFeat~\cite{wei2022masked} instead utilizes a handcrafted (HOG) feature as an effective regression target. Further, the seminal MAE~\cite{he2022masked} introduces an asymmetric encoder-decoder to predict pixels directly, showing that random masking with a high ratio (75\%) enables efficient and scalable pre-training, while yielding high-quality representations. Beyond what to predict, the masking policy governs the reconstruction target. Subsequent studies~\cite{kakogeorgiou2022hide, shi2022adversarial, wang2023hard} explore various masking strategies to challenge spatial reasoning for more meaningful visual features. In general, naïve MIM focuses on encoding local information to predict the missing parts, but lacks a global discriminative constraint.

\noindent\textbf{Masked Image Modeling (MIM)} formulates a reconstruction task that masks portions of an image and takes visible parts to recover the masked contents, such as visual tokens in BEiT~\cite{bao2021beit}, auxiliary features in MaskFeat~\cite{wei2022masked}, or pixel values in SimMIM~\cite{xie2022simmim} and MAE~\cite{he2022masked}. The tokenizer-free MAE introduces an asymmetric encoder-decoder to restore pixels directly, showing that a high ratio (75\%) random masking enables efficient and scalable pre-training, while yielding high-quality representations. Beyond what to predict, the masking policy governs the reconstruction target. Subsequent studies~\cite{kakogeorgiou2022hide, shi2022adversarial, wang2023hard} explore various masking strategies to challenge visual reasoning for more meaningful features. In general, naïve MIM focuses on encoding local information to predict the missing parts, but lacks a global discriminative constraint.

% \noindent\textbf{Instance discrimination (ID)} composes a metric learning problem that distinguishes each image instance from others. This typically employs Siamese encoders to maximize agreement among positive pairs (augmented views) of the same image. To avoid collapsing solutions, contrastive learning~\cite{chen2020simple, he2020momentum, chen2021empirical} simultaneously minimizes similarity between negative pairs from different images, \eg, SimCLR~\cite{chen2020simple} and MoCo~\cite{he2020momentum} use large batch sizes or memory banks to provide sufficient negatives, with strong data augmentations. MoCov3~\cite{chen2021empirical} further builds the training recipes for the ViTs. To circumvent negative pairs, distillation methods~\cite{grill2020bootstrap, chen2021exploring, caron2021emerging} align latent representations through an asymmetrical teacher-student architecture. \eg, BYOL~\cite{grill2020bootstrap} uses a momentum updated teacher network and an additional predictor in the student branch to match the teacher’s embedding, SimSiam~\cite{chen2021exploring} shows that even without a momentum encoder, a stop-gradient on one branch suffices to prevent collapse, and DINO~\cite{caron2021emerging} adapts the ViT to distillation via the momentum encoder and multi-crop training. In summary, ID excels at learning global semantics for image invariance, yet overlooks fine-grained texture awareness.

\noindent\textbf{Instance Discrimination (ID)} comprises a metric learning problem that distinguishes each image instance from others. This typically employs Siamese encoders to pull positive pairs (augmented views of the same image) closer. To avoid collapsing solutions, contrastive learning approaches~\cite{chen2020simple, chen2021exploring, caron2021emerging} simultaneously push negative pairs (from different images) away, yet require sufficient negatives and strong data augmentations. To circumvent negative pairs, distillation methods like BYOL~\cite{grill2020bootstrap}, SimSiam~\cite{chen2021exploring}, and DINO~\cite{caron2021emerging}, align latent representations by asymmetric teacher-student architectures, \eg, a momentum updated encoder~\cite{grill2020bootstrap, caron2021emerging}, an additional predictor~\cite{grill2020bootstrap, chen2021exploring}, or a stop-gradient operation~\cite{chen2021exploring, caron2021emerging}. In summary, ID excels at learning global semantics for image invariance, yet overlooks fine-grained texture awareness.

\noindent\textbf{Joint Embedding Architectures (JEA)} Previous studies~\cite{DBLP:conf/iclr/ParkKH0Y23, zhu2023understanding, ozbulak2023know} have revealed that MIM and ID are complementary: MIM captures fine-grained details while ID aligns high-level semantics. Accordingly, recent SSL frameworks converge toward joint embedding architectures (JEA) that integrate MIM with ID via Siamese designs~\cite{tao2023siamese, assran2023self, eymael2024efficient}: inject contrastive learning into MIM to ensure global consistency when reconstructing spatial details~\cite{DBLP:conf/iclr/YiGLYLWSQ23, huang2023contrastive, li2023mage, wei2024towards, jamal2025multi}, or leverage distillation for robust teacher-student alignment\cite{zhou2021ibot, chen2024context, zhao2024asymmetric}. Overall, JEA-based SSL methods have delivered stronger representations for general vision tasks than using either alone. However, these JEA-based SSL progresses for facial representations, especially security tasks, remain limited.

\subsection{Facial Representation Learning}

SSL for facial representation poses distinct challenges versus generic vision, owing to the unique textures yet highly similar semantics. Several works~\cite{nguyen2023micron, sun2024lafs, tourani2024pose} have tailored facial SSL to mitigate overfitting and improve performance, but are task-specific. Notably, FaRL~\cite{zheng2022general} combines image–text contrastive learning with MIM to transfer across diverse facial analysis tasks, excluding face security. However, as a non-pure visual SSL, it requires extensive face–caption pairs (20M) and computation for the text encoder, where web-crawled captions often describe trivial context rather than facial details that security tasks demand. More recent efforts target SSL to learn task-agnostic representations via masked image modeling~\cite{cai2023marlin, wang2023toward}, contrastive learning~\cite{bulat2022pre,liu2023pose, wang2023toward}, and distillation~\cite{gao2024self, wang2023toward}, for various tasks: expression~\cite{bulat2022pre, cai2023marlin, liu2023pose, gao2024self} and attribute~\cite{cai2023marlin, gao2024self} recognition, AU detection~\cite{bulat2022pre, liu2023pose}, and face alignment~\cite{bulat2022pre, wang2023toward, gao2024self}, etc. For instance, MARLIN~\cite{cai2023marlin} introduces a facial tube masking for the video MAE to learn spatio-temporal features, and MCF~\cite{wang2023toward} formulates a JEA-based SSL framework that couples MIM, contrastive learning, and distillation to enhance facial semantic learning.

% Despite their success in conventional face analyses, these works struggle to extrapolate to face security tasks. In particular, existing methods primarily focus on calibrating salient facial appearances that forgery and spoofing faces may also exhibit, rather than explicitly modeling facial ``realness'' representations that convey authenticity cues. Moreover, these works optimize for intra-domain performance when transferring to a downstream evaluation, whereas face security tasks necessitate cross-domain generalization within the specific task. In this work, we seek to bridge this gap by learning a fundamental, transferable, and realness-aware facial representation, improving generalizability across diverse face security tasks that prior models cannot adequately adapt.

Despite advancing conventional face analyses, these works struggle with face security tasks. Existing methods calibrate salient facial features that forgery or spoofing also exhibit well, but overlook ``realness'' representations. Furthermore, prior works adopt intra-domain evaluation for downstream tasks, whereas face security tasks call for cross-domain generalization. These gaps motivate us to learn generalizable and realness-aware facial representations for face security tasks.

{
\setlength{\abovecaptionskip}{0pt}
\begin{figure*}[t!]
\centering
\includegraphics[width=.9\textwidth]{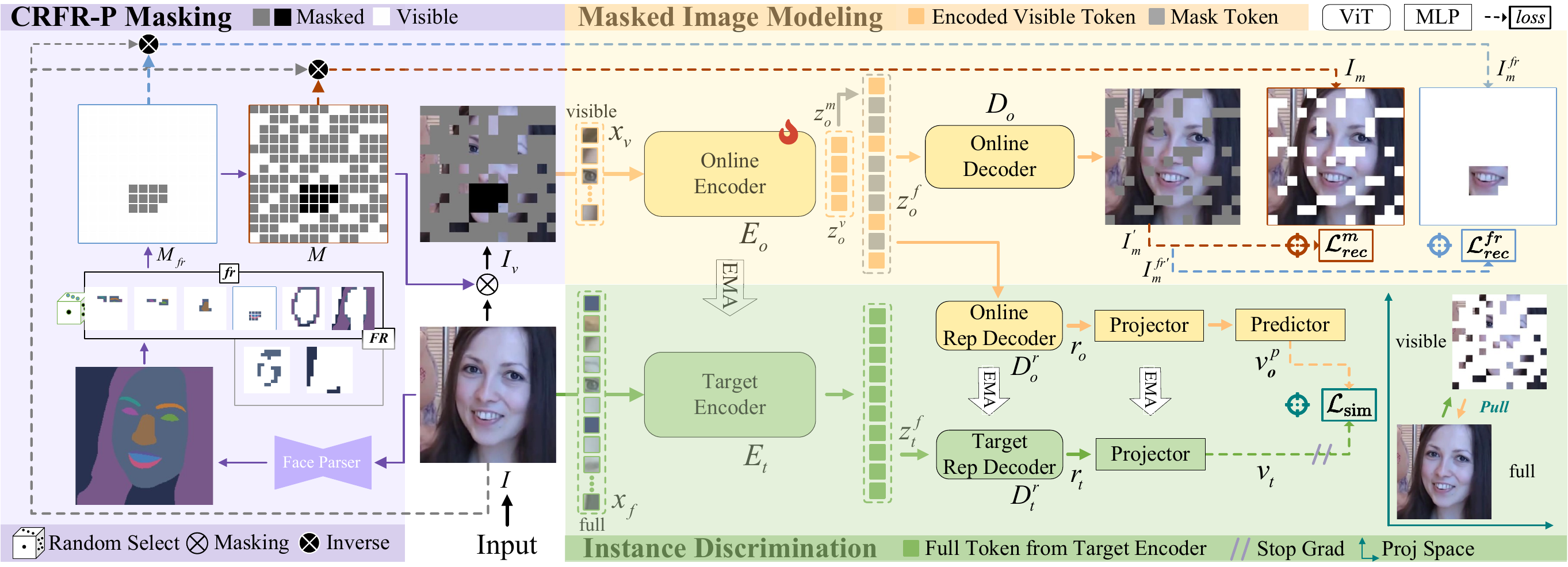}
% \vspace{-15pt}
\caption{\textbf{Overview of FS-VFM} self-supervised pre-training framework for learning foundational representations of real faces (\textit{\textbf{3C}}~\faCrosshairs). Guided by the \textcolor{MScolor}{\textbf{CRFR-P masking}} strategy, the \textcolor{MIMcolor}{\textbf{masked image modeling (MIM)}} network promotes \textit{intra-region \textbf{C}onsistency} with $\mathcal{L}_\mathit{rec}^\mathit{m}$ and enforces \textit{inter-region \textbf{C}oherency} via $\mathcal{L}_\mathit{rec}^\mathit{fr}$, while the \textcolor{CLcolor}{\textbf{instance discrimination (ID)}} network collaborates to foster \textit{local-to-global \textbf{C}orrespondence} through $\mathcal{L}_\mathit{sim}$. Given an input image $I$, the \textcolor{MScolor}{\textbf{CRFR-P masking}} generates a facial region mask $M_\mathit{fr}$ and an image mask $M$ sequentially. The \textcolor{MIMcolor}{\textbf{MIM network}}, a masked autoencoder, reconstructs the masked face $I_\mathit{m}$ from visible patches $x_\mathit{v}$ (masked by $M$), emphasizing the fully masked region $I_\mathit{m}^\mathit{fr}$ (specified by $M_\mathit{fr}$). The \textcolor{CLcolor}{\textbf{ID network}} maximizes the representation similarity between the masked online view $I_\mathit{v}$ and the full (unmasked) target view $I$ of the same sample by projection onto a disentangled space structured via Siamese representation decoders. After pre-training, the online encoder $E_\mathit{o}$, a vanilla ViT {\color{red} \faFire}, is applied to boost downstream face security tasks.}
\label{fig:framework}
\end{figure*}
}

\subsection{Downstream Face Security Tasks}\label{sec:rel_fst}
With numerous benchmarks for deepfake detection~\cite{rossler2019faceforensics++, li2020celeb, dolhansky2020deepfake, dolhansky2019dee, zi2020wilddeepfake, li2025celeb}, face anti-spoofing~\cite{zhang2012face, chingovska2012effectiveness, wen2015face, boulkenafet2017oulu}, and diffusion face forensic~\cite{cheng2024diffusion} tasks, deep learning models have achieved strong intra-dataset results but suffer from unseen forgeries or spoofs in real-world scenarios. Thus, SOTA methods for face security aim to improve generalization within their respective task.

% \subsubsection{Deepfake Detection (\texttt{DFD})} evolves to pursue cross-dataset generalizability by moving beyond dataset‑related artifacts. Along the “what to detect” axis, recent works mainly explore specific forgery patterns like spatial-temporal inconsistency~\cite{han2025towards, nguyen2025vulnerability}, frequency clues~\cite{gu2022exploiting, wang2023dynamic}, and identity‑information mismatches~\cite{nirkin2021deepfake, huang2023implicit, chen2024diffusionfake}, alongside specialized designs such as forgery feature disentanglement~\cite{zhu2023face, yan2023ucf, fu2025exploring, li2025critical} or multiple auxiliary objectives~\cite{shao2024detecting, nguyen2024laa, hong2024contrastive} for regularization. In addition, sophisticated generative data augmentation is prevalent, either by synthesizing pseudo‑fakes or simulating artifacts, at image~\cite{li2020face, shiohara2022detecting, bai2023aunet, xia2024advancing, nguyen2024laa, sun2025towards, yu2025unlocking}, video~\cite{yan2025generalizing, nguyen2025vulnerability}, or feature levels~\cite{yan2024transcending, cheng2024can}, to enrich forgery diversity and mitigate overfitting. 

\noindent\textbf{Deepfake Detection (\texttt{DFD})} evolves to pursue cross-dataset generality by moving beyond dataset‑related patterns. Recent works mainly explore specific forgery artifacts like spatial-temporal inconsistency~\cite{han2025towards, nguyen2025vulnerability, bai2023aunet}, frequency clues~\cite{gu2022exploiting, wang2023dynamic}, and identity mismatches~\cite{nirkin2021deepfake, huang2023implicit, chen2024diffusionfake}, alongside specialized regularizations, such as forgery feature disentanglement~\cite{zhu2023face, yan2023ucf, fu2025exploring, li2025critical} and multiple auxiliary objectives~\cite{shao2024detecting, nguyen2024laa, hong2024contrastive}. In addition, tailored data augmentations, which generate pseudo‑fakes or simulate artifacts at image~\cite{li2020face, shiohara2022detecting, bai2023aunet, xia2024advancing, nguyen2024laa, sun2025towards, yu2025unlocking}, video~\cite{yan2025generalizing, nguyen2025vulnerability}, or feature levels~\cite{yan2024transcending, cheng2024can}, are widely used to enrich forgery diversity and mitigate overfitting. 

\noindent\textbf{Face Anti-Spoofing (\texttt{FAS})} primarily targets domain shifts across presentation attacks, \eg, sensors and materials, to detect face liveness. Domain generalization (DG) methods have been employed to learn domain-invariant features via adversarial~\cite{liu2022spoof, shao2019multi, jia2020single, wang2022domain, zhou2023instance}, contrastive~\cite{wang2022domain, sun2023rethinking, liu2023towards}, test-time~\cite{zhou2024test}, and continual~\cite{cai2025rehearsal} learning. Recent studies demonstrate the cross‑domain robustness of source-free domain adaptation~\cite{liu2024source, li2025optimal} and priors from instance~\cite{zhou2023instance}, prototype~\cite{hu2024rethinking}, and domain~\cite{le2024gradient}. With auxiliary depth and infrared supervisions beyond RGB, generalized multi-modal \texttt{FAS}~\cite{lin2025reliable, lin2024suppress, yang2025dadm} revisits modality imbalance and alignment.

Most existing face security methods are task-specific, built on generic VFMs without facial domain focus, employ divergent backbones, and require full fine-tuning or bespoke adaptation, with non-trivial, specialized designs per task, lacking a universal and generalizable facial representation that transfers across diverse face security tasks effectively and efficiently.

\section{FS-VFM Pre-training Architecture}
\label{sec:framework}

To improve generalizability across diverse downstream face security tasks, we focus on learning intrinsic, fundamental, and transferable facial representations from unlabeled real faces. As illustrated in~\Cref{fig:framework}, the proposed FS-VFM pre-training framework comprises two complementary pretext tasks for self-supervised learning (SSL): masked image modeling (MIM) and instance discrimination (ID). The MIM network ($E_o \circ D_o$), a masked autoencoder (MAE)~\cite{he2022masked} driven by our CRFR-P facial masking strategy, reconstructs the masked face to explicitly promote meaningful intra-region \textit{\textbf{C}}onsistency and enforce challenging inter-region \textit{\textbf{C}}oherency. In parallel, the ID network employs the MIM encoder ($E_o$) in its online branch ($E_o \circ D_o^r \circ \mathit{proj} \circ \mathit{pred}$) to process masked local views, where the target branch ($E_t \circ D_t^r \circ \mathit{proj}$) distills unmasked global views, to establish underlying local-to-global \textit{\textbf{C}}orrespondence. These three pre-training objectives, termed \textit{\textbf{3C}}, collectively endow the online encoder ($E_o$) with pixel-level context perceptiveness, region-level relation awareness, and instance-level face invariance.

This section outlines the architecture and pre-training objectives of FS-VFM. In~\Cref{sec:studies}, we delve deeper into its key components and the design rationales.
%TODO: could move to the end of introduction

%-------------------------------------------------------------------------
\subsection{Facial MIM with Local Perception}
\label{subsec:MIM}
In a nutshell, the MIM network ($E_o \! \circ \! D_o$) in FS-VFM is an MAE~\cite{he2022masked} model steered by our CRFR-P masking strategy, reconstructs masked patches using only visible ones. Let $x_{f}=\left\{x_{i}\right\}_{i=1}^{N}$ denote the full set of $N$ non-overlapping patches split from an input face image $I$.

\noindent\textbf{CRFR-P Masking} The mask sampling strategy plays a critical role in MIM for both representation quality and downstream performance. Building on our studies in~\cref{sec:studies}, we introduce CRFR-P, \underline{C}overing a \underline{R}andom \underline{F}acial \underline{R}egion followed by \underline{P}roportional facial masking strategy, as shown in Fig.~\ref{fig:framework} and Alg.~\ref{alg:CRFR_P}. CRFR-P first partitions facial parts into predefined semantic regions $\mathit{FR}=$ \{\textit{eyebrows, eyes, mouth, face boundary, nose, hair, skin, background}\} using an off-the-shelf face parser. Next, it entirely masks all patches within a randomly selected region $\mathit{fr}\notin$ \{\textit{skin, background}\} and obtains the facial region mask $M_\mathit{fr}\in\{0,1\}^{N}$, where $0$ for the visible and $1$ for the masked patch. Then, based on the number of already masked patches and the overall masking ratio $r$, it randomly masks an equal portion of patches across each of the remaining $\{\mathit{FR}$$-$$\mathit{fr}\}$ regions to generate the image mask $M\in\{0,1\}^{N}$. Finally, CRFR-P returns both the image mask $M$ and the facial region mask $M_\mathit{fr}$.

\begin{algorithm}[t!]
\renewcommand{\thealgorithm}{1}
\caption{CRFR-P Masking Strategy}
\footnotesize  % \footnotesize scriptsize（smaller）
{\bf Input:} Real face image $I$, Masking ratio $r$\\
{\bf Output:}  Image mask $M$, Facial region mask $M_\mathit{fr}$
\begin{algorithmic}[1]
\State $\mathit{PM} \gets \mathit{Face\_Parser}(I)$
\State $P_\mathit{pm} \in \mathbb{R}^N \gets \mathit{patchify}(\mathit{PM})$
\State $M, M_\mathit{fr} \gets [0] \in \mathbb{R}^N, [0] \in \mathbb{R}^N$
\State $\mathit{FR} \! \leftarrow \! \{$eyebrows \! $\supseteq \! [$right eyebrow, left eyebrow$]$, eyes \! $\supseteq \! [$right eye, left eye$]$, mouth \! $\supseteq \! [$upper lip, inner mouth, lower lip$]$, face boundary \! $\supseteq \! [$skin$\cap$background, skin$\cap$hair$]$, nose, hair, skin, background\}
\State Randomly select a $\mathit{fr} \in \{\mathit{FR} - \{\mathit{skin}, \mathit{background}\}\}$ 
\State $M_\mathit{fr}[P_\mathit{pm} \cap \mathit{fr}] \gets 1$ \Comment{\textit{\textbf{C}overing a \textbf{R}andom \textbf{F}acial \textbf{R}egion}}
\If{$\sum\! M_{\mathit{fr}} > N \cdot r$} \Comment{\textit{Extreme-case}}
    \State Randomly unmask $(\sum\! M_{\mathit{fr}} - N \cdot r)$ patches in $M_{\mathit{fr}}$
    \State $M \gets M_\mathit{fr}$
    \State \textbf{break}
\EndIf
\State \textbf{end if}
\State $M \gets M_\mathit{fr}$
\For{$pr \in \{\mathit{FR} - \{\mathit{fr}\}\}$} \Comment{\textit{\textbf{P}roportional masking in other regions}}
    \State $r=(N \cdot r-\sum\! M)~ / ~(N-\sum\! M)$
    \State $M[(P_\mathit{pm} \cap pr) \cdot r] \gets 1$
\EndFor
\State \textbf{end for}
\State \textbf{Return:} $M, M_\mathit{fr}$
\end{algorithmic}
\label{alg:CRFR_P}
\end{algorithm}

\noindent\textbf{Online Encoder} $E_{o}$ operates exclusively on visible patches $x_{v} \leftarrow M \odot x_{f}$, and maps $x_{v}$ into latent features $z_o^v$, where $\odot$ denotes the element-wise product for masking and $\leftarrow$ selects the visible ones. Following ViT~\cite{dosovitskiy2020image} and MAE~\cite{he2022masked}, the online encoder first embeds the visible patches $x_{v}$ by a linear projection as patch embeddings, adds corresponding positional embeddings $p_{v}$, and passes the fused embeddings through a series of Transformer blocks to produce $z_o^v$:
\begin{equation}
z_o^v=E_o(x_v+p_v).
\end{equation}

\noindent\textbf{Online Decoder} $D_{o}$ reconstructs the pixels of the input image. It first concatenates the encoded visible token $z_o^v$ with learnable mask tokens $z_o^m$, and appends relative positional embeddings to form the full token set $z_o^f$. The online decoder, another stack of transformer blocks, receives $z_o^f$ as input, followed by a linear head to restore the masked patches:
\begin{equation}
I_m^{\prime}=(1-M)\odot D_o(z_o^f).
\end{equation}

\noindent\textbf{MIM Objective} Following~\cite{he2022masked}, we adopt normalized pixels as the reconstruction target and minimize the mean squared error (MSE) loss over masked patches between the predicted $I_{m}^{\prime}$ and the original $I_m\leftarrow(1-M) \odot I$:
\begin{equation}
\mathcal{L}_\mathit{rec}^m=\frac1{N_m}\sum\nolimits_{i=1}^{N_m}\left(I_m^{(i)}-I_m^{'(i)}\right)^2,
\label{eq:loss_rec_m}\end{equation}
where $N_m$$=$$N$$\times$$r$$=$$\sum\! M$ is the number of masked patches. Additionally, our CRFR-P masking strategy provides a supplementary mask $M_\mathit{fr}$. As a sub-mask of $M$, it covers all patches placed in the randomly selected facial region $I_m^\mathit{fr}\leftarrow(1-M_\mathit{fr})\odot I$. To reinforce inter-region coherency and prevent trivial solutions, we apply an auxiliary reconstruction loss to the masked patches of the facial region $\mathit{fr}$:
\begin{equation}
\mathcal{L}_\mathit{rec}^\mathit{fr}=\frac1{N_\mathit{fr}}\sum\nolimits_{j=1}^{N_\mathit{fr}}\left(I_m^\mathit{fr(j)}-I_m^{\mathit{fr}^{\prime}(j)}\right)^2,
\label{eq:loss_rec_fr}
\end{equation}
where $N_\mathit{fr}$$=$$\sum\! M_\mathit{fr}$ is the number of patches in $\mathit{fr}$, and $I_m^{\mathit{fr}^{\prime}}\leftarrow(1-M_\mathit{fr})\odot I_m^{\prime}$ be the decoder’s predictions for that region. Thus, the overall MIM objective becomes a weighted sum to update the MAE ($E_o \! \circ \! D_o$) network:
\begin{equation}
\mathcal{L}_\mathit{rec}=\mathcal{L}_\mathit{rec}^m+\lambda_\mathit{fr}\mathcal{L}_\mathit{rec}^\mathit{fr}.
\label{eq:loss_mim}
\end{equation}

%-------------------------------------------------------------------------
\subsection{Facial ID with Global Alignment}
\label{subsec:ID}

In a nutshell, the ID network in FS-VFM features symmetric designs between the online and target branches \wrt the encoder and representation decoder, while adopting asymmetric designs \wrt the input view, projection, negative-free loss, and model updates. These designs, tailored for face security tasks, complement MIM with more precise and reliable global semantic alignment, distinguishing from prior JEA (joint embedding architecture) works that graft ID onto MIM.

\noindent\textbf{Target Encoder} 
$E_{t}$ receives full patches $x_f=\{x_i\}_{i=1}^N$ as the target view to yield target latent features $z_t^f$, which prompt the online encoder $E_{o}$ in learning holistic representations. Passing all patches through $E_{t}$ is crucial for embedding complete facial semantics to steer the online encoder $E_{o}$ toward coherent local-to-global representations. Thus, the target encoder $E_{t}$ acts as a teacher that shares the same structure as the student $E_{o}$. Analogously, with positional embeddings $p_{f}$ of full patches, $E_{t}$ produces global embeddings:
\begin{equation}
z_t^{f}=E_t(x_f+p_f)
\end{equation}

\noindent\textbf{Online Rep Decoder} $D_{o}^r$ transforms the full tokens $z_o^f$ into online representations $r_o$. Unlike the online decoder $D_{o}$, which restores raw pixel values, $D_o^r$ recovers the representations of masked tokens to align with the uncorrupted target. $D_{o}^r$ resembles the structure of $D_{o}$ but has significantly shallower transformer blocks, followed by a linear layer that predicts features. The token features are output via a simple mean pooling as the online representations:
\begin{equation}
r_o=D_o^r(z_o^f)
\end{equation}

\noindent\textbf{Target Rep Decoder} 
In the target branch, the momentum encoder $E_{t}$ is updated using past iterations of the online encoder $E_{o}$, which also serves for MIM. This gap makes it suboptimal to directly match $r_o$ with the target embeddings $z_t^f$, as the model may struggle to recover high-level target features while restoring low-level pixel values. Thus, we add a target rep decoder $D_{t}^r$ that mirrors $D_{o}^r$ to represent the target features in the same disentangled space:
\begin{equation}
r_t=D_t^r(z_t^f)
\end{equation}

\noindent\textbf{ID Objective} Following the asymmetric projector/predictor design in~\cite{grill2020bootstrap, chen2021exploring, chen2021empirical}, we employ a projector followed by a predictor to map the online representation $r_o$ to a lower-dimensional vector $v_o^p$, and use only a projector for the target representation $r_t$ to obtain $v_t$. We minimize the negative cosine similarity~\cite{chen2021exploring} between these two $\ell_2$-normalized vectors:
\begin{equation}
\mathcal{L}_{sim}(v_o^p,\mathrm{sg}[v_t])=-\frac{v_o^p}{\|v_o^p\|_2}\cdot\frac{v_t}{\|v_t\|_2},
\label{eq:loss_id}
\end{equation}
where $\mathrm{sg}[\cdot]$ is a stop-gradient, \ie, gradients are only calculated \wrt the online branch ($E_o \circ D_o^r \circ \mathit{proj} \circ \mathit{pred}$). The parameters $\theta_{t}$ of the target branch ($E_t \circ D_t^r \circ \mathit{proj}$) are updated by an exponential moving average (EMA)~\cite{grill2020bootstrap} from the online counterparts $\theta_t\leftarrow\tau\theta_t+(1-\tau)\theta_o$. Note that our $\mathcal{L}_{sim}$ is asymmetric due to the different input views (\ie, masked versus full patches) for the two branches, unlike the symmetrized loss for both sides~\cite{grill2020bootstrap, chen2021exploring, chen2021empirical, caron2021emerging}.

%-------------------------------------------------------------------------
\subsection{Joint Objective for Foundational Face Representation}
\noindent\textbf{Overall Loss} FS-VFM learns foundational representations of real faces by jointly tackling the MIM (\Cref{eq:loss_mim}) and the ID (\Cref{eq:loss_id}) pretext task. Thus, the overall pre-training objective is a weighted sum:
\begin{equation}
\mathcal{L}=\mathcal{L}_\mathit{rec}+\lambda_\mathit{cl}\mathcal{L}_\mathit{sim}\overset{\Cref{eq:loss_mim}}{\operatorname*{=}}\mathcal{L}_\mathit{rec}^m+\lambda_\mathit{fr}\mathcal{L}_\mathit{rec}^\mathit{fr}+\lambda_\mathit{cl}\mathcal{L}_\mathit{sim}.
\end{equation}

\noindent\textbf{Scalable Facial Learners} FS-VFM can be readily pre-trained on various real face datasets or arbitrary combinations thereof, without annotations, to learn a general facial representation that transcends specific domains or tasks. Thus, it can benefit from the larger and more diverse unlabeled faces widely available in the open world. Built upon the standard ViT architecture, FS-VFM scales seamlessly across ViT variants without backbone modifications specific to face security.

\section{Diving Deep into FS-VFM}
\label{sec:studies}

% This section further illuminates the underpinning mechanisms and design rationales of pre-training FS-VFM (\Cref{sec:framework}). At the core lies the CRFR-P masking strategy, which drives MIM (masked image modeling) to promote both intra-region consistency and inter-region coherency, while also fostering robust local-to-global correspondence by delivering partial (masked) views to ID (instance discrimination). Thus, we first explore alternative facial masking strategies, with intuitive insights that motivate the formulation of CRFR-P (\Cref{subsec:masking_strategies}). We then quantitatively and qualitatively evaluate how these masking strategies shape attention or representation behaviors of the MIM model (\Cref{subsec:reveal}). Finally, we elaborate on our ID branch and compare it with other JEA (joint embedding architectures) for SSL (self-supervised learning), clarifying its distinctive advantages in strengthening reliable and discriminative representations (\Cref{subsec:id}). These studies explain why FS-VFMs are transferable, robust, and generalized across diverse face security tasks.

This section further illuminates the underpinning mechanisms and design rationales of pre-training FS-VFM (\Cref{sec:framework}). We first explore alternative facial masking strategies, with intuitive insights that motivate the formulation of CRFR-P (\Cref{subsec:masking_strategies}). We then quantitatively and qualitatively evaluate how these masking strategies shape attention or representation behaviors of the MIM (masked image modeling) (\Cref{subsec:reveal}). Finally, we elaborate on our ID (instance discrimination) branch and compare it with other JEA (joint embedding architectures), clarifying its distinctive advantages in strengthening reliable and discriminative representations (\Cref{subsec:id}). These studies explain why FS-VFMs are transferable, robust, and generalized across face security tasks.

%-------------------------------------------------------------------------
\subsection{Facial Masking Strategies: from Random to CRFR-P}
\label{subsec:masking_strategies}

{
\setlength{\abovecaptionskip}{0pt}
\begin{figure}[tb!]
\centering
\includegraphics[width=.9\linewidth]{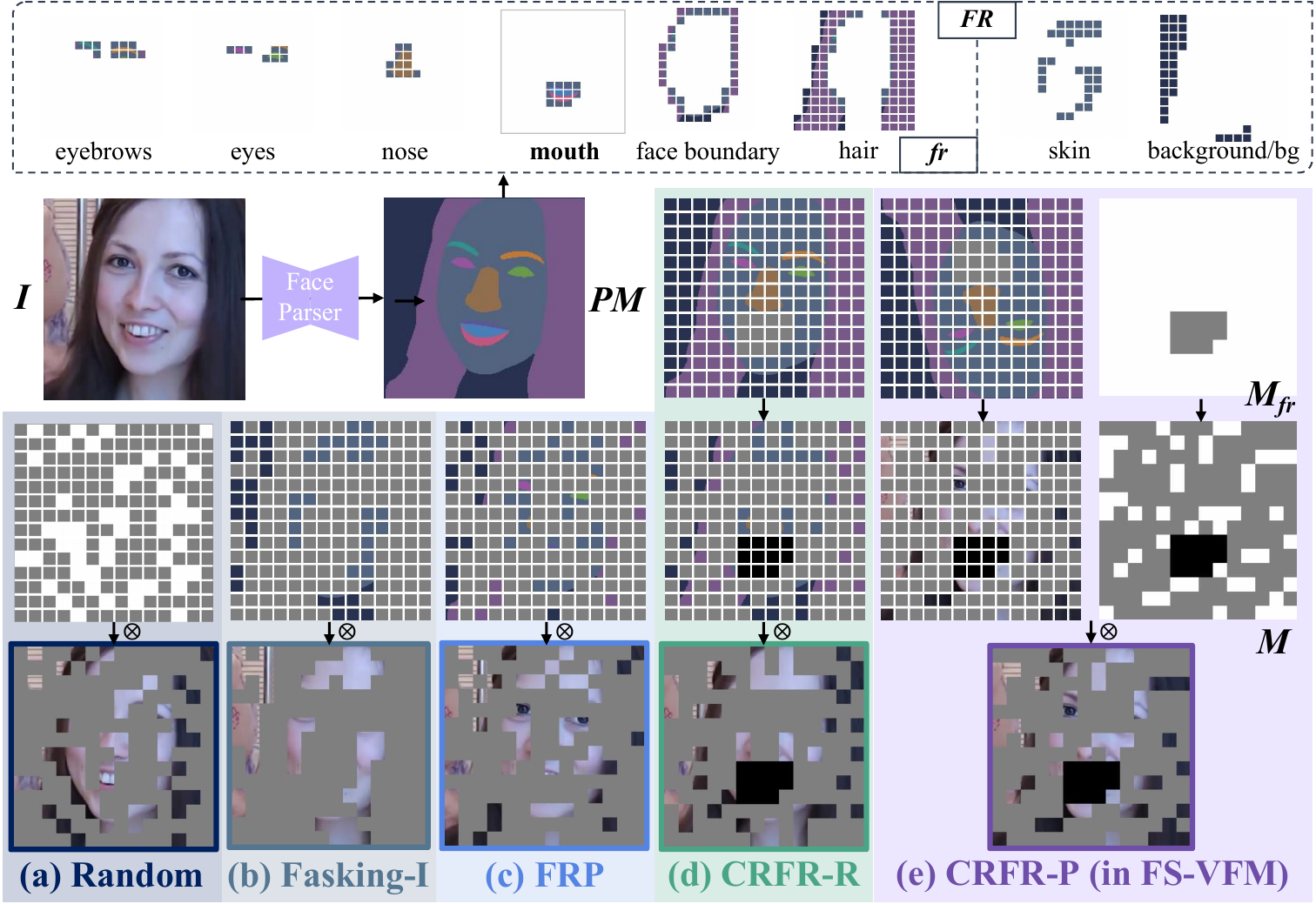}
% \vspace{-15pt}
\caption{\label{fig:MIM_mask_stra} Comparison of masking strategies for face images (75\% masking ratio). (a) Simple random masking. (b) Fasking-I, adapted from MARLIN~\cite{cai2023marlin}, priority masking regions $\notin$ \{bg, skin\}. (c) Our FRP masking for intra-region consistency: Proportional masking within each Facial Region $\in$ \{$\mathit{FR}$\}. (d) Our CRFR-R masking for inter-region coherency: Covering a Random Facial Region $\in$ \{$\mathit{fr}$\} and then Random masking other patches. (e) Our CRFR-P masking for both intra-region consistency and inter-region coherency: Covering a Random Facial Region $\in$ {$\mathit{fr}$} and then Proportional masking other regions $\in$ \{$\mathit{FR}-\mathit{fr}$\}. All masks are binary (black solely highlights $\mathit{fr}$).}
\end{figure}
}

\subsubsection{Motivation}\label{subsec:motivation}

Simple random masking with a high ratio is widely employed in both natural~\cite{he2022masked, xie2022simmim} and facial~\cite{zheng2022general, wang2023toward} MIM, yet it ignores facial inductive bias, impeding the learned facial representations. As our sole focus, human faces, which comprise well-defined regions with heterogeneous textures, we opt to segment facial semantics explicitly rather than learning additional masking modules~\cite{kakogeorgiou2022hide, shi2022adversarial, wang2023hard}, for reasonable and efficient facial mask sampling. With an off-the-shelf face parser to divide facial parts, MARLIN~\cite{cai2023marlin} proposes a masking strategy named Fasking for facial video MIM. We adapt it to image as Fasking-I, as shown in~\Cref{fig:MIM_mask_stra} (b), which partitions facial parts into $\{$\textit{left-eye, right-eye, nose, mouth, hair, skin, background}$\}$ and prioritizes masking non-skin and non-background regions. However, as visible tokens stem mainly from skin or background, Fasking-I struggles to preserve sufficient facial details crucial for security tasks. 

% For more effective facial masking, we explore the intrinsic properties of real faces. Unlike diverse manipulations posed in forged/spoofing faces, authentic/live faces generally maintain a natural, realness appearance. Drawing on FACS~\cite{ekman1978facial} and facial psychology~\cite{russell1997psychology, haxby2000distributed}, we articulate these local patterns as intra-region consistency and inter-region coherency: intra-region consistency means similar textures or features within the same facial region, \eg, consistent pupil color or symmetrical nostril; inter-region coherency exhibits facial semantic correlations for a cohesive look, \eg, a grin‌ co-occurs with curved eyes. In contrast, forged/spoofed faces often disrupt these endogenous patterns. However, tailoring an efficient masking strategy for these properties is non-trivial.

For more effective facial masking, we explore the intrinsic properties of real faces. Unlike diverse manipulations posed in forged/spoofing faces, authentic/live faces generally maintain a natural, realness appearance. Drawing on FACS~\cite{ekman1978facial} and facial psychology~\cite{russell1997psychology, haxby2000distributed}, we articulate these local patterns as intra-region consistency, which means similar textures or features within the same facial region, \eg, consistent pupil color or symmetrical nostril; and inter-region coherency, which exhibits facial semantic correlations for a cohesive look, \eg, a grin‌ co-occurs with curved eyes. In contrast, manipulated faces often disrupt these endogenous patterns. However, tailoring an efficient masking strategy for these properties is non-trivial.

{
\setlength{\abovecaptionskip}{0pt}
\begin{figure}[t!]
\centering
\includegraphics[width=\linewidth]{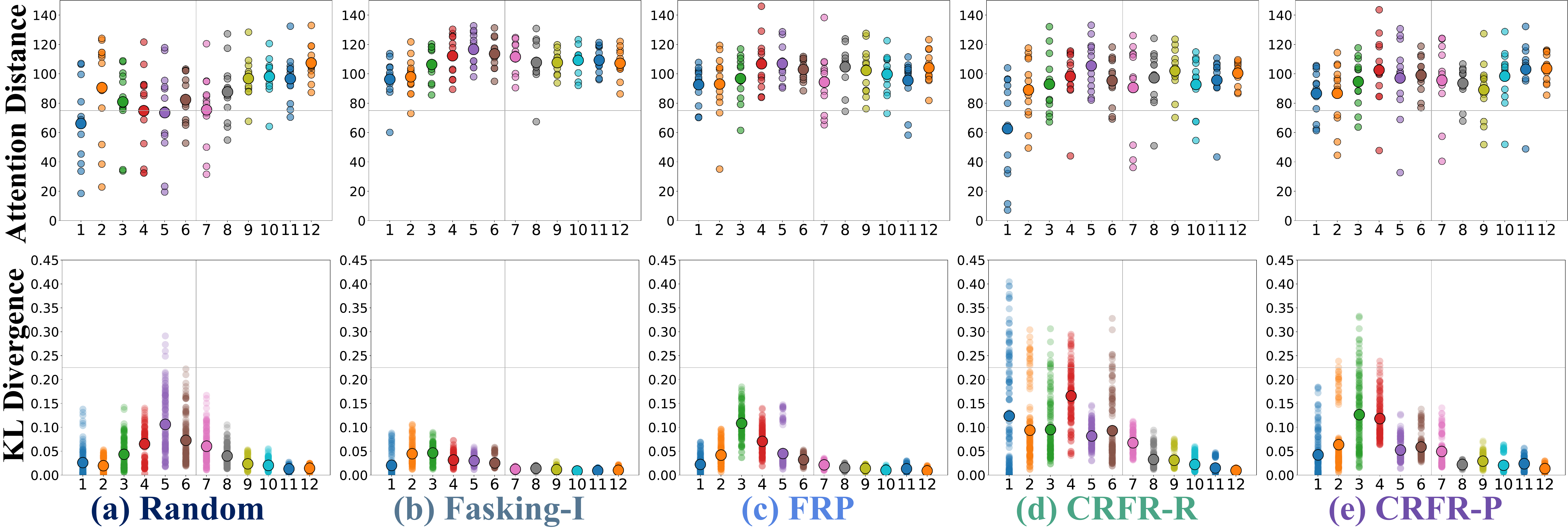}
\vspace{-10pt}
\caption{Mean attention distance~\cite{dosovitskiy2020image} (\textit{Top}, global ↑) and Kullback-Leibler divergence~\cite{xie2023revealing} (\textit{Bottom}, diverse ↑) of each attention head (small dot) across all blocks (\textit{x-axis}) in the MAE~\cite{he2022masked} ViT-B/16 encoder pre-trained by different facial masking strategies, with the average one (large dot) for each block.}
\label{fig:mask_analysis} 
\end{figure}
}

\subsubsection{Intuition} \label{subsec:intuition} 
As shown in~\Cref{fig:MIM_mask_stra}, simple random masking and Fasking-I are susceptible to fully occluding small, informative regions (\eg, eyes), which impedes accurate learning of rich textures therein (\eg, eyelids, pupils, iris). To promote intra-region consistency, we formulate FRP, \underline{F}acial \underline{R}egion \underline{P}roportional masking: randomly masks patches within each region in the same proportion. This intuitive way ensures that all regions retain visible patches, steering attention to the same region when restoring masked patches. Yet, it also risks a potential shortcut, \ie, restoring masked patches directly from adjacent unmasked patches in the same region, may yield a trivial reconstruction and neglect cross-region relationships. To foster inter-region coherency, we devise CRFR-R, \underline{C}overing a \underline{R}andom \underline{F}acial \underline{R}egion followed by \underline{R}andom masking: a randomly selected facial region is fully masked and must be inferred from visible patches outside it, forcing the model to learn its correlations \wrt other regions. However, the subsequent random masking may again obscure small regions elsewhere, compromising the intra-region consistency of them. As our preliminary masking strategies, FRP and CRFR-R exhibit individual constraints but complement each other.

\subsubsection{Design of CRFR-P}\label{subsec:design}
Building upon the above insights, we propose the \underline{C}overing a \underline{R}andom \underline{F}acial \underline{R}egion followed by \underline{P}roportional facial masking strategy, a straightforward design illustrated in~\Cref{fig:MIM_mask_stra} (e) and~\Cref{alg:CRFR_P}. The facial regions divided by CRFR-P (also FRP and CRFR-R), \ie, $\mathit{FR}$, differ from those of Fasking-I: similar parts are merged into a distinctive region (\eg, eyes), avoiding the shortcut restoration of the fully masked region (left-eye) from a proportionally masked region (right-eye). We reserve $M_\mathit{fr}$ to calculate the auxiliary reconstruction loss in~\Cref{eq:loss_rec_fr} for the fully masked region, as an arduous task that emphasizes long-range dependencies. This induces negligible overhead because $M_\mathit{fr}$ is a prerequisite for computing $M$. Despite its simplicity, CRFR-P masking poses a non-trivial and meaningful facial MIM task, which not only avoids the shortcut solution but also naturally resolves the major challenge: promoting both intra-region consistency and inter-region coherency.
% Despite its simplicity, CRFR-P masking poses a non-trivial and meaningful facial MIM pretext task, which naturally resolves the major challenge of promoting both intra-region consistency and inter-region coherency, by effectively directing attention to critical facial regions with appropriate range and diversity.

{
\setlength{\abovecaptionskip}{0pt}
\begin{figure}[t!]
\centering
\includegraphics[width=.9\linewidth]{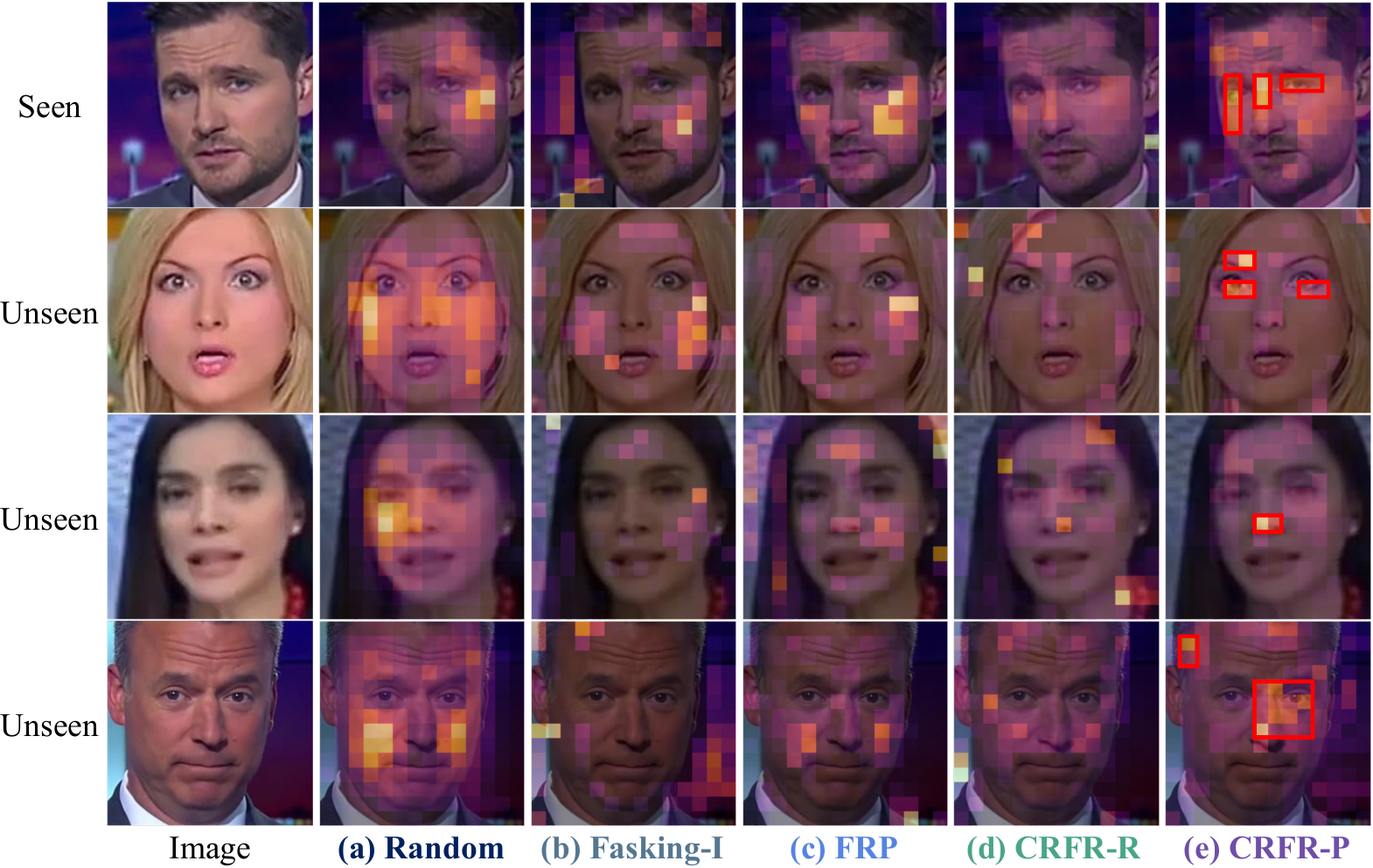}
% \vspace{-15pt}
\caption{Visualization of the self-attention map averaged across all heads from the last block of the ViT-B/16 encoder pre-trained by MAE~\cite{he2022masked} with different facial masking strategies.}
\label{fig:MIM_att_mask_mae}
\end{figure}
}

%-------------------------------------------------------------------------
\subsection{Impacts of Masking Strategies on Facial MIM}
\label{subsec:reveal}

% How do different facial masking strategies affect the MIM pre-trained model or its learned representations? Except for downstream ablations (\Cref{subsec:exp_ablation}), we analyze the attention properties. Since most MIM models, including ours, are built upon the ViT~\cite{dosovitskiy2020image} blocks, whose main component, the attention mechanism, is naturally interpretable~\cite{xie2023revealing}. In this subsection, we employ the vanilla MAE (\ie, our MIM network) with a ViT-B/16 encoder at a 75\% mask ratio, and pre-train it on real faces (\ie, FF++\_O~\cite{rossler2019faceforensics++}, our default dataset for ablations). We alter only the masking strategy across simple random, Fasking-I, FRP, CRFR-R, and CRFR-P, then examine attention behaviors in the pre-trained encoders: \emph{1) mean attention distance}, to measure the distribution from local to global; \emph{2) Kullback-Leibler (KL) divergence}, to evaluate the diversity among attention heads; \emph{3) self-attention map visualizations}, to uncover salient regions of focus.

How do different facial masking strategies affect the MIM pre-trained model or its representations? Most MIM models, including ours, are built upon the ViT~\cite{dosovitskiy2020image} blocks, whose main component, the attention mechanism, is naturally interpretable~\cite{xie2023revealing}. In this subsection, we employ the vanilla MAE (\ie, our MIM network) with a ViT-B/16 encoder at a 75\% mask ratio, and pre-train it on real faces (\ie, FF++\_O~\cite{rossler2019faceforensics++}, our default dataset for ablations). We alter only the masking strategy across simple random, Fasking-I, FRP, CRFR-R, and CRFR-P, then examine attention behaviors in the pre-trained encoders: \emph{1) mean attention distance}, to measure the distribution from local to global; \emph{2) Kullback-Leibler (KL) divergence}, to evaluate the diversity among attention heads; \emph{3) self-attention map visualizations}, to uncover focused regions.

% {
% \setlength{\abovecaptionskip}{0pt}
% \begin{figure*}[t!]
% \centering
% \includegraphics[width=.825\linewidth]{fig/mask_analysis.pdf}
% % \vspace{-5pt}
% \caption{Mean attention distance~\cite{dosovitskiy2020image} (\textit{Top}, ↑ global) and Kullback-Leibler divergence~\cite{xie2023revealing} (\textit{Bottom}, ↑ diverse) of each attention head (small dot) across all blocks (\textit{x-axis}) in the MAE~\cite{he2022masked} ViT-B/16 encoder pre-trained by different facial masking strategies, with the average one (large dot) for each block.}
% \label{fig:mask_analysis} 
% \end{figure*}
% }

\subsubsection{Local or Global Patterns?} 
To investigate whether the pre-trained model looks over local details or global context, we compute the mean attention distance~\cite{dosovitskiy2020image} for each attention head across all transformer blocks/layers, as plotted in~\Cref{fig:mask_analysis} (\textit{Top}). The model (MAE ViT-B/16 encoder) pre-trained with simple random masking exhibits more local attention in shallower blocks and gradually shifts to global attention in deeper, similar to supervised ViTs~\cite{dosovitskiy2020image}. Fasking-I shows large distances from the outset \ie, primarily global attention, as the visible patches are mostly sampled from broad background/skin regions. FRP masking also increases attention distances, but slightly lower than Fasking-I, since FRP keeps visible patches evenly distributed across all facial regions. When applying CRFR-R, one entire facial region is blanked out before random masking, which pivots attention to the disparate regions, yielding more global attention in the intermediate (3\textsuperscript{rd} to 8\textsuperscript{th}) blocks relative to the simple random masking counterparts. In contrast, after covering a region, CRFR-P proportionally masks the remaining regions rather than randomly masking patches, which retains the visibility across those regions, leading to a more global 1\textsuperscript{st} block than CRFR-R. Compared with FRP, before proportional masking, CRFR-P fully masks a region, which tightens the masking budget and exposes more visible patches in the remaining regions, achieving more local attention than FRP. 

By comparison, the model pre-trained with CRFR-P combines the effects of FRP and CRFR-R, delivering well-balanced attention distances throughout blocks and paying appropriate attention to both local details and global context.

\subsubsection{Similar or Different Tokens?}
To explore whether the pre-trained model attends to similar or different tokens, we follow~\cite{xie2023revealing} to calculate the Kullback-Leibler (KL) divergence for each attention head across all blocks, as plotted in~\Cref{fig:mask_analysis} (\textit{Bottom}). The model pre-trained with Fasking-I exhibits lower KL divergences across all heads, indicating limited diversity due to restricted (skin/background) regions dominating visible tokens. Interestingly, the proportional mask sampling also decreases the attention diversity, as evidenced by a lower KL divergence in FRP versus the Random and CRFR-P versus the CRFR-R counterparts. This is likely because proportional masking exposes patches in a more homogeneous pattern, \ie, drawn from each facial region. Conversely, covering a random facial region increases attention diversity, as observed in comparisons between CRFR-R versus Random, and between CRFR-P versus FRP. In essence, when fully masking a randomly selected facial region, the model cannot overly rely on any single region and is forced to inspect others. 

As a result, in terms of attention diversity, the FRP pre-trained model yields somewhat homogeneous heads, while CRFR-R shows overly heterogeneous ones. CRFR-P again strikes a balance that provides sufficient diversity without excessive dispersion, and attends to varied yet robust tokens.

{
\setlength{\abovecaptionskip}{0pt}
\begin{figure*}[htb!]
\centering
\includegraphics[width=.875\linewidth]{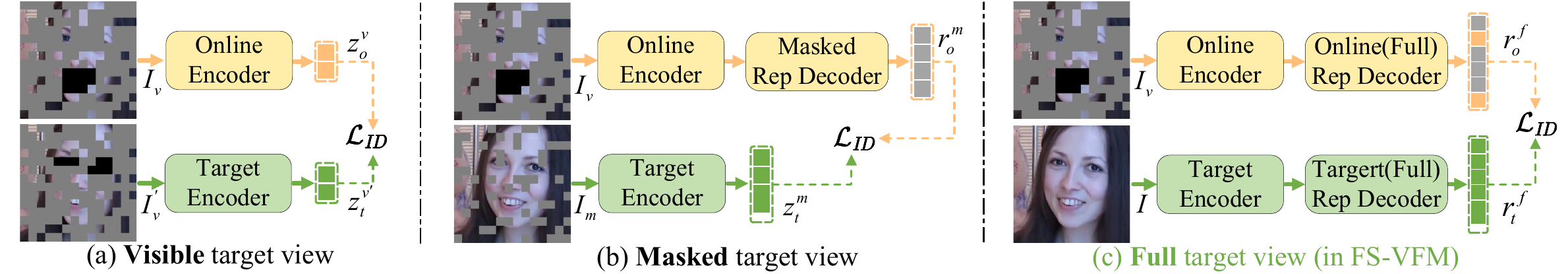}
% \vspace{-15pt}
\caption{Comparison of target views \& network couplings adapted for FS-VFM, drawn from JEA-based self-supervised pre-training methods. (a) Visible patches from a different mask~\cite{assran2023self, li2023mage, zhao2024asymmetric, wang2023toward}. (b) Masked patches from the same mask~\cite{chen2024context, wei2024towards}. (c) Full patches without masking~\cite{zhou2021ibot, DBLP:conf/iclr/YiGLYLWSQ23, huang2023contrastive, tao2023siamese, eymael2024efficient, jamal2025multi}.}
\label{fig:ID_target_view}
\end{figure*}
}

\subsubsection{Key or Trivial Focus?} 
To reveal whether the pre-trained model focuses on key or trivial regions, beyond quantitative analyses, we visualize the mean attention map from the last block and overlay it on the input face in~\Cref{fig:MIM_att_mask_mae}, as the pretext decoder or downstream head typically follows the final block. Under random masking, attention regions appear to cover the face, but primarily on large skin areas that can be trivially recovered from adjacent visible pixels. This suggests that the model solves the MIM task by shortcuts rather than learning meaningful features from challenging facial regions. Fasking-I behaves similarly and attends mostly to skin/background. Although FRP and CRFR-R broaden attention regions beyond skin/background, they still struggle to pinpoint the salient features. By contrast, CRFR-P consistently highlights key regions like the nose and eyes, focusing on meaningful region-level representations beyond superficial low-level pixel values.

% In summary, to encode both intra-region consistency and inter-region coherency for facial MIM, our CRFR-P masking effectively directs attention to key facial regions with appropriate distances and diversity across blocks, promoting the model to learn intrinsic properties of real faces while avoiding collapses. Furthermore, we hope these studies provide new insights into learning fundamental facial representations.

In sum, to encode both intra-region consistency and inter-region coherency for facial MIM, our CRFR-P masking effectively directs attention to key facial regions with appropriate distances and diversity across blocks, promoting the model to learn intrinsic properties of real faces while avoiding collapses. 

%-------------------------------------------------------------------------
\subsection{Connections and Analyses of Facial ID}
\label{subsec:id}

% We now clarify the relations and distinctions between our method and existing joint-embedding-architecture (JEA)-based works for self-supervised learning (SSL), \ie, integrate instance discrimination (ID) or Siamese encoder, with masked image modeling (MIM) or degraded input. Although prior efforts have shown efficacy in natural visual and facial analysis, our empirical studies suggest that face security tasks demand finer and more precise alignment within the ID network. FS-VFM addresses this need by refining: \emph{1) target view \wrt network coupling}, \emph{2) data augmentation}, and \emph{3) loss formulation}, to foster a reliable local-to-global correspondence.

We now clarify how our approach relates to and differs from existing JEA (joint embedding architectures) works for SSL (self-supervised learning), \ie, integrate the ID or Siamese encoder, with the MIM or degraded input. Although prior efforts have shown efficacy in natural vision and facial analysis, our empirical studies suggest that face security tasks demand finer and more precise alignment within the ID network. FS-VFM addresses this by refining: \emph{1) target view \wrt network coupling}, \emph{2) data augmentation}, and \emph{3) loss formulation}, to foster a reliable local-to-global correspondence.

\subsubsection{Target view \& Siamese network}
In most JEA-based frameworks, the online (student) branch processes visible patches from the masked image, while the target (teacher) branch varies. Accordingly, for FS-VFM, we explore different target views and corresponding network paradigms in~\Cref{fig:ID_target_view}: \emph{(a) Visible patches from a different mask}~\cite{assran2023self, li2023mage, zhao2024asymmetric, wang2023toward}: the online and target encoders yield latent features $z_o^v$ and $z_t^v$ that are directly contrasted; \emph{(b) Masked patches from the same mask}~\cite{chen2024context, wei2024towards}: to align with the target features $z_t^m$ of masked patches, the online branch uses a masked representation (rep) decoder that predicts masked representations $r_o^m$ from the visible tokens $z_o^v$. This decoder takes learnable masked tokens (as $\mathit{Q}$) and full tokens (as $\mathit{K}$ and $\mathit{V}$) to compute cross-attention, which follows the latent regressor in CAE~\cite{chen2024context}; \emph{(c) Full patches without masking}~\cite{zhou2021ibot, DBLP:conf/iclr/YiGLYLWSQ23, huang2023contrastive, tao2023siamese, jamal2025multi}: some decoder-free methods~\cite{zhou2021ibot, DBLP:conf/iclr/YiGLYLWSQ23} directly match online visible features $z_o^v$ with full target features $z_o^f$ by optimizing $\mathcal{L}_\mathit{ID}(z_o^v, z_t^f)$. In contrast, CMAE~\cite{huang2023contrastive} attaches a feature (rep) decoder after the online encoder to aid alignment, \ie, $\mathcal{L}_\mathit{ID}(z_o^v, z_t^f)$. Further, we introduce an additional target rep decoder, \ie, Siamese rep decoders for both branches, and compute $\mathcal{L}_\mathit{ID}(r_o^f,r_t^f)$ in the same, disentangled space, further bridging the distribution gap from low-level pixels to high-level semantics.

% Our ablations on downstream face security tasks show that FS-VFM performs better when using \emph{(c) full patches} as the target view, along with Siamese rep decoders. By predicting complete facial embeddings from partially visible patches, the ID network reconciles global and local views of the same face. Building on this alignment, FS-VFM structures the representation space through ``local-to-global'' correspondence, thus endowing its encoder with improved facial discriminability.

Our ablations on downstream face security tasks show that FS-VFM performs better when using \emph{(c) full patches} as the target view, along with Siamese rep decoders. By predicting complete facial embeddings from partially visible patches, the ID network structures the representation space through ``local-to-global'' correspondence, thereby endowing the encoder with improved facial discriminability.

\subsubsection{Data augmentation}
Most ID methods rely heavily on aggressive data augmentations, including spatial and color enhancements, to avoid model collapse~\cite{chen2020simple, he2020momentum, chen2021empirical, grill2020bootstrap, chen2021exploring, caron2021emerging}. However, strong augmentations like color perturbations are suboptimal for MIM~\cite{he2022masked}, as masking corruption itself introduces adequate regularization. Consequently, JEA-based SSL methods~\cite{DBLP:conf/iclr/YiGLYLWSQ23, li2023mage, huang2023contrastive, eymael2024efficient} keep simple augmentations like random cropping or flipping for the masked online view, and use either strong or simple ones for the full (unmasked) target view.

% FS-VFM stands out by eliminating all explicit augmentations on either branch and behaves well. This may stem from the preserved facial semantics in unaltered inputs, which aid in learning intact information~\cite{wang2023toward}, crucial for face security tasks where forgery and spoofing cues may be implicit anywhere. Moreover, our CRFR-P masking inherently induces sufficient spatial variations tailored to facial structures, obviating even simple (crop and flip) augmentations. Thus, FS-VFM only processes a single, original view per image, without compromising generalizability.

FS-VFM stands out by eliminating all explicit augmentations for both branches without compromising generalizability. This may stem from the preserved facial semantics in unaltered inputs, which aid in learning intact information~\cite{wang2023toward}, crucial for face security tasks where forgery and spoofing cues may be implicit anywhere. Moreover, our CRFR-P masking inherently induces sufficient spatial variations tailored to facial structures, obviating even simple (crop and flip) augmentations. Thus, FS-VFM only processes a single, original view per image.

\subsubsection{Loss Formulation}
% We evaluate two dominant loss types for the ID pretext task. For contrastive loss, which pulls positive pairs from the same sample closer while pushing negative pairs from different samples apart, we adopt the widely used InfoNCE~\cite{oord2018representation}. For non-contrastive loss, which solely maximizes the similarity between positives, we employ the mean squared error (MSE) from BYOL~\cite{grill2020bootstrap} and negative cosine similarity (NCS) from SimSiam~\cite{chen2021exploring} in an asymmetric form. Our empirical studies suggest that the NCS outperforms InfoNCE in FS-VFM, though most JEA-based (MIM\&ID) methods~\cite{li2023mage, DBLP:conf/iclr/YiGLYLWSQ23, huang2023contrastive, wang2023toward, wei2024towards, jamal2025multi} prefer the latter. We speculate that, during large-scale pre-training on real faces, the inter-sample contrast between negative pairs, which pushes real faces apart, may hinder our model to learn the intrinsic facial ``realness'' representations. Thus, FS-VFM adopts the asymmetric NCS (\Cref{eq:loss_id}) by default, matching each online anchor with its target view without negative pairs, to learn intra-face correspondence effectively and efficiently.

We compare two dominant loss types for the ID pretext task: for contrastive loss, which pulls positive pairs from the same sample closer while pushing negative pairs from different samples apart, we adopt the widely used InfoNCE~\cite{oord2018representation}; for non-contrastive loss, which solely maximizes the similarity between positives, we employ the mean squared error (MSE) from BYOL~\cite{grill2020bootstrap} and negative cosine similarity (NCS) from SimSiam~\cite{chen2021exploring} in an asymmetric form. We found that the NCS performs better for FS-VFM, although most JEA-based methods~\cite{li2023mage, DBLP:conf/iclr/YiGLYLWSQ23, huang2023contrastive, wang2023toward, wei2024towards, jamal2025multi} prefer the InfoNCE. We speculate that, for pre-training on real faces, the inter-sample contrast between negatives, which pushes real faces apart, may hinder our model to learn common facial “realness’’ representations. We thus adopt the asymmetric NCS (\Cref{eq:loss_id}) by default, matching each online anchor to its target view without negatives, to learn intra-face correspondence effectively and efficiently.

\section{Adaptations and FS-Adapter}\label{sec:fs_ada}

\subsection{Adaptations on Face Security Tasks}

% As discussed in~\Cref{sec:rel_fst}, most vision foundation models (VFMs) are pre-trained for natural data recognition~\cite{he2016deep, chollet2017xception, tan2019efficientnet, dosovitskiy2020image, caron2021emerging, radford2021learning, he2022masked} or generic face analysis~\cite{zheng2022general, wang2023toward, cai2023marlin}. And thus, full fine-tuning remains the dominant transfer strategy for face security tasks~\cite{cheng2024can, sun2025towards, li2025critical, long2024generalized, liu2024cfpl, li2025optimal}. While benefiting from scaling up VFMs, it updates the entire backbone per task, which incurs heavy compute and storage overhead. An alternative, linear probing, which only learns a task-specific linear head, though efficient, performs poorly and is rarely employed on face security, as it cannot pursue the fine-grained, nonlinearly separable~\cite{he2022masked} facial features.

As most vision foundation models (VFMs) are pre-trained for natural recognition~\cite{he2016deep, chollet2017xception, tan2019efficientnet, dosovitskiy2020image, caron2021emerging, radford2021learning} or facial analysis~\cite{zheng2022general, wang2023toward, cai2023marlin}, full fine-tuning remains the dominant transfer strategy for face security tasks~\cite{cheng2024can, sun2025towards, li2025critical, long2024generalized, li2025optimal}. While benefiting from scaling up VFMs, it updates the entire backbone per task, which incurs heavy compute and storage overhead. An alternative, linear probing, which only learns a task-specific linear head, though efficient, performs poorly and is rarely employed in face security, as it cannot leverage the nonlinearly separable, fine-grained features~\cite{he2022masked}.

% Stemming from the NLP community, parameter-efficient fine-tuning (PEFT) updates only part of the backbone or additional parameters while maintaining or improving performance as much as possible, and has been successfully adopted in the CV domain~\cite{xin2024parameter}. As a promising PEFT solution, the adapter~\cite{houlsby2019parameter} freezes the pre-trained parameters and appends small learnable bottleneck modules in the encoder. Most visual adaptation methods~\cite{houlsby2019parameter, chen2022adaptformer, sung2022vl, yin20255} integrate adapters into every transformer layers and learn them independently, as shown in~\Cref{fig:adaptions} (c). However, without face-security domain knowledge, simple fine-tuning multiple additional modules for naïve binary classification may still struggle with overfitting and limit generalization.

Parameter-efficient fine-tuning (PEFT) mitigates this trade-off by updating only part of the backbone or additional modules, which stems from the NLP and has been successfully adopted in the CV community~\cite{xin2024parameter}. A prominent PEFT strategy, the adapter~\cite{houlsby2019parameter} freezes the pre-trained weights and inserts lightweight bottlenecks into every transformer layer, as shown in~\Cref{fig:adaptions} (b), whose simplicity and effectiveness have been widely extended to visual adapter tuning~\cite{chen2022adaptformer, sung2022vl, yin20255}. However, without domain knowledge, tuning multiple modules on naïve binary classification may still suffer from overfitting and limit generalization. Thus, we explore: without modifying the backbone architecture, how to harness the generic real face representations of FS-VFM, through a plug-and-play adapter that is agnostic to specific forgery or spoofing types, substantially reduces trainable parameters while preserving generalization as much as possible, enabling ultra-efficient transfer to downstream face security tasks. To this end, we propose a bottleneck FS-Adapter, as illustrated in ~\Cref{fig:adaptions} (c).

% Thus, we explore: without modifying the backbone architecture, how to simply harness the generic real face representations of FS-VFM ViTs, through a plug-and-play adapter that is agnostic to specific forgery types or spoofing ways, substantially reduces trainable parameters while preserving generalization, thus enabling ultra-efficient transfer to downstream face security tasks and achieving performance on par with, or even superior to, that of fully fine-tuned other VFMs. To achieve these properties, we propose a bottleneck FS-Adapter, as illustrated in ~\Cref{fig:adaptions} (d). 

{
\setlength{\abovecaptionskip}{0pt}
\begin{figure}[tb!]
\centering
\includegraphics[width=.95\linewidth]{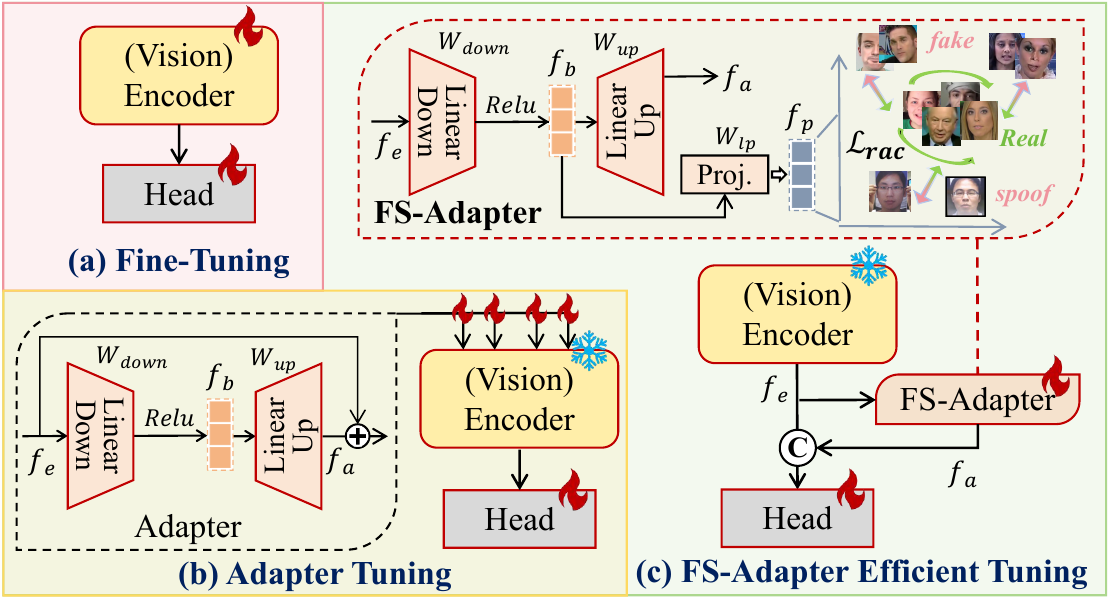}
% \caption{Adaptations for transferring vision foundation models to face security tasks. (a) Full fine-tuning updates the entire backbone and the task head. (b) Linear probing freezes the pre-trained encoder and learns only a linear classifier. (c) Adapter tuning keeps the backbone frozen and inserts trainable bottleneck adapters into Transformer layers. (d) Our FS-Adapter introduces a lightweight, plug-and-play bottleneck on top of the frozen encoder, delivering strong generalization with real-anchor contrastive learning ($\mathcal{L}_\mathit{rac}$), while achieving efficient tuning.}
\caption{Adaptation methods for transferring vision foundation models to face security tasks. (a) Simple fine-tuning updates the entire backbone and the task head. (b) Adapter tuning freezes the backbone and trains bottleneck adapters throughout Transformer layers. (c) Our FS-Adapter introduces a lightweight, plug-and-play bottleneck on top of the frozen encoder, delivering effective and efficient tuning with real-anchor contrastive learning ($\mathcal{L}_\mathit{rac}$).}
\label{fig:adaptions}
\end{figure}
}

\subsection{Effective and Efficient FS-Adapter Tuning}

% \noindent\textbf{Vanilla Adapters} A standard adapter~\cite{houlsby2019parameter} is a small bottleneck module composed of a linear down-sampling layer parameterized by $w_{\mathrm{down}}\!\in\!\mathbb{R}^{d \times b}$, a non-linear $\operatorname{ReLU}$ activation, and a linear up-sampling layer $w_{\mathrm{up}}\!\in\!\mathbb{R}^{b \times d}$. Here, $d$ and $b$ are the input and bottleneck middle dimensions, respectively, where $b\!\ll\!d$. For an input feature $f_{\mathrm{e}}\!\in\!\mathbb{R}^{N \times d}$ from the frozen encoder, the adapter produces the feature:
\noindent\textbf{Vanilla Adapter} is a small bottleneck module~\cite{houlsby2019parameter} consisted of a linear down-sampling layer parameterized by $w_{\mathrm{down}}\!\in\!\mathbb{R}^{d \times b}$, a non-linear $\operatorname{ReLU}$ activation, and a linear up-sampling layer $w_{\mathrm{up}}\!\in\!\mathbb{R}^{b \times d}$. Here, $d$ and $b$ are the input and bottleneck dimensions, where $b\!\ll\!d$. For an input feature $f_{\mathrm{e}}\!\in\!\mathbb{R}^{N \times d}$ from the frozen encoder, the adapter produces the feature:
\begin{equation}
f_{a}\!\in\!R^{N \times d}=w_{up} \cdot\left(\operatorname{ReLU}\left(w_{down} \cdot f_{e}\right)\right),
\end{equation}
which is then added back to $f_e$ by a scale factor and residual connection. With adapters in a standard $n$-layers ViT, the trainable parameters scale as $n\!\times\!2\!\times\!d\!\times b$.

\noindent\textbf{FS-Adapter} We extend the simple bottleneck of Adapters with a minimalist design to introduce the inductive bias for face security. As shown in~\Cref{fig:adaptions} (c), our FS-Adapter includes a novel \textit{Real-Anchor Contrastive Loss} $\mathcal{L}_\mathit{rac}$ that effectively leverages and constrains real face representations in the bottleneck space. It adds only a one-layer linear projector $w_{\mathit{lp}}\!\in\!\mathbb{R}^{b \times b}$ to map the bottleneck features $f_{\mathrm{b}}\!\in\!\mathbb{R}^{N \times b}$ into $f_{\mathrm{p}}\!\in\!\mathbb{R}^{N \times b}$:
\begin{equation}
f_{p}= w_{\mathit{lp}} \cdot f_{b} = w_{\mathit{lp}} \cdot \operatorname{ReLU}\left(w_{\text {down}} \cdot f_{e}\right).
\end{equation}
After normalization, the projection features $f_{p}$ are used to compute $\mathcal{L}_\mathit{rac}$. Meanwhile, the adapter features $f_{\mathrm{a}}\!\in\!\mathbb{R}^{N \times d}$ are fused with the original feature $f_e$ for the task head. We empirically find that concatenation yields better downstream performance than residual connections, despite adding negligible $\mathit{2}\!\times\!d$ parameters for the binary classifier. Crucially,  we attach the FS-Adapter solely after the last ViT block, which not only acquires task knowledge but also preserves the rich semantics and full expressivity of the frozen FS-VFM encoder.

% \noindent\textbf{Real-Anchor Contrastive Loss} In the context of downstream face security tasks, fake faces may derive from unknown digital or physical attacks, which also involve various forgery types (e.g., face swapping, attribute manipulation, and entire face synthesis) or spoofing manners (e.g., print attacks, video replay, and 3D masks). Thus, following the motivations of pre-training FS-VFM, we continue to focus on real faces for downstream adaptation rather than prior assumptions about specific forgeries or spoofs. To enhance both the discrimination and generalization, we introduce a Real-Anchor Contrastive loss that pulls together features of real faces, and pushes apart features between real and non-real faces, while not considering the distance between fake samples. 

\noindent\textbf{Real-Anchor Contrastive Loss} In real-world face security tasks, fake faces may derive from diverse unknown digital forgeries (\eg, face swapping and face synthesis) or physical attacks (\eg, print photo and replay video). Thus, following the motivations of pre-training FS-VFM, we center downstream adaptation on real faces rather than prior assumptions about specific forgeries or spoofs. To enhance both discrimination and generalization, we introduce the $\mathcal{L}_\mathit{rac}$: it pulls features of real faces together, and pushes apart real versus fake faces, while neglecting distances between fake samples.

Formally, let $\mathcal{R}$ denote the set of real faces in a batch and $\mathcal{F}$ denote the non-real. For each anchor from real faces, \ie $i\!\in\!\mathcal{R}$ with the projected bottleneck features $f_p^i$, we define its positive set (other real faces) as $\mathcal{P}(i) = \{ j\!\in\!\mathcal{R}, j\!\neq\!i\}$ and negative set (all non-real) as $\mathcal{N}(i) = \{k\!\in\!\mathcal{F}\}$. The $\mathcal{L}_\mathit{rac}$ is:
\begin{align}
\mathcal{L}_{\mathrm{rac}}
&= \frac{1}{|\mathcal{R}|}
   \sum_{i\in\mathcal{R}}
   \Bigl\{
     -\frac{1}{|\mathcal{P}(i)|}
      \sum_{j\in\mathcal{P}(i)}
      \log\\
&\quad
      \frac{\exp\bigl(f_p^i\!\cdot\!f_p^j/\tau\bigr)}
           {\sum_{j\in\mathcal{P}(i)}\exp\bigl(f_p^i\!\cdot\!f_p^j/\tau\bigr)
            +\sum_{k\in\mathcal{N}(i)}\exp\bigl(f_p^i\!\cdot\!f_p^k/\tau\bigr)}
   \Bigr\}\,,\nonumber
\end{align}
\noindent where $\tau$ is the temperature. By anchoring only on real faces, FS-Adapter leverages the relative stability and the suggested \textbf{\textit{3C}} of real faces to calibrate the feature space, which promotes tight clustering of real faces and better separation margins to non-real ones. Moreover, it leaves non-real faces unstructured to prevent overfitting in specific forgery or spoof patterns, improving generalization across diverse face security tasks.

% \noindent\textbf{Efficient Tuning} During downstream adaptations, only the FS-Adapter and the head are optimized from the overall loss:
% \begin{equation}
% \mathcal{L} = \mathcal{L}_{task} + \lambda_\mathit{rac} \mathcal{L}_\mathit{rac},
% \end{equation}
% where $\mathcal{L}_{task}$ is the task loss (\eg, cross-entropy) and $\lambda_\mathit{rac}$ is a weighting factor. Building upon FS-VFM, FS-Adapter can be appended only after the last transformer block, rather than throughout all layers. This is because the ViT pre-trained from FS-VFM has already encoded strong and transferable facial representations, largely alleviating the demand for early-layer intervention. Meanwhile, for calculating $\mathcal{L}_\mathit{rac}$, we apply an one-layer linear projector $w_{\mathrm{proj}} \in \mathbb{R}^{b \times b}$ to the bottleneck features instead of the input $f_{\mathrm{e}}$ or the output $f_{\mathrm{a}}$, which not only improves the discriminability by a compact mapping space, but also reduces the trainable parameters from $d \times d$ to $b \times b$ ($b \ll d$). Together, the number of trainable parameters of FS-Adapter is  $2 \times d \times b + b \times b$, and is around 1\% of the whole model, less than the 12\% of vanilla Adapters. Furthermore, this plug-and-play design on top of ViT only backpropagates gradients to the lightweight adapter. Thus, the FS-Adapter significantly reduces trainable parameters and computational overhead, yielding an ultra-efficient tuning for adaptation to downstream face security tasks.

\noindent\textbf{Efficient Tuning} During downstream adaptation, we optimize only the FS-Adapter and the task head with $\mathcal{L} = \mathcal{L}_{task} + \lambda_\mathit{rac} \mathcal{L}_\mathit{rac}$, where $\mathcal{L}_{task}$ is the task loss (\eg, cross-entropy), $\lambda_\mathit{rac}$ is a weighting factor. Building upon the strong, transferable facial representations from FS-VFM, FS-Adapter can be appended only after the last ViT block rather than throughout all layers. Meanwhile, we apply the linear projector $w_{\mathit{lp}}\!\in\!\mathbb{R}^{b\!\times\!b}$ to the bottleneck features instead of the input $f_{\mathrm{e}}$ or the output $f_{\mathrm{a}}$, which not only improves the discriminability by a compact mapping space but also reduces parameters from $d\!\times\!d$ to $b\!\times\!b$ ($b\!\ll\!d$). In total, FS-Adapter introduces only $2\!\times\!d\!\times\!b + b\!\times\!b$ trainable parameters, roughly $1/n$ of those required by vanilla adapters in an $n$-layer ViT. Further, it only backpropagates gradients to the lightweight adapter preceding the backbone. Thus, the FS-Adapter significantly reduces trainable parameters and computational overhead, enabling ultra-efficient adaptation to downstream face security tasks.

\section{Experiments}
\label{sec:exp}
We evaluate the effectiveness of learning and adapting FS-VFMs on three challenging face security tasks: cross-dataset deepfake detection (\texttt{DFD}, \cref{subsec:exp_deeepfake}), cross-domain face anti-spoofing (\texttt{FAS}, \cref{subsec:exp_fas}), and unseen diffusion face forensic (\texttt{DiFF}, \cref{subsec:exp_diff}) by thoroughly examining:

\textbf{\hypertarget{q1}{Q1}}~Do our facial representations transfer better than common model initialization practices? % cite table?

\textbf{\hypertarget{q2}{Q2}}~How do FS-VFMs compare to existing vision foundation models (VFMs)—both natural and facial—across supervised, self-supervised, and vision–language pre-training paradigms?

\textbf{\hypertarget{q3}{Q3}}~Further challenging \hyperlink{q2}{Q2}, with FS-VFMs frozen, can FS-Adapter efficient-tuning rival fully fine-tuning existing VFMs?

\textbf{\hypertarget{q4}{Q4}}~Can our pre-trained FS-VFM outperform SOTA task-specific methods just by simple fine-tuning its vanilla ViT? 

% \textbf{\hypertarget{q5}{Q5}}~While transferable and generalizable, are FS-VFMs scalable \wrt model and data sizes?
\textbf{\hypertarget{q5}{Q5}}~Are the gains consistent with scaling model/data up?

\noindent We also present ablation studies on FS-VFM (\cref{subsec:exp_ablation}) and FS-Apapter (\cref{subsec:abbla_fsa}), as well as visualizations (\cref{subsec:exp_vis}), to ascertain our contributions. More experimental details are provided in the supplementary material. %TODO

%-------------------------------------------------------------------------
\subsection{Pre-training Setups and Baselines}
% For our main experiments, we pre-train FS-VFMs on the VGGFace2~\cite{cao2018vggface2} dataset (VF2, $\sim$3M images). We use DLIB~\cite{king2009dlib} for face detection and cropping with a 30\% addition margin, and FACER~\cite{zheng2022general} toolkit for face parsing instead of alignment. We resize cropped face images to $224 \times 224$, with parsing maps saved as binary stream files for efficient CRFR-P masking.
\noindent\textbf{Data and Preprocessing} For main experiments, we pre-train FS-VFMs on VGGFace2 (VF2, $\sim$3M images)~\cite{cao2018vggface2} dataset. We use DLIB~\cite{king2009dlib} to detect and crop faces (with a 30\% margin), and FACER~\cite{zheng2022general} toolkit for face parsing instead of alignment. We resize cropped faces to $224\times224$, with parsing maps saved as binary streams for efficient CRFR-P masking.

\noindent\textbf{Architecture} In FS-VFMs, the MIM network is a naïve MAE~\cite{he2022masked} guided by our CRFR-P, with a vanilla ViT-\{S, B, L\}/16 as the encoder $E_o$. In the ID network, rep decoders $D_{o}^r$ and $D_{t}^r$ are 2-layer ViT blocks with the same width as the encoder, where the projector and predictor are 2-layer MLPs like BYOL~\cite{grill2020bootstrap}. After pre-training, we retain only $E_o$ as the backbone and append a task head for downstream adaptation.

\noindent\textbf{Implementation} We set the mask ratio $r$ to 0.75 and use \textbf{no} data augmentation during pre-training. We empirically set loss weights $\lambda_\mathit{fr}=0.007$ and $\lambda_\mathit{cl}=0.1$. The EMA momentum coefficient follows~\cite{grill2020bootstrap}. We pre-train our FS-VFMs from scratch for 600 epochs. Other setups follow MAE~\cite{he2022masked} defaults.

\noindent\textbf{Pre-trained Baseline VFMs} To probe \hyperlink{q1}{Q1}-\hyperlink{q3}{Q3}, we evaluate the following VFMs across mainstream pre-training paradigms and ViT sizes, chosen by availability (released weights), fairness (vanilla ViTs), and relevance (natural and facial domain):

$\bullet$ \textit{Scratch}~\cite{dosovitskiy2020image} \{S/16, B/16, L/16\}: random initialization, to discern pre‑training benefits versus backbone effects;

$\bullet$ \textit{Supervised}~\cite{dosovitskiy2020image} \{S/16, B/16, L/16\}: standard ImageNet supervised pre-training (Sup), the most common weight initialization for face security tasks;

% $\bullet$ \textit{MAE}~\cite{he2022masked} \{B/16, L/16\}: self-supervised learning via masked image modeling (SSL/MIM), our MIM network, the baseline across all experiments and ablations;
$\bullet$ \textit{MAE}~\cite{he2022masked}\{B/16, L/16\}: self-supervised masked image modeling (SSL/MIM), our MIM network \& ablative baseline.

$\bullet$ \textit{DINO}~\cite{caron2021emerging} \{S/16, B/16\}: self-supervised learning via instance discrimination (SSL/ID), a self-distillation method for learning local-to-global correspondence;

$\bullet$ \textit{CLIP}~\cite{radford2021learning} \{B/16, L/14\}: contrastive vision–language pre-training (VLP) on web-scale image–text pairs from LAION400M, which includes $\sim$50M facial images~\cite{zheng2022general};

$\bullet$ \textit{FaRL}~\cite{zheng2022general} \{B/16\}: joint CLIP with masked image modeling (VLP/JEA), pre-trained on 20M face–text pairs for weakly-supervised facial representation learning;

$\bullet$ \textit{MCF}~\cite{wang2023toward} \{B/16\}: self-supervised facial representation learning that also joint MIM and ID (SSL/JEA), pre-trained on 20M face images from FaRL.

% For downstream tasks, we compare FS-VFMs with these prevalent VFMs by initializing the backbones with their pre-trained weights, while keeping other settings identical, so performance differences stem solely from representation quality.
In downstream tasks, these prevalent VFMs, including ours, share identical settings except for the pre-trained weights, so performance essentially depends on the representation quality.

{
\setlength{\abovecaptionskip}{0pt}
\begin{table*}
\centering
\caption{Cross-dataset evaluation of simple fine-tuning VFMs on deepfake detection (\textbf{\texttt{DFD}}). All models are fine-tuned on FF++ (c23) and tested on unseen datasets under identical settings. \textit{\&FS-Adapter ET (Efficient Tuning)} only updates the FS-Adapter and head, freezing the ViT backbone. Left: frame-level, Right: video-level. \textbf{Best results}, \underline{second-best}.}
\label{tab:dfd_vfm}
\begingroup
\renewcommand{\arraystretch}{1.1}
\resizebox{.925\linewidth}{!}{%
\begin{tabular}{lcccccccccccccccccc} \noalign{\hrule height 1pt} 
\multirow{2}{*}{Method} & \multirow{2}{*}{Backbone} & \multicolumn{2}{c}{Pre-train} & \multirow{2}{*}{\begin{tabular}[c]{@{}c@{}}Train.\\Param.\end{tabular}} & \multirow{2}{*}{\begin{tabular}[c]{@{}c@{}}Train\\Set\end{tabular}} & \multicolumn{5}{c}{Test Set \textbf{Frame-level AUC↑ (\%)}} & \multirow{2}{*}{\textbf{Avg.}} &  & \multicolumn{5}{c}{Test Set \textbf{Video-level AUC↑ (\%)}} & \multirow{2}{*}{\textbf{Avg.}} \\ \cline{3-4}\cline{7-11}\cline{14-18}
 &  & Manner & Type &  &  & CDFV2 & DFDCP & DFDC & WDF & CDF++ &  &  & CDFV2 & DFDCP & DFDC & WDF & CDF++ &  \\ \noalign{\hrule height 1pt} 
Xception~\cite{chollet2017xception} & CNN & Sup & Natural\textsuperscript{IN} & 20.9M & FF++ & 69.52 & 68.94 & 68.20 & 68.83 & 73.70 & 69.84 &  & 76.39 & 72.24 & 70.62 & 76.11 & 79.10 & 74.89 \\
EfficientNet-B4~\cite{tan2019efficientnet} & CNN & Sup & Natural\textsuperscript{IN} & 17.6M & FF++ & 73.37 & 64.37 & 69.47 & 71.95 & 70.60 & 69.95 &  & 79.81 & 66.95 & 71.85 & 76.42 & 73.80 & 73.77 \\ \hline
Scratch~\cite{dosovitskiy2020image} & ViT-S/16 & None & Rand.Init. & 21.6M & FF++ & 62.46 & 68.91 & 64.01 & 59.38 & 65.41 & 64.03 &  & 64.82 & 72.89 & 66.82 & 62.17 & 67.77 & 66.89 \\
Supervised~\cite{dosovitskiy2020image} & ViT-S/16 & Sup & Natural\textsuperscript{IN} & 21.6M & FF++ & 65.67 & 70.76 & 60.53 & 68.57 & 70.23 & 67.15 &  & 73.04 & 76.58 & 58.23 & 70.42 & 74.53 & 70.56 \\
DINO~\cite{caron2021emerging} & ViT-S/16 & SSL/ID & Natural\textsuperscript{IN} & 21.6M & FF++ & 69.88 & 72.86 & 70.31 & 72.48 & 66.52 & 70.41 &  & 74.74 & 76.70 & 72.79 & 79.89 & 70.35 & 74.89 \\
\rowcolor[rgb]{1,0.965,0.8} FS-VFM (Ours) & ViT-S/16 & SSL/JEA & Facial\textsuperscript{3M} & 21.6M & FF++ & \textbf{83.15} & \textbf{82.60} & \textbf{76.94} & \textbf{81.29} & \textbf{81.06} & \textbf{81.01} &  & \textbf{90.78} & \textbf{89.41} & \textbf{80.78} & \textbf{84.45} & \textbf{85.71} & \textbf{86.23} \\ \hdashline[1pt/1pt]
\rowcolor[rgb]{1,0.965,0.8}\textit{\&FS-Adapter ET}& ViT-S/16 & SSL/JEA & Facial\textsuperscript{3M} & \textbf{0.085M} & FF++ & \uline{70.73} & \uline{73.64} & \uline{71.02} & \uline{72.99} & \uline{75.71} & \uline{72.82} &  & \uline{75.40} & \uline{77.54} & \uline{73.17} & \uline{75.62} & \uline{79.48} & \uline{76.24} \\ \hline
Scratch~\cite{dosovitskiy2020image} & ViT-B/16 & None & Rand.Init. & 85.6M & FF++ & 61.14 & 69.00 & 64.27 & 60.68 & 64.67 & 63.95 &  & 64.08 & 72.62 & 66.73 & 60.36 & 67.42 & 66.24 \\
Supervised~\cite{dosovitskiy2020image} & ViT-B/16 & Sup & Natural\textsuperscript{IN} & 85.6M & FF++ & 77.43 & 74.07 & 71.09 & 75.86 & 72.52 & 74.19 &  & 86.24 & 82.11 & 74.48 & 81.20 & 77.20 & 80.25 \\
CLIP~\cite{radford2021learning} & ViT-B/16 & VLP & Natural\textsuperscript{LA} & 85.8M & FF++ & 73.47 & 78.40 & 71.88 & 75.78 & 72.75 & 74.46 &  & 82.03 & 85.26 & 75.36 & 82.19 & 78.08 & 80.58 \\
MAE~\cite{he2022masked} & ViT-B/16 & SSL/MIM & Natural\textsuperscript{IN} & 85.6M & FF++ & 72.64 & 79.81 & 72.18 & 73.94 & 71.61 & 74.04 &  & 79.51 & 87.10 & 75.93 & 80.96 & 75.47 & 79.79 \\
DINO~\cite{caron2021emerging}  & ViT-B/16 & SSL/ID & Natural\textsuperscript{IN} & 85.6M & FF++ & 73.88 & 77.31 & 72.78 & 75.08 & 68.51 & 73.51 &  & 80.47 & 84.64 & 76.90 & 82.06 & 72.39 & 79.29 \\
FaRL~\cite{zheng2022general} & ViT-B/16 & VLP/JEA & Facial\textsuperscript{20M} & 85.8M & FF++ & 73.13 & 76.56 & 73.90 & 76.61 & 71.04 & 74.25 &  & 80.13 & 81.38 & 77.75 & 83.47 & 75.73 & 79.69 \\
MCF~\cite{wang2023toward} & ViT-B/16 & SSL/JEA & Facial\textsuperscript{20M} & 85.6M & FF++ & 73.16 & 75.78 & 69.63 & 74.10 & 71.59 & 72.85 &  & 80.25 & 82.55 & 73.61 & 79.79 & 76.26 & 78.49 \\
\textit{FSFM}~\cite{wang2025fsfm} $^{\textit{(Pre)}}$ & \textit{ViT-B/16} & \textit{SSL/JEA} & \textit{Facial\textsuperscript{3M}} & \textit{85.6M} & \textit{FF++} & \textit{85.05} & \textit{85.50} & \textit{80.20} & \textit{85.26} & \textit{81.29} & \textit{83.46} &  & \textit{91.44} & \textit{89.71} & \textit{83.47} & \textit{86.96} & \textit{85.76} & \textit{87.47} \\
\rowcolor[rgb]{1,0.965,0.8} FS-VFM (Ours) & ViT-B/16 & SSL/JEA & Facial\textsuperscript{3M} & 85.6M & FF++ & \textbf{86.13} & \textbf{88.87} & \textbf{81.84} & \textbf{85.34} & \textbf{84.27} & \textbf{85.29} &  & \textbf{93.03} & \textbf{93.11} & \textbf{85.08} & \textbf{88.20} & \textbf{88.74} & \textbf{89.63} \\ \hdashline[1pt/1pt]
\rowcolor[rgb]{1,0.965,0.8}\textit{\&FS-Adapter ET}& ViT-B/16 & SSL/JEA & Facial\textsuperscript{3M} & \textbf{0.335M} & FF++ & \uline{77.63} & \uline{85.06} & \uline{76.61} & \uline{84.11} & \uline{79.99} & \uline{80.68} &  & \uline{83.40} & \uline{88.45} & \uline{78.73} & \uline{85.96} & \uline{84.19} & \uline{84.14} \\ \hline
Scratch~\cite{dosovitskiy2020image} & ViT-L/16 & None & Rand.Init. & 303.1M & FF++ & 61.41 & 66.06 & 63.82 & 59.28 & 63.31 & 62.78 &  & 64.09 & 69.99 & 66.65 & 60.74 & 65.98 & 65.49 \\
Supervised~\cite{dosovitskiy2020image} & ViT-L/16 & Sup & Natural\textsuperscript{IN} & 303.1M & FF++ & 79.80 & 78.80 & 71.99 & 74.11 & 71.44 & 75.23 &  & 86.12 & \uline{85.62} & 75.43 & 81.32 & 75.08 & 80.71 \\
CLIP~\cite{radford2021learning} & ViT-L/14 & VLP & Natural\textsuperscript{LA} & 303.2M & FF++ & 73.35 & 77.54 & 73.17 & 77.81 & 67.96 & 73.97 &  & 83.32 & 81.27 & 76.46 & 83.44 & 73.83 & 79.66 \\
MAE~\cite{he2022masked} & ViT-L/16 & SSL/MIM & Natural\textsuperscript{IN} & 303.1M & FF++ & 74.25 & 81.53 & 75.14 & 78.99 & 70.65 & 76.11 &  & 80.69 & 88.63 & 79.71 & 83.57 & 74.32 & 81.38 \\
\rowcolor[rgb]{1,0.965,0.8} FS-VFM (Ours) & ViT-L/16 & SSL/JEA & Facial\textsuperscript{3M} & 303.1M & FF++ & \textbf{87.64} & \textbf{88.27} & \textbf{83.57} & \textbf{90.34} & \uline{86.38} & \textbf{87.24} &  & \textbf{95.15} & \textbf{93.35} & \textbf{87.74} & \textbf{91.60} & \textbf{91.07} & \textbf{91.78} \\ \hdashline[1pt/1pt]
\rowcolor[rgb]{1,0.965,0.8}\textit{\&FS-Adapter ET}& ViT-L/16 & SSL/JEA & Facial\textsuperscript{3M} & \textbf{0.594M} & FF++ & \uline{84.31} & \uline{83.27} & \uline{80.34} & \uline{85.54} & \textbf{86.80} & \uline{84.05} &  & \uline{89.07} & \uline{85.62} & \uline{82.62} & \uline{85.10} & \uline{89.79} & \uline{86.44} \\ \noalign{\hrule height 1pt} 
\multicolumn{19}{l}{\begin{tabular}[c]{@{}l@{}}\small \textbf{\textit{Abbreviation:}}~ Sup(Supervised)~ SSL(Self-Supervised-Learning)~ VLP(Vision-Language-Pretraining)~ MIM(Masked-Image-Modeling)~ ID(Instance-Discrimination)\\ \small JEA(Joint-Embedding-Architecture)~ IN(ImageNet)~ LA(Laion)~ Train.Param.(Trainable Parameters)\end{tabular}}
\end{tabular}
}
\endgroup
\end{table*}
}

{
\setlength{\abovecaptionskip}{0pt}
\begin{table*}
\centering
\caption{Cross-dataset evaluation on deepfake detection (\textbf{\texttt{DFD}}). For a fair comparison, results of SOTA task-specialized methods are cited from their original papers, and the results of CDF++ are from its benchmark. Avg.$\Delta$Ours denotes the average AUC improvement of FS-VFM (Ours) over other methods across their tested sets. Left: frame-level, Right: video-level. \textbf{Best results}, \underline{second-best}.}
\label{tab:dfd_sota}
\begingroup
\setlength{\tabcolsep}{2pt} % default 6pt
\renewcommand{\arraystretch}{1.15}
\resizebox{.95\linewidth}{!}{%
\begin{tabular}{lcccccccclcccccccc} \noalign{\hrule height 1pt}
\multirow{2}{*}{Method} & \multirow{2}{*}{\begin{tabular}[c]{@{}c@{}}Pre-train\\Manner/Type\end{tabular}} & \multirow{2}{*}{\begin{tabular}[c]{@{}c@{}}Train\\Set\end{tabular}} & \multicolumn{5}{c}{Test Set \textbf{Frame-level AUC↑ (\%)}} & \multirow{2}{*}{\begin{tabular}[c]{@{}c@{}}\textbf{Avg.}\\\textbf{$\Delta$Ours}\end{tabular}} & \multirow{2}{*}{Method} & \multirow{2}{*}{\begin{tabular}[c]{@{}c@{}}Pre-train\\Manner/Type\end{tabular}} & \multirow{2}{*}{\begin{tabular}[c]{@{}c@{}}Train\\Set\end{tabular}} & \multicolumn{5}{c}{Test Set \textbf{Video-level AUC↑ (\%)}} & \multirow{2}{*}{\begin{tabular}[c]{@{}c@{}}\textbf{Avg.}\\\textbf{$\Delta$Ours}\end{tabular}} \\ \cline{4-8}\cline{13-17} &  &  & CDFV2 & DFDCP & DFDC & WDF & CDF++ &  &  &  &  & CDFV2 & DFDCP & DFDC & WDF & CDF++ & \\ \noalign{\hrule height 1pt}
\multicolumn{8}{l}{\textit{\small SOTA \texttt{DFD}-specialized method (Venue)}} &  & \multicolumn{9}{l}{\small \textit{SOTA \texttt{DFD}-specialized method (Venue)}} \\ 
OST~\cite{chen2022ost} (NIPS'22)† & Sup\textsuperscript{IN}/Natural & FF++ & 74.80 & 83.30 &  &  &  & {\cellcolor[rgb]{1,0.965,0.8}}8.91↑ & SBIs~\cite{shiohara2022detecting} (CVPR'22)‡ & Sup\textsuperscript{IN}/Natural & FF++\textsuperscript{SD} & 93.18 & 86.15 & 72.42 &  & 73.40 & {\cellcolor[rgb]{1,0.965,0.8}}10.54↑ \\
RECCE~\cite{cao2022end} (CVPR'22)† & Sup\textsuperscript{IN}/Natural & FF++ & 68.71 &  & 69.06 & 64.31 & \uline{75.50} & {\cellcolor[rgb]{1,0.965,0.8}}17.59↑ & RealForensics~\cite{haliassos2022leveraging} (CVPR'22) & SSL\textsuperscript{ID}/Facial & FF++ & 86.90 &  & 75.90 &  &  & {\cellcolor[rgb]{1,0.965,0.8}}10.05↑ \\
UIA-ViT~\cite{zhuang2022uia} (ECCV'22)* & SSL/Facial & FF++ & 82.41 & 75.80 &  &  &  & {\cellcolor[rgb]{1,0.965,0.8}}8.85↑ & TALL-Swin\cite{xu2023tall} (ICCV'23)* & Sup\textsuperscript{IN}/Natural & FF++\textsuperscript{SD} & 90.79 &  & 76.78 &  &  & {\cellcolor[rgb]{1,0.965,0.8}}7.66↑ \\
CC-Net~\cite{zhu2023face} (TPAMI'23)† & Sup\textsuperscript{IN}/Natural & FF++ & 72.04 &  & 72,35 & &  & {\cellcolor[rgb]{1,0.965,0.8}}13.16↑ & AUNet~\cite{bai2023aunet} (CVPR'23)† & Sup\textsuperscript{IN}/Natural & FF++\textsuperscript{SD} & 92.77 & 86.16 & 73.82 &  &  & {\cellcolor[rgb]{1,0.965,0.8}}7.83↑ \\
UCF~\cite{yan2023ucf} (ICCV'23)† & Sup\textsuperscript{IN}/Natural & FF++ & 82.40 & 80.50 &  &  & 72.40 & {\cellcolor[rgb]{1,0.965,0.8}}9.00↑ & MLR~\cite{hong2024contrastive} (CVPR'24)* & Sup\textsuperscript{IN}/Natural & FF++ & 91.56 &  & 75.17 & 73.41 &  & {\cellcolor[rgb]{1,0.965,0.8}}11.45↑ \\
SFDG~\cite{wang2023dynamic} (CVPR'23)‡ & Sup\textsuperscript{IN}/Natural & FF++ & 75.83 &  & 73.64 & 69.27 &  & {\cellcolor[rgb]{1,0.965,0.8}}14.27↑ & NACO~\cite{zhang2024learning} (ECCV'24)* & SSL\textsuperscript{JEA}/Facial & FF++ & 89.50 &  & 76.70 &  &  & {\cellcolor[rgb]{1,0.965,0.8}}8.35↑ \\
IID~\cite{huang2023implicit} (CVPR'23) & Sup\textsuperscript{IN}/Natural & FF++ & 83.80 & 81.23 &  &  & 71.40 & {\cellcolor[rgb]{1,0.965,0.8}}8.62↑ & FPG~\cite{xia2024advancing} (MM'24)‡ & Sup\textsuperscript{IN}/Natural & FF++\textsuperscript{SD} & 94.49 & 87.24 & 74.75 &  &  & {\cellcolor[rgb]{1,0.965,0.8}}6.59↑ \\
CFM~\cite{luo2023beyond} (TIFS'24)‡ & Sup\textsuperscript{IN}/Natural & FF++ & 82.78 & 75.82 &  & 78.39 & 73.30 & {\cellcolor[rgb]{1,0.965,0.8}}10.59↑ & CFM~\cite{luo2023beyond} (TIFS'24)‡ & Sup\textsuperscript{IN}/Natural & FF++ & 89.65 & 80.22 &  & 82.27 & \uline{76.50} & {\cellcolor[rgb]{1,0.965,0.8}}10.63↑ \\
LSDA~\cite{yan2024transcending} (CVPR'24)‡ & Sup\textsuperscript{IN}/Natural & FF++\textsuperscript{SD} & 83.00 & 81.50 & 73.60 &  & 70.00 & {\cellcolor[rgb]{1,0.965,0.8}}9.44↑ & LSDA~\cite{yan2024transcending} (CVPR'24)‡ & Sup\textsuperscript{IN}/Natural & FF++\textsuperscript{SD} & 91.10 &  & 77.00 &  & 72.70 & {\cellcolor[rgb]{1,0.965,0.8}}11.05↑ \\
ProDet~\cite{cheng2024can} (NIPS'24)‡ & Sup\textsuperscript{IN}/Natural & FF++\textsuperscript{SD} & 84.48 & 81.16 & 72.40 & 77.18 & 69.20 & {\cellcolor[rgb]{1,0.965,0.8}}10.36↑ & ProDet~\cite{cheng2024can} (NIPS'24)‡ & Sup\textsuperscript{IN}/Natural & FF++\textsuperscript{SD} & 92.50 &  & 77.00 & 82.87 & 73.60 & {\cellcolor[rgb]{1,0.965,0.8}}9.90↑ \\
DiffFake~\cite{chen2024diffusionfake} (NIPS'24)* & Sup\textsuperscript{IN}/Natural & FF++ & 80.46 & 80.95 &  & 80.14 &  & {\cellcolor[rgb]{1,0.965,0.8}}8.23↑ & VB~\cite{yan2025generalizing} (CVPR'25)* & VLP\textsuperscript{CLIP}/Natural & FF++\textsuperscript{SD} & 94.70 & 90.90 & \uline{84.30} & 84.80 &  & {\cellcolor[rgb]{1,0.965,0.8}}3.29↑ \\
UDD~\cite{fu2025exploring} (AAAI'25)* & VLP\textsuperscript{CLIP}/Natural & FF++ & \uline{86.90} & \uline{85.60} & \uline{75.80} &  &  & {\cellcolor[rgb]{1,0.965,0.8}}3.73↑ & UDD~\cite{fu2025exploring} (AAAI'25)* & VLP\textsuperscript{CLIP}/Natural & FF++ & 93.10 & 88.10 & 81.20 &  &  & {\cellcolor[rgb]{1,0.965,0.8}}4.61↑ \\
FakeDiffer~\cite{wang2025fakediffer} (AAAI'25)† & Sup\textsuperscript{IN}/Natural & FF++ & 69.24 &  & 68.46 &  &  & {\cellcolor[rgb]{1,0.965,0.8}}16.76↑ & FCGA~\cite{han2025towards} (CVPR'25)* & VLP\textsuperscript{CLIP}/Natural & FF++ & \uline{95.00} &  & 81.80 & \uline{87.20} &  & {\cellcolor[rgb]{1,0.965,0.8}}3.50↑ \\
EDF~\cite{li2025critical} (AAAI'25)‡ & Sup\textsuperscript{IN}/Natural & FF++ & 76.30 &  & 70.27 & 69.29 &  & {\cellcolor[rgb]{1,0.965,0.8}}15.23↑ & KFD~\cite{yu2025unlocking} (ICML'25) & LVLM/Hybrid & FF++\textsuperscript{SD} & 94.71 & \uline{91.81} & 79.12 &  &  & {\cellcolor[rgb]{1,0.965,0.8}}3.53↑ \\
VLFFD~\cite{sun2025towards} (CVPR'25)* & VLP\textsuperscript{CLIP}/Natural & FF++\textsuperscript{SD} & 83.15 & 83.21 &  & \uline{85.10} &  & {\cellcolor[rgb]{1,0.965,0.8}}4.93↑ & FakeSTormer~\cite{nguyen2025vulnerability} (ICCV'25)* & SSL\textsuperscript{MAE}/Natural & FF++\textsuperscript{SD} & 92.40 & 90.00 & 74.60 & 74.20 &  & {\cellcolor[rgb]{1,0.965,0.8}}9.16↑ \\ \hline
\multicolumn{9}{l}{\textit{\small Simple Fine-Tuning w/o task-specific methodology}} & \multicolumn{9}{l}{\textit{\small Simple Fine-Tuning w/o task-specific methodology}} \\ 
\rowcolor[rgb]{1,0.965,0.8} FS-VFM (Ours)* & SSL\textsuperscript{JEA}/Facial & FF++ & \textbf{87.64} & \textbf{88.27} & \textbf{83.57} & \textbf{90.34} & \textbf{86.38} & $\Delta$ & FS-VFM (Ours)* & SSL\textsuperscript{JEA}/Facial & FF++ & \textbf{95.15} & \textbf{93.35} & \textbf{87.74} & \textbf{91.60} & \textbf{91.07} & $\Delta$ \\ \noalign{\hrule height 1pt}
\multicolumn{18}{l}{\begin{tabular}[c]{@{}l@{}}\small \textbf{\textbf{\textit{Abbreviation:}}}~ Sup(Supervised)~ IN(ImageNet)~ SSL(Self-Supervised-Learning)~ VLP(Vision-Language-Pretraining)~ ID(Instance-Discrimination)~ JEA(Joint-Embedding-Architecture)\\\small LVLM(Large-Vision-Language-Model)~ \textbf{\textit{FF++}}\textit{\textbf{\textbf{\textsuperscript{SD}}:}} Synthetic (or Self-made) Data from FF++~~~~\textit{\textbf{Backbone (or modified):}} Xception~†~~~~EfficientNet-B4~‡~~~~ViTs~*\end{tabular}}
\end{tabular}
}
\endgroup
\end{table*}
}

%-------------------------------------------------------------------------
\subsection{Cross-Dataset Deepfake Detection}
\label{subsec:exp_deeepfake}

\noindent\textbf{Setting} To evaluate the generalizability of our method across diverse \texttt{DFD} scenarios, we follow the challenging cross-dataset evaluation. Specifically, we train \textit{one} detector on the FaceForensics++ (FF++, c23/HQ)~\cite{rossler2019faceforensics++} dataset and test it on unseen datasets: CelebDF-v2 (CDFv2)~\cite{li2020celeb}, Deepfake Detection Challenge preview (DFDCp)~\cite{dolhansky2019dee}, Deepfake Detection Challenge (DFDC)~\cite{dolhansky2020deepfake}, Wild Deepfake (WDF)~\cite{zi2020wilddeepfake}, and CelebDF++~\cite{li2025celeb}. For simple fine-tuning the vanilla ViT from baseline VFMs and our FS-VFMs, we add only \textit{one} linear layer as the binary classifier after averaging all non-[CLS] token features. We also append the FS-Adapter to FS-VFMs and further freeze the ViT backbone for parameter-efficient tuning. We report both the frame-level and video-level Area Under Curve (AUC), the most widely used metric for \texttt{DFD}.

\noindent\textbf{Comparison with existing VFMs}~\Cref{tab:dfd_vfm} shows that FS-VFM (Ours) transcends all natural and facial VFMs by a substantial margin at both frame and video levels, boosting generalization across all unseen deepfakes and ViT scales. 1) FS-VFMs outperform ImageNet supervised Xception, EfficientNet-B4, and ViTs, which are common practices in \texttt{DFD}, suggesting that FS-VFMs provide a much stronger initialization for deepfake detectors. 2) The ViT-B/16 pre-trained by MIM-based MAE and ID-based DINO show comparable performance but differ across datasets, given MIM targets local patterns while ID operates globally~\cite{zhu2023understanding}. FS-VFM surpasses both, confirming the efficacy of learning both local and global representations. 3) FS-VFM also surpasses the VLP-based CLIP, which benefits from web-scale image-text pairs and has emerged as a strong \texttt{DFD} model~\cite{sun2025towards, fu2025exploring, yan2025generalizing, han2025towards}. 4) As for facial VFMs, the FaRL integrates VLP with MIM, but underperforms the original CLIP. The MCF is also an MIM and ID joint SSL, yet generalizes even worse than most natural VFMs, stressing the gap between face analysis and security tasks. Notably, our FS-VFM, despite being pre-trained on only 3M faces, distinctly exceeds both FaRL and MCF, which are pre-trained on 20M faces. 5) Even the FS-VFM ViT-S/16 outperforms other VFMs built on larger ViT-B and ViT-L, highlighting our superior pre-training quality. 6) In summary, FS-VFMs set a new bar for generalizable \texttt{DFD} by simple fine-tuning of the vanilla ViT, demonstrating that our methods effectively learns fundamental real face representations that are sensitive to deepfakes.

% \noindent\textbf{Comparison by efficient FS-Adapter tuning} With the FS-VFM ViT backbone frozen, \textit{\&FS-Adapter Efficient Tuning} updates merely 0.39\%, 0.39\%, and 0.20\% of parameters for ViT-S/16, ViT-B/16, and ViT-L/16, respectively, yet still achieves superior generalization compared to full fine-tuning other VFMs, as shown in~\Cref{tab:dfd_vfm}. Specifically: 1) Across ViT-S/B/L, \&FS-Adapter Efficient Tuning is the runner-up to fully fine-tuned FS-VFMs on both frame and video level AUCs, indicating that the adapter preserves most of the pre-training gains while drastically reducing trainable parameters. 2) Using only ViT-B/16, \&FS-Adapter already attains higher average AUCs than fully fine-tuned counterparts of other VFMs, underscoring the competitiveness of parameter-efficient tuning when coupled with strong facial representations from FS-VFMs. 3) Scaling \&FS-Adapter to ViT-L/16 further improves the performance while optimizing a smaller fraction of the model (0.20\%). Notably, it approaches the full fine-tuning results of FS-VFM ViT-B/16 and is even comparable to our previous FSFM~\cite{wang2025fsfm} ViT-B/16, while cutting trainable parameters by $>144\times(85.6M/0.594M)$. 4) Taken together, the lightweight, plug-and-play FS-Adapter retains most generalizability of FS-VFMs, scales favorably with larger ViTs, and thus delivers a highly cost-effective path for real-world deployment scenarios under computational resource constraints.

\noindent\textbf{Comparison by FS-Adapter Efficient Tuning} As shown in~\Cref{tab:dfd_vfm}, with the FS-VFM ViT backbone frozen, \textit{\&FS-Adapter ET (Efficient Tuning)} updates merely 0.394\%, 0.391\%, and 0.196\% parameters of ViT-S/16, ViT-B/16, and ViT-L/16, respectively, yet still generalizes better than full fine-tuning other VFMs. 1) Across ViT-S/B/L, \&FS-Adapter ET keeps the runner-up only to fully fine-tuned FS-VFMs, indicating that most of the pre-training gains are preserved while drastically reducing trainable parameters. 2) Using only ViT-B/16, \&FS-Adapter already yields a higher average AUC than fully fine-tuning other VFMs, which underscores the potential of coupling strong facial representations from FS-VFMs with parameter-efficient tuning. 3) Scaling \&FS-Adapter to ViT-L/16 further improves the performance while optimizing a smaller fraction (0.196\%) of the model. Notably, it approaches the full fine-tuning results of FS-VFM ViT-B/16 and is even comparable to \textit{our previous FSFM}~\cite{wang2025fsfm} ViT-B/16, while cutting trainable parameters $>\!144\times$ (85.6M$\rightarrow$0.594M). 4) Taken together, the lightweight, plug-and-play FS-Adapter retains most generalizability of FS-VFMs, scales better with larger ViTs, and thus delivers a highly cost-effective path for real-world deployment scenarios under computational constraints.

{
\setlength{\abovecaptionskip}{0pt}
\begin{table}
\centering
\caption{Cross-domain evaluation of simple fine-tuning VFMs on face anti-spoofing (\textbf{\texttt{FAS}}). All models are fine-tuned under identical settings. \textit{\&FS-Adapter Efficient Tuning} only updates the FS-Adapter and head, freezing the ViT backbone. \textbf{Best}, \underline{second-best}.}
\label{tab:fas_vfm}
\begingroup
\setlength{\tabcolsep}{1pt}
\renewcommand{\arraystretch}{1.1}
\resizebox{\linewidth}{!}{%
\begin{tabular}{l c c c c r@{~/~}l r@{~/~}l r@{~/~}l r@{~/~}l r@{~/~}l} \noalign{\hrule height 1pt}
\multirow{2}{*}{Method} & \multirow{2}{*}{Backbone} & \multicolumn{2}{c}{Pre-train} & \multirow{2}{*}{\begin{tabular}[c]{@{}c@{}}Train.\\Param.\end{tabular}} & \multicolumn{2}{c}{OCI→\textbf{M}} & \multicolumn{2}{c}{OMI→\textbf{C}} & \multicolumn{2}{c}{OCM→\textbf{I}} & \multicolumn{2}{c}{ICM→\textbf{O}} & \multicolumn{2}{c}{\textbf{Avg.}} \\ \cline{3-4}\cline{6-15}
 &  & Manner & Type &  & \scriptsize HTER & \scriptsize AUC & \scriptsize HTER & \scriptsize AUC & \scriptsize HTER & \scriptsize AUC & \scriptsize HTER & \scriptsize AUC & \scriptsize HTER↓ & \scriptsize AUC↑ \\ \noalign{\hrule height 1pt}
Scratch~\cite{dosovitskiy2020image} & ViT-S/16 & None & Rand.Init. & 21.9M & 13.61 & 91.29 & 38.57 & 63.95 & 15.03 & 91.59 & 28.35 & 75.41 & 23.89 & 80.56 \\
Sup.~\cite{dosovitskiy2020image} & ViT-S/16 & Sup & Natural\textsuperscript{IN} & 21.9M & \uline{4.47} & \uline{98.14} & \uline{6.86} & \uline{97.99} & 11.79 & 95.36 & 13.88 & 92.45 & \uline{9.25} & \uline{95.99} \\
DINO~\cite{caron2021emerging} & ViT-S/16 & SSL/ID & Natural\textsuperscript{IN} & 21.9M & 16.14 & 90.91 & 30.08 & 76.43 & 15.97 & 92.69 & 28.04 & 76.05 & 22.56 & 84.02 \\
\rowcolor[rgb]{1,0.965,0.8} FS-VFM & ViT-S/16 & SSL/JEA & Facial\textsuperscript{3M} & 21.9M & \textbf{4.00} & \textbf{99.26} & \textbf{5.98} & \textbf{98.19} & \textbf{3.27} & \textbf{99.38} & \textbf{9.93} & \textbf{94.24} & \textbf{5.79} & \textbf{97.77} \\ \hdashline[1pt/1pt]
\rowcolor[rgb]{1,0.965,0.8}\textit{\&FS-Adapter} & ViT-S/16 & SSL/JEA & Facial\textsuperscript{3M} & \textbf{0.47M} & 8.94 & 96.96 & 11.36 & 95.35 & \uline{10.20} & \uline{96.13} & 14.71 & \uline{92.68} & 11.30 & 95.28 \\ \hline
Scratch~\cite{dosovitskiy2020image} & ViT-B/16 & None & Rand.Init. & 86.2M & 15.37 & 90.73 & 35.37 & 68.23 & 14.75 & 94.18 & 31.65 & 71.55 & 24.29 & 81.17 \\
Sup.~\cite{dosovitskiy2020image} & ViT-B/16 & Sup & Natural\textsuperscript{IN} & 86.2M & \textbf{3.52} & 98.74 & \uline{2.42} & \uline{99.52} & 8.45 & 96.91 & 11.86 & 94.62 & \uline{6.56} & \uline{97.45} \\
CLIP~\cite{radford2021learning} & ViT-B/16 & VLP & Natural\textsuperscript{LA} & 86.2M & 6.00 & 98.66 & \uline{2.42} & 99.43 & 13.37 & 94.02 & \uline{8.04} & \uline{97.42} & 7.46 & 97.38 \\
MAE~\cite{he2022masked} & ViT-B/16 & SSL/MIM & Natural\textsuperscript{IN} & 86.2M & 10.32 & 94.87 & 15.91 & 89.96 & 15.54 & 91.13 & 16.51 & 90.29 & 14.57 & 91.56 \\
DINO~\cite{caron2021emerging} & ViT-B/16 & SSL/ID & Natural\textsuperscript{IN} & 86.2M & 6.73 & 97.15 & 13.44 & 93.90 & 14.27 & 93.56 & 15.55 & 90.99 & 12.50 & 93.90 \\
FaRL~\cite{zheng2022general} & ViT-B/16 & VLP/JEA & Facial\textsuperscript{20M} & 86.2M & 5.58 & 98.15 & 3.58 & 99.40 & 9.70 & 96.98 & 16.65 & 90.27 & 8.88 & 96.20 \\
MCF~\cite{wang2023toward} & ViT-B/16 & SSL/JEA & Facial\textsuperscript{20M} & 86.2M & \uline{4.00} & \uline{98.84} & 8.46 & 96.90 & 8.02 & 97.39 & 10.70 & 95.64 & 7.80 & 97.19 \\
\textit{FSFM}~\cite{wang2025fsfm}
% $^{\textit{(Pre)}}$ 
& \textit{ViT-B/16} & \textit{SSL/JEA} & \textit{Facial\textsuperscript{3M}} & \textit{86.2M} & \textit{3.78} & \textit{99.15} & \textit{3.16} & \textit{99.41} & \textit{4.63} & \textit{99.03} & \textit{7.68} & \textit{97.11} & \textit{4.81} & \textit{98.68} \\
\rowcolor[rgb]{1,0.965,0.8} FS-VFM & ViT-B/16 & SSL/JEA & Facial\textsuperscript{3M} & 86.2M & 4.15 & \textbf{98.92} & \textbf{2.40} & \textbf{99.67} & \textbf{2.43} & \textbf{99.55} & \textbf{4.99} & \textbf{98.62} & \textbf{3.49} & \textbf{99.19} \\ \hdashline[1pt/1pt]
\rowcolor[rgb]{1,0.965,0.8}\textit{\&FS-Adapter} & ViT-B/16 & SSL/JEA & Facial\textsuperscript{3M} & \textbf{1.1M} & 9.92 & 96.27 & 4.81 & 98.97 & \uline{7.74} & \uline{97.59} & 11.04 & 95.20 & 8.38 & 97.01 \\ \hline
Scratch~\cite{dosovitskiy2020image} & ViT-L/16 & None & Rand.Init. & 303.8M & 18.94 & 85.77 & 31.47 & 72.67 & 16.81 & 90.50 & 34.65 & 68.96 & 25.47 & 79.47 \\
Sup.~\cite{dosovitskiy2020image} & ViT-L/16 & Sup & Natural\textsuperscript{IN} & 303.8M & 7.11 & 96.88 & 10.39 & 95.63 & 15.37 & 91.25 & 10.39 & 95.63 & 10.82 & 94.85 \\
CLIP~\cite{radford2021learning} & ViT-L/14 & VLP & Natural\textsuperscript{LA} & 303.8M & \uline{4.90} & \uline{98.95} & 2.40 & 99.44 & 8.22 & 97.18 & 5.37 & 98.37 & \uline{5.23} & \uline{98.48} \\
MAE~\cite{he2022masked} & ViT-L/16 & SSL/MIM & Natural\textsuperscript{IN} & 303.8M & 11.06 & 94.45 & 22.20 & 82.80 & 11.09 & 95.37 & 22.20 & 82.80 & 16.64 & 88.85 \\
\rowcolor[rgb]{1,0.965,0.8} FS-VFM & ViT-L/16 & SSL/JEA & Facial\textsuperscript{3M} & 303.8M & \textbf{2.00} & \textbf{99.50} & \textbf{1.30} & \textbf{99.87} & \textbf{1.42} & \textbf{99.80} & \textbf{4.22} & \textbf{98.09} & \textbf{2.23} & \textbf{99.31} \\ \hdashline[1pt/1pt]
\rowcolor[rgb]{1,0.965,0.8}\textit{\&FS-Adapter} & ViT-L/16 & SSL/JEA & Facial\textsuperscript{3M} & \textbf{1.6M} & 9.11 & 96.55 & \uline{2.33} & \uline{99.61} & \uline{7.16} & \uline{97.54} & \uline{5.26} & \uline{98.44} & 5.96 & 98.04 \\
\noalign{\hrule height 1pt}
\end{tabular}
}
\endgroup
\end{table}
}

% \noindent\textbf{Comparison with SOTA specialized methods} As shown in~\Cref{tab:dfd_sota}, FS-VFM outperforms all \texttt{DFD}-specialized counterparts, regardless of their pre-training paradigms or backbones, achieving best performance across unseen datasets, at both frame and video levels. 1) Our method significantly surpasses SSL-based methods like NACO (likewise a JEA-based SSL to learn consistent representations of real face videos) and FakeStormer (models spatio-temporal inconsistencies in pseudo-fakes with an MAE encoder). 2) FS-VFM also transcends SOTA detectors with the CLIP ViT-L/14 backbone, including VLFFD, VB, and FCGA, which introduce synthetic image-text pairs, video blending, and facial component guidance, respectively. Our model even exceeds the LVLM-based KFD, which leverages an LLM and a larger ImageBind-Huge encoder. 3) Notably, many well-generalization detectors (FF++\textsuperscript{SD}) rely on simulating pseudo-fake at the image, video, or feature level, especially blending artifacts that are common in CDFV2 and DFDCP datasets, yet struggle with forgeries lacking these clues. In contrast, FS-VFM is grounded in real face representations rather than specific artifacts, and yields pronounced generalization gains on more challenging DFDC (diverse unknown manipulations), WDF (in-the-wild), and CDF++ (three forgery types from 22 recent methods). 4) In summary, with simple fine-tuning of a vanilla ViT, FS-VFM delivers SOTA generalization without any task-specific modules or tailored data generation for \texttt{DFD}.

\noindent\textbf{Comparison with SOTA specialized methods} In~\Cref{tab:dfd_sota}, FS-VFM outperforms all \texttt{DFD}-specialized counterparts, regardless of their pre-training paradigms or backbones, achieving best performance across unseen datasets at both frame and video levels. 1) Our method significantly surpasses SSL-based methods like NACO (likewise a JEA-based SSL to learn consistent representations of real face videos) and FakeStormer (models spatio-temporal inconsistencies in pseudo-fakes with an MAE encoder). 2) FS-VFM also transcends SOTA detectors with the CLIP ViT-L/14 backbone, including VLFFD, VB, and FCGA, which introduce synthetic image-text pairs, video blending, and facial component guidance, respectively, and even exceeds the LVLM-based KFD, which leverages an LLM and a larger ImageBind-Huge encoder. 3) Notably, many well-generalization detectors (FF++\textsuperscript{SD}) rely on simulating pseudo-fake at the image, video, or feature level, especially blending artifacts that are common in CDFV2 and DFDCP datasets, yet struggle with forgeries lacking these clues. In contrast, FS-VFM is grounded in real face representations rather than specific artifacts, yielding pronounced generalization on more challenging DFDC (diverse unknown manipulations), WDF (in-the-wild), and CDF++ (three forgery types from 22 recent methods). 4) In summary, with simple fine-tuning of a vanilla ViT, FS-VFM delivers SOTA generalization without any task-specific modules or tailored data generation for \texttt{DFD}.

{
\setlength{\abovecaptionskip}{0pt}
\begin{table}
\centering
\caption{Cross-domain evaluation on face anti-spoofing (\textbf{\texttt{FAS}}). The results of SOTA specialized methods are cited from their original papers. \textbf{Best results}, \underline{second-best}.}
\label{tab:fas_sota}
\begingroup
\setlength{\tabcolsep}{1.75pt}
\renewcommand{\arraystretch}{1.1}
\resizebox{1\linewidth}{!}{%
\begin{tabular}{l c r@{~/~}l r@{~/~}l r@{~/~}l r@{~/~}l c} \noalign{\hrule height 1pt}
\multirow{2}{*}{Method} & \multirow{2}{*}{\begin{tabular}[c]{@{}c@{}}Pre-train\\Manner/Type\end{tabular}} & \multicolumn{2}{c}{OCI→M} & \multicolumn{2}{c}{OMI→C} & \multicolumn{2}{c}{OCM→I} & \multicolumn{2}{c}{ICM→O} & \multirow{2}{*}{\begin{tabular}[c]{@{}c@{}}\textbf{Avg. }\\\textbf{HTER↓}\end{tabular}} \\ \cline{3-10}
 &  & \scriptsize HTER & \scriptsize AUC & \scriptsize HTER & \scriptsize AUC & \scriptsize HTER & \scriptsize AUC & \scriptsize HTER & \scriptsize AUC &  \\ \noalign{\hrule height 1pt}
\multicolumn{11}{l}{\small \textit{SOTA FAS-specialized method (Venue)}} \\
MADDG~\cite{shao2019multi} (CVPR'19)† & Scratch/None & 17.69 & 88.06 & 24.50 & 84.51 & 22.19 & 84.99 & 27.98 & 80.02 & 23.09 \\
NAS-FAS~\cite{yu2020fas} (TPAMI'20) & Scratch/NAS & 16.85 & 90.42 & 15.21 & 92.64 & 11.63 & 96.98 & 13.16 & 94.18 & 14.21 \\
SSDG-R~\cite{jia2020single} (CVPR'20)‡ & Sup\textsuperscript{IN}/Natural & 7.38 & 97.17 & 10.44 & 95.94 & 11.71 & 96.59 & 15.61 & 91.54 & 11.29 \\
PatchNet~\cite{wang2022patchnet} (CVPR'22)‡ & Sup\textsuperscript{IN}/Natural & 7.10 & 98.46 & 11.33 & 94.58 & 13.40 & 95.67 & 11.82 & 95.07 & 10.91 \\
SSAN-R~\cite{wang2022domain} (CVPR'22)‡ & Sup\textsuperscript{IN}/Natural & 6.67 & 98.75 & 10.00 & 96.67 & 8.88 & 96.79 & 13.72 & 93.63 & 9.82 \\
UDG-FAS~\cite{liu2023towards} (ICCV'23)‡ & SSL\textsuperscript{ID}/LOO & 7.14 & 97.31 & 11.44 & 95.59 & 6.28 & 98.61 & 12.18 & 94.36 & 9.26 \\
UDG-FAS~\cite{liu2023towards} (ICCV'23)‡ & Sup\textsuperscript{IN}/Natural & 5.95 & 98.47 & 9.82 & 96.76 & 5.86 & 98.62 & 10.97 & 95.36 & 8.15 \\
IADG~\cite{zhou2023instance} (CVPR'23)† & Scratch/None & 5.41 & 98.19 & 8.70 & 96.44 & 10.62 & 94.50 & 8.86 & 97.14 & 8.40 \\
SAFAS~\cite{sun2023rethinking} (CVPR'23)‡ & Sup\textsuperscript{IN}/Natural & 5.95 & 96.55 & 8.78 & 95.37 & 6.58 & 97.54 & 10.00 & 96.23 & 7.83 \\
GAC-FAS~\cite{le2024gradient} (CVPR'24)‡ & Sup\textsuperscript{IN}/Natural & 5.00 & 97.56 & 8.20 & 95.16 & 4.29 & 98.87 & 8.60 & 97.16 & 6.52 \\
TTDG-V~\cite{zhou2024test} (CVPR'24)* & Sup\textsuperscript{IN}/Natural & 4.16 & 98.48 & 7.59 & 98.18 & 9.62 & 98.18 & 10.00 & 96.15 & 7.84 \\
AG-FAS~\cite{long2024generalized} (TPAMI'24) & Hybrid & 5.71 & 98.03 & 5.44 & 98.55 & 6.71 & 98.23 & 9.43 & 95.52 & 6.82 \\ \hline
ViTAF-ViT\cite{dosovitskiy2020image} (ECCV'22)* & Sup\textsuperscript{IN}/Natural & \uline{1.58} & \textbf{99.68} & 5.70 & 98.91 & 9.25 & 97.15 & 7.47 & 98.42 & 6.00 \\
FLIP-MCL\cite{srivatsan2023flip} (ICCV'23)* & VLP\textsuperscript{CLIP}/Natural & 4.95 & 98.11 & \textbf{0.54} & \textbf{99.98} & 4.25 & 99.07 & \uline{2.31} & \textbf{99.63} & 3.01 \\
CFPL~\cite{liu2024cfpl} (CVPR'24)* & VLP\textsuperscript{CLIP}/Natural & \textbf{1.43} & 99.28 & 2.56 & 99.10 & 5.43 & 98.41 & 2.50 & 99.42 & 2.98 \\
FGPL~\cite{hu2024fine} (MM'24)* & VLP\textsuperscript{CLIP}/Natural & 2.86 & 98.12 & 3.89 & 98.19 & \uline{3.50} & \uline{99.54} & \textbf{1.77} & 99.23 & 3.01 \\
OTA~\cite{li2025optimal} (CVPR'25)* & VLP\textsuperscript{CLIP}/Natural & 2.14 & 99.47 & 2.00 & 99.75 & 4.85 & 98.81 & 2.61 & \uline{99.52} & \uline{2.91} \\ \hdashline[1pt/1pt]
\multicolumn{11}{l}{\small \textit{Simple Fine-Tuning w/o task-specific methodology}} \\
\cellcolor[rgb]{1,0.965,0.8} FS-VFM (Ours) & \cellcolor[rgb]{1,0.965,0.8} SSL\textsuperscript{JEA}/Facial
& \multicolumn{2}{>{\columncolor[rgb]{1,0.965,0.8}}c}{\textbf{~2.00}~/~\uline{99.50}}
& \multicolumn{2}{>{\columncolor[rgb]{1,0.965,0.8}}c}{\uline{~1.30}~/~\uline{99.87}}
& \multicolumn{2}{>{\columncolor[rgb]{1,0.965,0.8}}c}{\textbf{~1.42}~/~\textbf{99.80}}
& \multicolumn{2}{>{\columncolor[rgb]{1,0.965,0.8}}c}{~4.22~/~98.09}
& \cellcolor[rgb]{1,0.965,0.8}\textbf{2.23} \\
\noalign{\hrule height 1pt}
\multicolumn{11}{l}{\small \textit{\textbf{\textbf{Backbone (or modified):}}}~ a CNN from MADDG~\cite{shao2019multi}~†~~~~ResNet-18~‡~~~~ViTs~*}
\end{tabular}
}
\endgroup
\end{table}
}

%-------------------------------------------------------------------------
\subsection{Cross-Domain Face Anti-Spoofing}
\label{subsec:exp_fas}

% \noindent\textbf{Setting} To evaluate the transferability of our method for \texttt{FAS} under domain shifts, we use four widely used benchmark datasets: MSU-MFSD (M)~\cite{wen2015face}, CASIA-FASD (C)~\cite{zhang2012face}, Idiap Replay-Attack (I)~\cite{chingovska2012effectiveness}, and OULU-NPU (O)~\cite{boulkenafet2017oulu}. We treat each dataset as the target domain and apply the leave-one-out (LOO) cross-domain evaluation. We follow the 0-shot setting and data setups of prior works~\cite{huang2022adaptive, srivatsan2023flip}, as they also fine-tune the vanilla ViT for this MCIO protocol. We append the task head after averaging all non-[CLS] tokens instead of the [CLS] one, to keep it the same as other tasks. We report the mean HTER (Half Total Error Rate) and AUC over 5 runs. 

\noindent\textbf{Setting} To evaluate the transferability of our method for \texttt{FAS} under domain shifts, we apply the leave-one-out (LOO) cross-domain evaluation on four widely used benchmark datasets: MSU-MFSD (M)~\cite{wen2015face}, CASIA-FASD (C)~\cite{zhang2012face}, Idiap Replay-Attack (I)~\cite{chingovska2012effectiveness}, and OULU-NPU (O)~\cite{boulkenafet2017oulu}. We follow the 0-shot setting and data setups of prior works~\cite{huang2022adaptive, srivatsan2023flip}, as they also fine-tune the vanilla ViT for this protocol. We append the task head after averaging all non-[CLS] tokens instead of the [CLS] one, to keep it the same as other tasks. We report the mean HTER (Half Total Error Rate) and AUC over 5 runs.

% \noindent\textbf{Comparison with existing VFMs} In~\Cref{tab:fas_vfm}, FS-VFM (ours) achieves the best cross-domain generalization upon existing VFMs across all ViT scales. We observe: 1) Supervised ImageNet pre-training is still a competitive initiation scheme for \texttt{FAS}, as also noted in~\cite{huang2022adaptive, zhou2024test}. 2) Fine-tuning generic SSL models, including both MIM-based MAE and ID-based DINO, exhibits inferior performance under domain shifts, while the VLP-based CLIP improves. 3) Even large-scale facial pre-training, FaRL and MCF ViT-B/16, underperform their corresponding CLIP and ImageNet Supervised baselines, again underscoring the gap between face analysis and security tasks. 4) Our FS-VFMs effectively bridge these gaps and achieve dramatically better generalization to unseen spoofing domains. 5) \textit{\&FS-Adapter ET}, which freezes the ViT, still retains the competitive cross-domain robustness. Similar to \texttt{DFD}, it scales cost-effectively with a larger backbone. With the FS-VFM ViT-L/16, \&FS-Adapter outperforms all other fully fine-tuned VFMs in 3 (C/I/O) out of 4 target domains, and only slightly lags behind the CLIP ViT-L/14 in average, while training solely its 0.527\% (1.6M/303.8M) parameters. 6) Overall, our FS-VFMs improve the generalizability of ViT for cross-domain \texttt{FAS}, effectively modeling domain-agnostic, credible features of live (real) faces.

\noindent\textbf{Comparison with existing VFMs} In~\Cref{tab:fas_vfm}, FS-VFM (ours) achieves the best cross-domain generalization upon existing VFMs across all ViT scales. We observe: 1) ImageNet supervised ViTs remain a competitive initiation for \texttt{FAS}, as also noted in~\cite{huang2022adaptive, zhou2024test}. 2) Fine-tuning generic SSL models, including both MIM-based MAE and ID-based DINO, transfer poorly to unseen spoof domains, while the VLP-based CLIP improves. 3) Even large-scale facial pre-training, FaRL and MCF ViT-B/16, underperform their corresponding CLIP and ImageNet Supervised baselines, again underscoring the gap between face analysis and security tasks. 4) Our FS-VFMs effectively bridge these gaps and achieve dramatically better generalization under domain shifts. 5) \textit{\&FS-Adapter ET}, with the ViT frozen, still retains the competitive cross-domain robustness. Similar to \texttt{DFD}, it scales cost-effectively with a larger backbone. With the FS-VFM ViT-L/16, \&FS-Adapter outperforms all other fully fine-tuned VFMs on 3 (C/I/O) out of 4 target domains, and nearly closes the CLIP ViT-L/14 in average, while training solely its 0.527\% (1.6M/303.8M) parameters. 6) Overall, our FS-VFMs effectively model domain-agnostic, credible features of live (real) faces, improving the generalizability of ViT for cross-domain \texttt{FAS}.

% 4) FS-VFM effectively bridges these gaps: using just a ViT-S/16, its averaged metrics surpass most VFMs built on larger ViT-B and ViT-L. Scaling up further, FS-VFMs with ViT-B/16 and ViT-L/16 achieve dramatically leading performance.
% 4) FS-VFM effectively bridges these gaps: using just a ViT-S/16, its averaged metrics surpass most VFMs built on larger ViT-B and ViT-L, and generalizes dramatically better to unseen spoofing domains as backbones scale.

\noindent\textbf{Comparison with SOTA specialized methods} Against \texttt{FAS}-specialized methods in~\Cref{tab:fas_sota}, FS-VFM achieves the lowest average HTER across the LOO scenarios, and the top-tier performance on 3 (M/C/I) out of 4 domains. Crucially, this is attained by simply fine-tuning a vanilla ViT from FS-VFM, using only a standard cross-entropy loss and the baseline setup in prior works~\cite{dosovitskiy2020image, srivatsan2023flip}, without any task-specific modules or domain generalization techniques. In contrast, recent arts leverage CLIP models and elaborate on tackling domain shifts, such as learnable content/style queries and text prompts~\cite{liu2024cfpl}, separated domain-agnostic and domain-specific prompts with a convolutional adapter~\cite{hu2024fine}, and a prototype model with test-time adaptation~\cite{li2025optimal}. FS-VFM matches or surpasses these methods, demonstrating that a strong facial representation is transferable for robust face presentation attack detection.

% \noindent\textbf{Comparison with SOTA specialized methods} Against \texttt{FAS}-specialized methods in~\Cref{tab:fas_sota}, FS-VFM achieves the lowest average HTER and the top-tier performance on 3 (M/C/I) out of 4 LOO scenarios. Crucially, this is attained by simply fine-tuning a vanilla ViT from FS-VFM, using only a standard cross-entropy loss and the baseline setup in prior works~\cite{dosovitskiy2020image, srivatsan2023flip}, without any task-specific modules or domain generalization techniques. In contrast, recent arts leverage CLIP models and elaborate on tackling domain shifts, \eg, learnable content/style queries and text prompts~\cite{liu2024cfpl}, separated domain-agnostic/specific prompts plus a convolutional adapter~\cite{hu2024fine}, and prototype-based test-time adaptation~\cite{li2025optimal}. FS-VFM matches or surpasses these methods, demonstrating that a strong facial representation is transferable for robust face presentation attack detection.

% While optimization for a specific downstream task is beyond this study, incorporating special auxiliary supervision or DG techniques into our pre-trained model may further improve its generalization for face presentation attack detection.

%-------------------------------------------------------------------------
\subsection{Unseen Diffusion-Generated Faces Forensic}
\label{subsec:exp_diff}
\noindent\textbf{Setting} To further assess the adaptability of our method against emerging unknown face forgeries, we extend the \texttt{DFD} (\Cref{subsec:exp_deeepfake}) to stress-test the cross-distribution DiFF~\cite{cheng2024diffusion} benchmark, which comprises high-quality synthetic face images from 13 recent diffusion models across four subsets: Text-to-Image (T2I), Image-to-Image (I2I), Face Swapping (FS), and Face Editing (FE). We train \textit{one} detector on the FF++ (c23) \textit{DeepFakes} subset (only an early face-swapping algorithm), and report AUCs on the DiFF test subsets. This setting is more challenging than the typical \texttt{DFD} (\Cref{subsec:exp_deeepfake}) given unseen, heterogeneous generators and manipulations.

{
\setlength{\abovecaptionskip}{0pt}
\begin{table}
\centering
\caption{Cross-dataset evaluation on the (\textbf{\texttt{DiFF}}) benchmark~\cite{cheng2024diffusion}. All models are fine-tuned only on FF++\_DeepFakes/c23 subset~\cite{rossler2019faceforensics++}.}
\label{tab:DiFF}
\begingroup
\setlength{\tabcolsep}{4pt}
\renewcommand{\arraystretch}{1.1}
\resizebox{\linewidth}{!}{%
\begin{tabular}{lcccccccccc} \noalign{\hrule height 1pt}
\multirow{2}{*}{Method} & \multirow{2}{*}{Backbone} & \multicolumn{2}{c}{Pre-train} & \multirow{2}{*}{\begin{tabular}[c]{@{}c@{}}Train.\\Param.\end{tabular}} & \multicolumn{5}{c}{Test Subset \textbf{AUC↑ (\%)}} & \multirow{2}{*}{\begin{tabular}[c]{@{}c@{}}\textbf{Avg.}\\\textbf{w/o} F+\end{tabular}} \\ \cline{3-4}\cline{6-10}
 &  & Manner & Type &  & FF++ & T2I & I2I & FS & FE &  \\ \noalign{\hrule height 1pt}
Scratch~\cite{dosovitskiy2020image} & ViT-S/16 & None & Rand.Init. & 21.6M & 86.78 & 38.28 & 29.49 & 36.99 & 31.39 & 34.04 \\
Supervised~\cite{dosovitskiy2020image} & ViT-S/16 & Sup & Natural\textsuperscript{IN} & 21.6M & 98.88 & 59.43 & 58.38 & 60.41 & 46.19 & 56.10 \\
DINO~\cite{caron2021emerging} & ViT-S/16 & SSL/ID & Natural\textsuperscript{IN} & 21.6M & \uline{99.12} & \uline{75.83} & \uline{70.92} & 53.68 & 42.86 & 60.82 \\
\rowcolor[rgb]{1,0.965,0.8} FS-VFM (Ours) & ViT-S/16 & SSL/JEA & Facial\textsuperscript{3M} & 21.6M & \textbf{99.13} & \textbf{77.10} & \textbf{77.36} & \textbf{72.01} & \textbf{75.72} & \textbf{75.55} \\ \hdashline[1pt/1pt]
\rowcolor[rgb]{1,0.965,0.8} \textit{\& FS-Adapter ET} & ViT-S/16 & SSL/JEA & Facial\textsuperscript{3M} & \textbf{0.085M} & 94.04 & 65.17 & 58.61 & \uline{73.82} & \uline{48.43} & \uline{61.51} \\ \hline
Scratch~\cite{dosovitskiy2020image} & ViT-B/16 & None & Rand.Init. & 85.6M & 88.02 & 41.88 & 33.69 & 40.42 & 36.15 & 38.04 \\
Supervised~\cite{dosovitskiy2020image} & ViT-B/16 & Sup & Natural\textsuperscript{IN} & 85.6M & 98.68 & 62.67 & 59.94 & 55.84 & 47.00 & 56.36 \\
CLIP~\cite{radford2021learning} & ViT-B/16 & VLP & Natural\textsuperscript{LA} & 85.8M & 99.52 & 38.71 & 37.03 & 38.40 & 38.65 & 38.20 \\
MAE~\cite{he2022masked} & ViT-B/16 & SSL/MIM & Natural\textsuperscript{IN} & 85.6M & 99.65 & 56.92 & 56.24 & 60.66 & 34.79 & 52.15 \\
DINO~\cite{caron2021emerging} & ViT-B/16 & SSL/ID & Natural\textsuperscript{IN} & 85.6M & 99.57 & \uline{76.49} & \uline{73.90} & 63.16 & 49.67 & 65.80 \\
FaRL~\cite{zheng2022general} & ViT-B/16 & VLP/JEA & Facial\textsuperscript{20M} & 85.8M & \textbf{99.74} & 43.79 & 43.05 & 48.79 & 45.12 & 45.19 \\
MCF~\cite{wang2023toward} & ViT-B/16 & SSL/JEA & Facial\textsuperscript{20M} & 85.6M & 99.54 & 70.62 & 67.74 & 65.11 & 44.54 & 62.00 \\
\textit{FSFM}~\cite{wang2025fsfm} $^{\textit{(Pre)}}$ & \textit{ViT-B/16} & \textit{SSL/JEA} & \textit{Facial\textsuperscript{3M}} & \textit{85.6M} & \textit{99.28} & \textit{88.20} & \textit{89.00} & \textit{81.99} & \textit{88.50} & \textit{86.92} \\
\rowcolor[rgb]{1,0.965,0.8} FS-VFM (Ours) & ViT-B/16 & SSL/JEA & Facial\textsuperscript{3M} & 85.6M & \uline{99.68} & \textbf{91.00} & \textbf{91.80} & \textbf{83.16} & \textbf{89.84} & \textbf{88.95} \\ \hdashline[1pt/1pt]
\rowcolor[rgb]{1,0.965,0.8} \textit{\& FS-Adapter ET} & ViT-B/16 & SSL/JEA & Facial\textsuperscript{3M} & \textbf{0.335M} & 98.27 & 75.09 & 71.42 & \uline{86.26} & \uline{68.28} & \uline{75.26} \\ \hline
Scratch~\cite{dosovitskiy2020image} & ViT-L/16 & None & Rand.Init. & 303.1M & 85.87 & 40.28 & 32.40 & 38.70 & 38.70 & 37.52 \\
Supervised~\cite{dosovitskiy2020image} & ViT-L/16 & Sup & Natural\textsuperscript{IN} & 303.1M & 99.06 & 56.75 & 52.86 & 56.49 & 43.67 & 52.44 \\
CLIP~\cite{radford2021learning} & ViT-L/14 & VLP & Natural\textsuperscript{LA} & 303.2M & \uline{99.31} & 48.17 & 45.91 & 64.45 & 50.93 & 52.37 \\
MAE~\cite{he2022masked} & ViT-L/16 & SSL/MIM & Natural\textsuperscript{IN} & 303.1M & 99.30 & 51.70 & 48.51 & 79.96 & 57.03 & 59.30 \\
\rowcolor[rgb]{1,0.965,0.8} FS-VFM (Ours) & ViT-L/16 & SSL/JEA & Facial\textsuperscript{3M} & 303.1M & \textbf{99.59} & \textbf{92.72} & \textbf{92.51} & \textbf{97.17} & \textbf{92.83} & \textbf{93.81} \\ \hdashline[1pt/1pt]
\rowcolor[rgb]{1,0.965,0.8} \textit{\& FS-Adapter ET} & ViT-L/16 & SSL/JEA & Facial\textsuperscript{3M} & \textbf{0.594M} & 98.40 & \uline{80.29} & \uline{81.47} & \uline{96.70} & \uline{82.78} & \uline{85.31} \\ \noalign{\hrule height 1pt}
\end{tabular}
}
\endgroup
\end{table}
}

% \noindent\textbf{Comparison} As shown in~\Cref{tab:DiFF}, FS-VFM (Ours) decisively outperforms other VFMs across all unseen diffusion methods and ViT scales, while maintaining superior in-domain performance. Most existing VFMs severely overfit to the DeepFakes distribution that fails to extrapolate to diffusion face forgeries. By contrast, FS-VFMs benefit from fundamental representations of real faces that transcend specific forgery patterns, thus generalizing significantly better. Moreover, \&FS-Adapter efficient tuning also yields top-tier average AUC scores and stands out for its performance-efficiency balance, especially coupled with ViT-L/16. In summary, these comparisons are mostly consistent with the cross-dataset \texttt{DFD}, with even more pronounced improvements, highlighting the out-of-distribution robustness of our method.

\noindent\textbf{Comparison} In~\Cref{tab:DiFF}, FS-VFM (Ours) decisively outperforms other VFMs across all unseen diffusion methods and ViT scales, while maintaining superior in-domain performance. Most existing VFMs severely overfit to the DeepFakes distribution, failing to extrapolate to diffusion face forgeries. By contrast, FS-VFMs benefit from fundamental representations of real faces that transcend specific forgery patterns, thus generalizing significantly better. Moreover, \&FS-Adapter efficient tuning also yields top-tier average AUC and stands out for its efficiency-performance balance, especially with ViT-L/16. These comparisons are mostly consistent with the cross-dataset \texttt{DFD}, with even more pronounced improvements, highlighting the out-of-distribution robustness of our method.

% Dino's ViT-S/16 and ViT-B/16 perform slightly better than the corresponding &FS-Adapter ET on the T2I and I2I test subsets. We speculate that this is mainly because ID-Based Dino tends to learn global patterns, which helps to identify images synthesized from the entire image in T2I and I2I, and the comparison is based on its full fine-tuning. However, it lags behind our &FS-Adapter ET in both FS FE and average, especially ViT-B/16.

%-------------------------------------------------------------------------
\subsection{Ablation Studies of FS-VFM}
\label{subsec:exp_ablation}
% In this subsection, we conduct extensive ablation experiments to evaluate the effectiveness of each component and its rational design in pre-training FS-VFM. Unless otherwise stated, we pre-trained the FS-VFM ViT-B/16 model on the FF++\_O dataset, which contains $\sim\!0.1M$ real face images extracted from 720 training and 140 validation videos of FF++ (c23) YouTube subset~\cite{rossler2019faceforensics++}. We report the average generalization results, including \texttt{DFD} across the CDFV2, DFDCP, DFDC, and WDF datasets, as well as \texttt{FAS} evaluated on the MCIO cross-domain protocol, as presented in \Cref{tab:ablation}.

In this subsection, we conduct extensive ablations to assess the effectiveness of each component and its rational design in pre-training FS-VFM. Unless specified, we pre-trained the FS-VFM ViT-B/16 model on the FF++\_O dataset, which contains $\sim$0.1M real face images from the FF++ (c23) YouTube subset~\cite{rossler2019faceforensics++}. We report the \textbf{average} generalization metrics, including \texttt{DFD} across \{CDFv2, DFDCP, DFDC, WDF\} datasets and \texttt{FAS} on the MCIO cross-domain protocol, in \Cref{tab:ablation}.

\noindent\textbf{Effect of 3C Objectives} We first evaluate different facial masking strategies on the vanilla MAE. Both our preliminary \textit{FRP} and \textit{CRFR-R} strategies outperform simple \textit{random} masking, confirming the significance of intra-region consistency ($C^{1}$) and inter-region coherency ($C^{2}$), respectively. Notably, the \textit{CRFR-P} strategy emerges as the most effective, highlighting that both $C^{1}$ and $C^{2}$ are essential and complementary for strong facial representations. Building on MAE with CRFR-P, we further introduce the ID network with varied target views in \Cref{fig:ID_target_view}. The consistent improvement with the \textit{\&Full} target view proves the benefit of establishing local-to-global correspondence ($C^{3}$) by complete facial semantics.

{
\setlength{\abovecaptionskip}{0pt}
\begin{table}[!t]
\centering
\caption{Ablations of FS-VFM on cross-dataset \texttt{DFD} and cross-domain \texttt{FAS} with \textbf{averaged} metrics. The FS-VFM ViT-B/16 model is pre-trained on FF++\_O. \colorbox[rgb]{0.85,0.85,0.85}{Default settings}.}
\label{tab:ablation}
\begingroup
\setlength{\tabcolsep}{8pt}
\renewcommand{\arraystretch}{1.1}
\resizebox{.9\linewidth}{!}{%
\begin{tabular}
% {p{2.4cm}<{\centering}p{0.4cm}<{\centering}p{0.4cm}<{\centering}p{0.4cm}<{\centering}p{1.3cm}<{\centering}p{1.3cm}<{\centering}p{1.3cm}<{\centering}p{1.3cm}<{\centering}} 
{cccccccc}
 \noalign{\hrule height 1pt} 
\multirow{2}{*}{\textbf{Component}} & \multicolumn{1}{c}{\multirow{2}{*}{$\mathbf{~C^{1}~}$}} & \multicolumn{1}{c}{\multirow{2}{*}{$\mathbf{~C^{2}~}$}} & \multicolumn{1}{c}{\multirow{2}{*}{$\mathbf{~C^{3}~}$}} & \multicolumn{2}{c}{\small \textbf{Deepfake Detection}} & \multicolumn{2}{c}{\small \textbf{Face Anti-spoofing}} \\ 
\cline{5-8}
 & \multicolumn{1}{c}{} & \multicolumn{1}{c}{} & \multicolumn{1}{c}{} & \textbf{F-AUC↑} & \textbf{V-AUC↑} & \textbf{HTER↓} & \textbf{AUC↑} \\ 
 \noalign{\hrule height 1pt} 
\multicolumn{8}{l}{\small \textit{Vanilla MAE~\&Masking Strategy~(w/o ID Network)}} \\
\&Random (MAE) &  &  &  & 74.19 & 79.51 & 19.05 & 87.42 \\
\&Fasking-I~\cite{cai2023marlin} &  &  &  & 73.80 & 78.33 & \textbf{17.81} & 87.75 \\
\&FRP & $\checkmark$ &  &  & 75.43 & 81.21 & 17.96 & 87.61 \\
\&CRFR-R &  & $\checkmark$ &  & 75.01 & 80.70 & 18.28 & 87.34 \\
\&CRFR-P & $\checkmark$ & $\checkmark$ &  & \textbf{76.11} & \textbf{81.58} & \uline{17.85} & \textbf{88.11} \\ 
\hdashline[1pt/1pt]
\multicolumn{8}{l}{\small \textit{ID \&Target View (w/ MAE\&CRFR-P)}} \\
\&Visible & $\checkmark$ & $\checkmark$ &  & 75.54 & 81.50 & 18.22 & 87.95 \\
\&Masked & $\checkmark$ & $\checkmark$ &  & 76.35 & 81.86 & 18.41 & 87.77 \\
\textbf{\&Full (FS-VFM)} &{\cellcolor[rgb]{0.85,0.85,0.85}}$\checkmark$ & {\cellcolor[rgb]{0.85,0.85,0.85}}$\checkmark$ &{\cellcolor[rgb]{0.85,0.85,0.85}}$\checkmark$ & \textbf{76.39} & \textbf{82.31} & \textbf{17.44} & \textbf{88.26} \\ 
 \noalign{\hrule height 1pt} 
\textbf{Design} & \multicolumn{3}{c}{\textbf{Setting}} & \multicolumn{1}{l}{} & \multicolumn{1}{l}{} & \multicolumn{1}{l}{} & \multicolumn{1}{l}{} \\ 
\hline
\multirow{5}{*}{\begin{tabular}[c]{@{}c@{}}Online\&Target \\ Rep Decoder\\ ($D_{o}^r$\&$D_{t}^r$)\\ Blocks \end{tabular}}  & \multicolumn{3}{c}{0 \& 0} & 75.63 & 81.48 & 18.37 & 86.77 \\
 & \multicolumn{3}{c}{2 \& 0} & 75.74 & 81.14 & 18.54 & 87.22 \\
 & \multicolumn{3}{c}{1 \& 1} & 75.06 & 80.68 & 18.86 & 87.64 \\
 & \multicolumn{3}{c}{{\cellcolor[rgb]{0.85,0.85,0.85}}\textbf{2 \& 2}} & \textbf{\textbf{76.39}} & \textbf{\textbf{82.31}} & \textbf{\textbf{17.44}} & \textbf{\textbf{88.26}} \\
 & \multicolumn{3}{c}{3 \& 3} & 75.08 & 80.71 & 17.93 & 87.80 \\ 
\hdashline[1pt/1pt]
\multirow{3}{*}{\begin{tabular}[c]{@{}c@{}}Online\&Target\\($I_v$\&$I$) Data\\ Augmentation \end{tabular}} & \multicolumn{3}{c}{(crop+flip)\&none} & 75.93 & 81.54 & 18.24 & 87.04 \\
 & \multicolumn{3}{c}{none\&(crop+flip)} & 73.39 & 78.80 & 19.13 & 86.11 \\
 & \multicolumn{3}{c}{{\cellcolor[rgb]{0.85,0.85,0.85}} \textbf{none\&none}} & {\textbf{76.39}} & \textbf{82.31} & \textbf{17.44} & \textbf{88.26} \\ 
\hdashline[1pt/1pt]
\multirow{3}{*}{\begin{tabular}[c]{@{}c@{}} Loss \\for ID\end{tabular}} & \multicolumn{3}{c}{InfoNCE} & 75.10 & 80.60 & 18.24 & 87.37 \\
 & \multicolumn{3}{c}{\cite{grill2020bootstrap}-like MSE} & 74.19 & 79.34 & 18.09 & 88.21 \\
 & \multicolumn{3}{c}{{\cellcolor[rgb]{0.85,0.85,0.85}}\textbf{Asym.~\Cref{eq:loss_id}}} & \textbf{76.39} & \textbf{82.31} & \textbf{17.44} & \textbf{88.26} \\
\hdashline[1pt/1pt]
 \multirow{3}{*}{\begin{tabular}[c]{@{}c@{}}Pre-training \\ Epoch  \end{tabular}}  
 & \multicolumn{3}{c}{200} & 74.20 & 79.14 & 17.96 & 87.71 \\
 & \multicolumn{3}{c}{{\cellcolor[rgb]{0.85,0.85,0.85}}400} & 76.39 & 82.31 & 17.44 & 88.26 \\
 & \multicolumn{3}{c}{600} & \textbf{77.37} & \textbf{83.86} & \textbf{15.97} & \textbf{91.28} \\ 
 \noalign{\hrule height 1pt} 
\end{tabular}}%}
\endgroup
\end{table}
}

% \noindent\textbf{Effect of Key Designs} \textbf{1) \textit{Masking Ratio}} $r{=}0.75$ yields a better trade-off. Lower ratios (\textit{0.35, 0.50, 0.65}) leave excessive patches visible, making the reconstruction and alignment tasks trivial. Conversely, a higher ratio (\textit{0.85}) renders pretext tasks too challenging to learn sufficient fine-grained representations for face security tasks. \textbf{2) \textit{Rep Decoders}} Using 2 ViT blocks for both online and target rep decoders (\textit{2\&2} for $D_{o}^r$\&$D_{t}^r$) strikes the complexity-generalization balance, outperforming fewer (\textit{1\&1}) or more (\textit{3\&3}) layers. Omitting the rep decoder (\textit{0\&0}) or appending it only to the online branch (\textit{2\&0}) degrades performance, verifying the gain of a disentangled representation space to bridge the feature distribution gap. \textbf{3) \textit{Data Augmentation}} FS-VFM performs best even without any augmentation to both online and target views (\textit{none\&none}). Unlike other SSL methods, applying simple augmentation in the MIM network (\textit{(crop+flip)\&none}) or target view (\textit{none\&(crop+flip)}) hurts generalization, suggesting that our CRFR-P masking already offers adequate spatial regularization and that preserving original faces intact for target views aids robustness. \textbf{4) \textit{ID Loss}} Our asymmetric negative cosine similarity (\Cref{eq:loss_id}) proves more effective than the BYOL-like~\cite{grill2020bootstrap} MSE and the widely used InfoNCE~\cite{oord2018representation}.

\noindent\textbf{Effect of Key Designs} \textit{1) Rep Decoders} Using 2 ViT blocks for both online and target rep decoders (\textit{2\&2} for $D_{o}^r$\&$D_{t}^r$) strikes the complexity-generalization balance, outperforming fewer (\textit{1\&1}) or more (\textit{3\&3}) layers. Omitting the rep decoder (\textit{0\&0}) or appending it only to the online branch (\textit{2\&0}) degrades performance, verifying the gain of a disentangled representation space to bridge the feature distribution gap. \textit{2) Data Augmentation} FS-VFM performs best even without any augmentation to both online and target views (\textit{none\&none}). Unlike other SSL methods, applying simple augmentation in the MIM network (\textit{(crop+flip)\&none}) or target view (\textit{none\&(crop+flip)}) hurts generalization, suggesting that our CRFR-P masking already offers adequate spatial regularization and that preserving original faces intact for target views aids robustness. \textit{3) ID Loss} Our asymmetric negative cosine similarity (\Cref{eq:loss_id}) proves more effective than the BYOL-like~\cite{grill2020bootstrap} MSE and the widely used InfoNCE~\cite{oord2018representation}. \textit{4) Pre-training Epochs} FS-VFM pre-trained for 200 epochs achieves performance comparable to that of the vanilla MAE baseline pre-trained for 400 epochs, suggesting that FS-VFM learns stronger facial representations more effectively. With longer pre-training schedules, FS-VFM consistently yields improved initialization weights for downstream tasks.

% \textit{4) Pre-training Epochs} The performance of FS-VFM pre-trained for 200 epochs is comparable to the MAE pre-trained for 400 epochs, which suggests FS-VFM is stronger to learn better representations. With longer pre-training schedules, FS-VFM consistently learns better initialization weights for downstream tasks.
% with extended pre-training, FS-VFM consistently yields better initialization weights for downstream tasks.
% With longer pre-training schedules, FS-VFM consistently yields improved initialization weights for downstream tasks.

% With longer pre-training schedules, FS-VFM consistently learns better initialization weights for downstream tasks.
% 此外，三表格表格种，FS-VFM均是预训练400 epoch的，其始终优于预训练400 epoch的FSFM。
% 预训练200epoch的FSFM和运训练400的MAE性能相当。

% the longer pre-training schedule makes the model learn better initialization weights for fine-tuning [44],
% This result suggests the stronger capability of CMAE on learning better representations.

% I-JEPA outperforms MAE while requiring less pretraining epochs when
% using a similar encoder architecture. I-JEPA, using a ViTH/14 architecture, matches the performance of a ViT-L/16
% pretrained with data2vec [7], while using significantly less
% computational effort (see Section 7).

{
\setlength{\abovecaptionskip}{0pt}
\begin{figure*}[ht!]
\centering
\includegraphics[width=.95\linewidth]{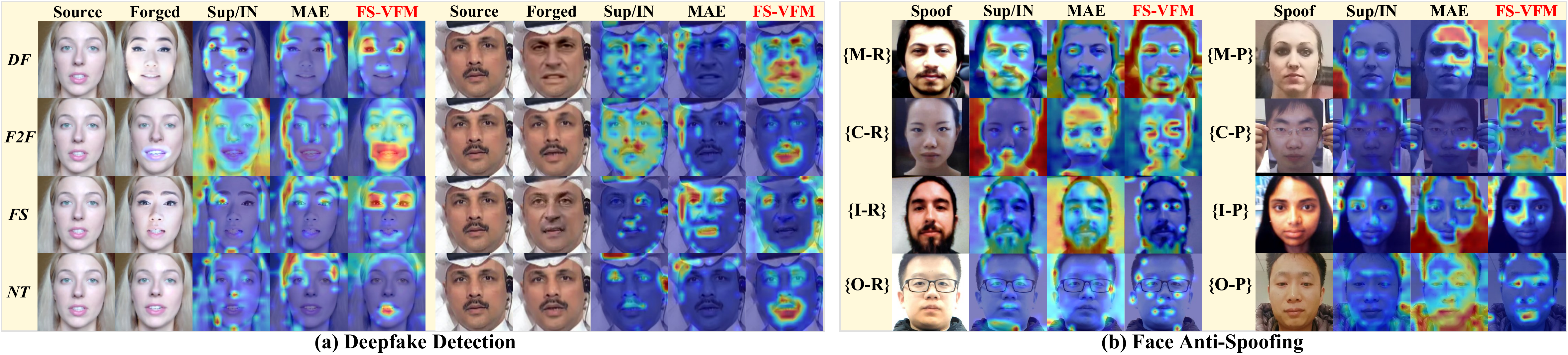}
\caption{\label{fig:vis_cam} CAM Visualizations. (a) \texttt{DFD} on various FF++~\cite{rossler2019faceforensics++} manipulations (DF/DeepFakes, F2F/Face2Face, FS/FaceSwap, NT/NeuralTextures). (b) \texttt{FAS} on the cross-domain MCIO protocol (R/Replays, P/Print or Photo). FS-VFM clearly highlights forgery artifacts and spoofing clues. Images are from the test set.}
\end{figure*}
}

{
\setlength{\abovecaptionskip}{0pt}
\begin{figure}[t!]
\centering
\includegraphics[width=\linewidth]{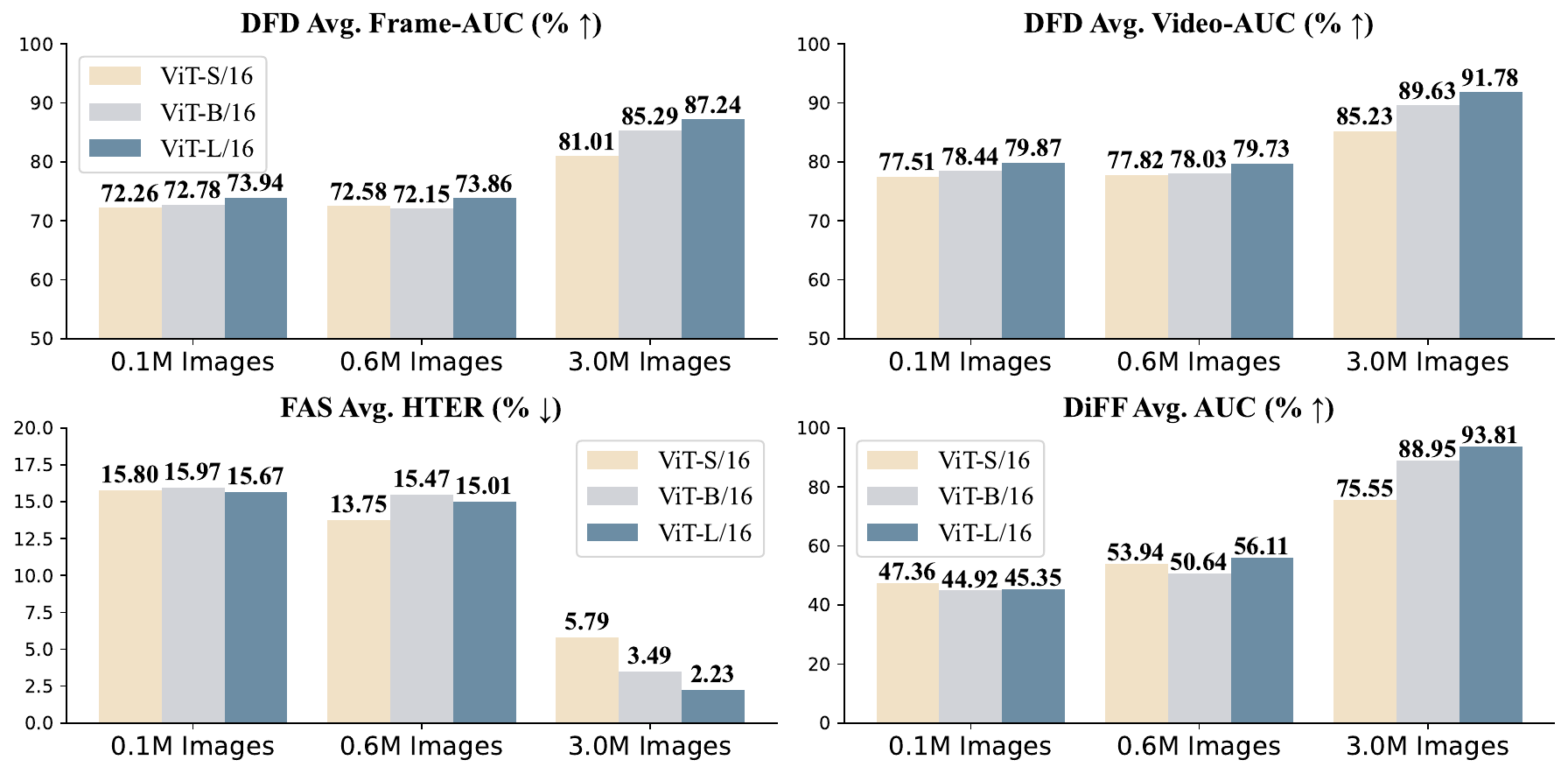}
% \vspace{-5pt}
\caption{Ablations of scaling data and model sizes for pre-training FS-VFM.}
\label{fig:scale} 
\end{figure}
}

\subsection{Scalability of FS-VFM}
\label{sec:scale}
As shown in~\Cref{fig:scale}, scaling up the pre-training data (from 0.1M, 0.6M, to 3.0M facial images) and model capacity (from ViT S/16, B/16, to L/16) systematically improves generalization across face security tasks. A larger, more diverse dataset enables the model to learn richer facial representations, substantially boosting downstream transfer robustness, which is encouraging given the abundant unlabeled face data available in both academia and the real world. Moreover, a larger model further enhances marginal capacity. In particular, larger ViT backbones see more pronounced gains from data scaling than smaller ViTs, \eg, from 0.1M to 3M pre-training images, ViT S/16, B/16, and L/16 increase 8.75\%, 12.51\%, and 13.30\% frame-level AUC on cross-dataset DFD, reduce 7.72, 11.19, and 11.91 HTER on cross-domain FAS, respectively. These results demonstrate the promising scalability of FS-VFM.
\subsection{Ablation Studies of FS-Adapter}\label{subsec:abbla_fsa}

We ablate the FS-Adapter for efficiency-performance trade-offs on downstream face security tasks, especially built upon the FS-VFM ViT-L/16, and report corresponding (\Cref{tab:dfd_vfm},~\Cref{tab:fas_vfm}, and~\Cref{tab:DiFF}) averaged metrics in~\Cref{tab:abla_fsa}. The frozen FS-VFM features a 24-layer ViT-L/16 with 1024-dimensional embeddings, while the adapters' bottleneck down-samples $4\times$, \ie, $l\!=\!24, d\!=\!1024$, and $b\!=\!d/4\!=\!256$. For reference, \textit{Vanilla Adapter}~\cite{houlsby2019parameter} inserted in all layers achieves strong results, but is relatively parameter-intensive and requires heavy backpropagation through the backbone. Next, we append the adapter only at the last layer as the baseline \textit{Variant 1}, which yields clear performance drops on all tasks, albeit reducing most trainable parameters. This suggests that a minimal, straightforward adaptation lacks sufficient domain knowledge for generalization. We thus explore: \textit{Variant 2} adds supervised contrastive learning, where the only difference from FS-Adapter is $\mathcal{L}_\mathit{scl}$ that further pulls fake faces closer, but improves minimally over Variant 1. This verifies that our real-anchor contrastive learning with $\mathcal{L}_\mathit{rac}$ regularizes a better feature space for face security. \textit{Variant 3} projects upon the adapter feature $f_a$ instead of the bottleneck feature $f_b$, which increases $2.67\!\times\!$ parameters but declines metrics, proving that the constraint in a compact bottleneck space is both efficient and effective. \textit{Variant 4} confirms that using a projector $w_{\mathit{lp}}$, the common practice in contrast learning, is necessary, despite introducing negligible $0.066M$ parameters. Finally, our \textit{FS-Adapter} trains just $4.69\%$ ($0.59M\!/\!12.38M$) parameters of the Vanilla Adapter, but delivers even better overall performance, especially in unseen \texttt{DiFF}. These ablations establish the plug-and-play FS-Adapter that enables ultra-efficient and highly flexible transfer of foundational facial representations to downstream face security tasks, retaining superior generalization.

{
\setlength{\abovecaptionskip}{0pt}
\begin{table}
\centering
\caption{Ablations of FS-Adapter (\&FS-VFM ViT-L/16) on cross-dataset \texttt{DFD}, cross-domain \texttt{FAS}, unseen \texttt{DiFF} with \textbf{averaged} metrics.}
\label{tab:abla_fsa}
\begingroup
\setlength{\tabcolsep}{2pt}
\renewcommand{\arraystretch}{1.1}
\resizebox{\linewidth}{!}{%
\begin{tabular}{ccccccccccccc} \noalign{\hrule height 1pt}
\multirow{2}{*}{\begin{tabular}[c]{@{}c@{}}Adapter\\(ViT-L/16)\end{tabular}} & \multirow{2}{*}{\begin{tabular}[c]{@{}c@{}}Insert\\layer\end{tabular}} & \multicolumn{4}{c}{Contrastive Learning} & \multicolumn{2}{c}{\multirow{2}{*}{\begin{tabular}[c]{@{}c@{}}Tuned Params.\\of Adapter\end{tabular}}} & \multicolumn{2}{c}{\textbf{\texttt{DFD}}} & \multicolumn{2}{c}{\textbf{\texttt{FAS}}} & \textbf{\texttt{DiFF}} \\ \cline{3-6}\cline{9-13}
 &  & Feat & $w_{\mathit{lp}}$ & $\mathcal{L}_\mathit{scl}$ & $\mathcal{L}_\mathit{rac}$ & \multicolumn{2}{c}{} &\scriptsize F-AUC↑ &\scriptsize V-AUC↑ &\scriptsize HTER↓ &\scriptsize AUC↑ &\scriptsize AUC↑ \\ \noalign{\hrule height 1pt}
Vanilla~\cite{houlsby2019parameter} & all-$l$ & \multicolumn{4}{c}{} & $l\!*\!(2db)$ & 12.58M & \textbf{83.62} & \uline{85.57} & \uline{5.98} & \textbf{98.45} & 80.98 \\ \hdashline[1pt/1pt]
Variant 1 & last & \multicolumn{4}{c}{} & $2db$ & 0.524M & 81.79 & 84.16 & 6.87 & 97.08 & 77.68 \\
Variant 2 & last & $f_{\mathrm{b}}$ & $\checkmark$ & $\checkmark$ &  & $2db\!+\!bb$ & 0.590M & 81.84 & 84.03 & 6.87 & 97.49 & 79.48 \\
Variant 3 & last & $f_{\mathrm{a}}$ & $\checkmark$ &  & $\checkmark$ & $2db\!+\!dd$ & 1.573M & 79.75 & 81.93 & 6.05 & 97.96 & \uline{81.33} \\
Variant 4 & last & $f_{\mathrm{b}}$ &  &  & $\checkmark$ & $2db$ & 0.524M & 77.36 & 80.68 & 6.57 & 97.94 & 79.01 \\ \hdashline[1pt/1pt]
FS-Adapter & last & $f_{\mathrm{b}}$ & $\checkmark$ &  & $\checkmark$ & $2db\!+\!bb$ & 0.590M & \uline{83.37} & \textbf{85.60} & \textbf{5.96} & \uline{98.04} & \textbf{85.31} \\ \noalign{\hrule height 1pt}
\end{tabular}
}
\endgroup
\end{table}
}

%-------------------------------------------------------------------------
\subsection{Qualitative Analysis}
\label{subsec:exp_vis}

% {
% \setlength{\abovecaptionskip}{0pt}
% \begin{figure}[htb!]
% \centering
% \includegraphics[width=1\linewidth]{fig/resconstuction.pdf}
% \caption{\textbf{Reconstruction Visualization} of real face images with a masking ratio of 75\%, using MIM models pretrained from: (a) a naïve MAE with simple random masking~\cite{he2022masked}, (b) a naïve MAE with our CRFR-P masking, and (c) our FSFM. All models were pretrained on the train and validation sets of FF++\_o~\cite{rossler2019faceforensics++} without adversarial learning, for 400 epochs. Images are from the test set.}
% \label{fig:rec}
% \end{figure}
% }

% \noindent\textbf{Reconstruction} To demonstrate the superiority of the facial representations pretrained with FSFM, we further follow MAE~\cite{he2022masked} to visualize reconstruction results, as shown in \cref{fig:rec}. We can see that FSFM demonstrates better reconstruction quality concerning intra-region consistency (preserving fine-grained textures within facial regions), inter-region coherency (maintaining spatial relationships across regions), and local-to-global correspondence (aligning local appearance with global facial looking).

% Visualization analysis by Grad-CAM.
To illustrate the superiority of our facial representations for discerning forgeries and spoofs, we visualize the GradCAM~\cite{selvaraju2017grad} maps of FS-VFM against the ImageNet supervised and MAE baselines in~\Cref{fig:vis_cam}: (a) \texttt{DFD} FS-VFM more accurately reveals forgery-relevant artifacts on FF++ corresponding manipulations, \eg, the altered mouth region in F2F and NT, whereas baselines confuse. (b) \texttt{FAS} FS-VFM highlights spoof-specific clues under the cross-domain MCIO evaluation, capturing inconsistent reflections across facial regions (M-Replay, I-Print), moiré patterns from screens (I-Replay, O-Replay, O-Photo), high-frequency presented textures (M-Paper, C-Replay), and cut edges of photos (C-Photo). These visualizations demonstrate that FS-VFM effectively responds to anomalies violating the suggested \textit{\textbf{3C}} of real faces, shedding light on the boosted generalization across face security tasks.

\section{Conclusion}
In this work, we present a scalable self-supervised pre-training framework, FS-VFM, that introduces the first universal \textbf{V}ision \textbf{F}oundation \textbf{M}odel for \textbf{F}ace \textbf{S}ecurity tasks. To learn fundamental and generalizable representations of real faces, we propose and pursue 3C pre-training objectives — intra-region Consistency, inter-region Coherency, and local-to-global Correspondence — by synergizing masked image modeling with instance discrimination. We show that FS-VFM consistently outperforms prior vision foundation models on cross-dataset deepfake detection, cross-domain face anti-spoofing, and unseen diffusion-based face forensic, and even outperforms SOTA task-specific methods via simple fine-tuning of a vanilla ViT. We further introduce FS-Adapter, a lightweight plug-and-play bottleneck module that facilitates efficient adaptation to downstream tasks while preserving superior generalization. Collectively, our contributions set a full-stack and robust groundwork for generalizable face security, and we hope this work spurs further research toward safeguarding facial authenticity against evolving threats.

{   
    \small
    \bibliographystyle{IEEEtran}
    \bibliography{main}
}

% \clearpage 

\vspace{-40pt}

\begin{IEEEbiography}[{\includegraphics[height=1.25in,clip,keepaspectratio]{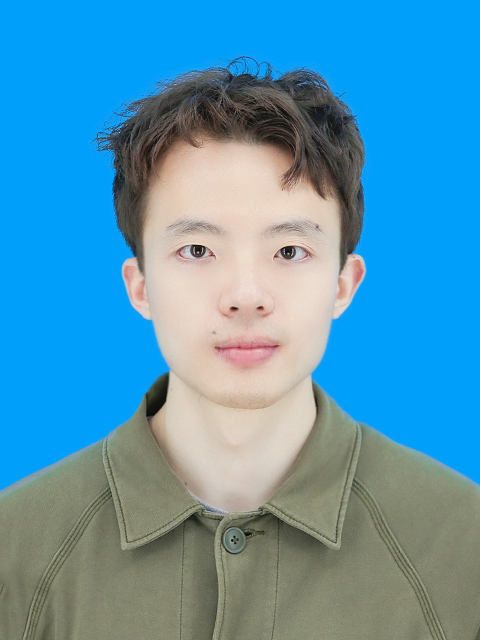}}]{Gaojian Wang}
is a Ph.D. candidate at the School of Cyber Science and Technology, Zhejiang University. His current research interests include AI security, representation learning, deepfake detection, face anti-spoofing, and trustworthy foundation models.
\end{IEEEbiography}

\vspace{-20pt}

\begin{IEEEbiography}[{\includegraphics[height=1.25in,clip,keepaspectratio]{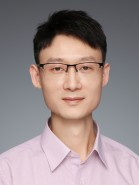}}]{Feng Lin}
(Senior Member, IEEE) received the Ph.D. degree from the Department of Electrical and Computer Engineering, Tennessee Technological University, USA, in 2015. He is currently a Professor with the School of Cyber Science and Technology, College of Computer Science and Technology, Zhejiang University, China. He was an Assistant Professor with the University of Colorado Denver, USA, a Research Scientist with the State University of New York (SUNY) at Buffalo, USA, and an Engineer with Alcatel-Lucent (currently, Nokia). His current research interests include IoT and Smart Vehicle Security, AI Security, Autonomous Driving, and Mobile Sensing. Dr. Lin was a recipient of the ACM SIGSAC China Rising Star Award, the Best Paper Awards from IEEE/ACM CHASE'22, ACM MobiSys'20, IEEE Globecom'19, IEEE BHI'17, the Best Demo Award from ACM HotMobile'18, and the Best Paper Award Nomination from SenSys'21 and INFOCOM'21. He serves as an associate editor for IEEE TIFS and IEEE Network.
\end{IEEEbiography}

\vspace{-20pt}

\begin{IEEEbiography}[{\includegraphics[height=1.25in,clip,keepaspectratio]{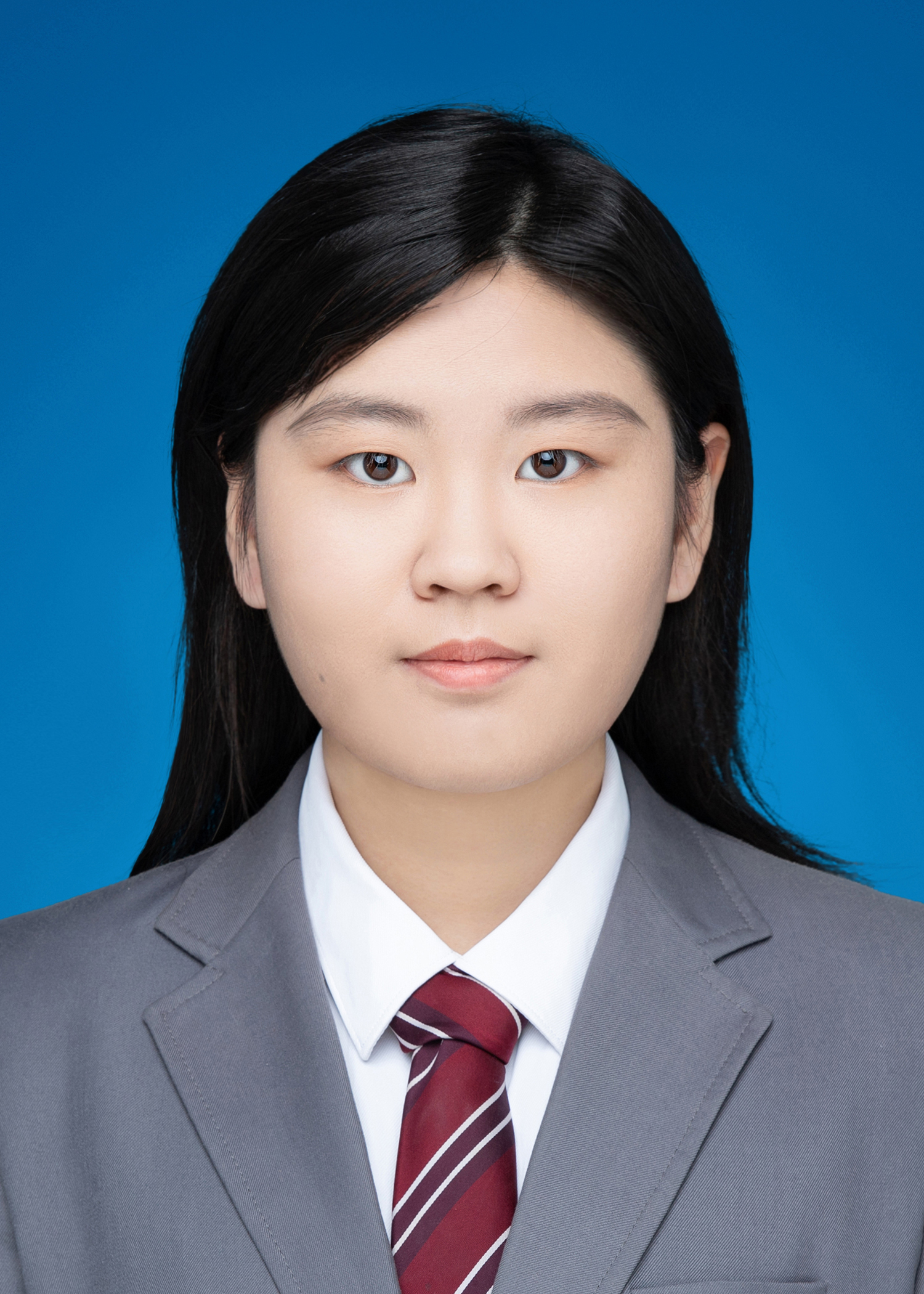}}]{Tong Wu}
received the B.S. degree in cyber science and engineering from Southeast University, in 2023. She is currently working toward the M.S. degree with the School of Cyber Science and Technology, Zhejiang University. Her research interests include AI security and IoT security.
\end{IEEEbiography}

\vspace{-20pt}

\begin{IEEEbiography}[{\includegraphics[height=1.25in,clip,keepaspectratio]{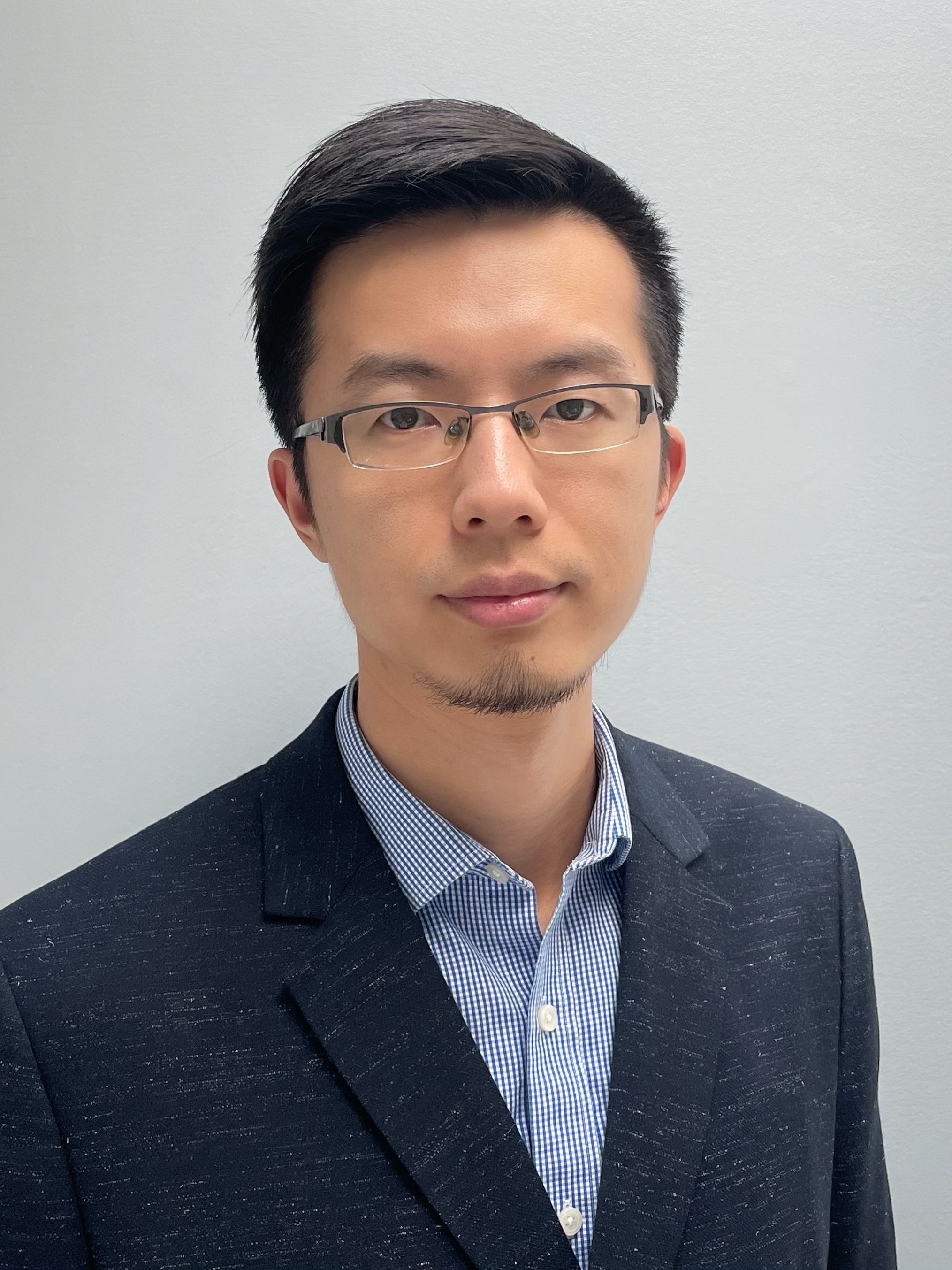}}]{Zhisheng Yan}
(Member, IEEE) received the PhD degree in computer science and engineering from University at Buffalo, The State University of New York. He is currently an associate professor in the Department of Information Sciences and Technology, School of Computing, George Mason University. He leads the Mason immErsive meDia computIng and Applications (MEDIA) Lab. Previously, he was an assistant professor in the Department of Computer Science, Georgia State University and a visiting researcher in the Department of Electrical Engineering, Stanford University. His research focuses on systems and security issues of immersive computing systems, such as VR, AR, imaging, and video systems. His research has been recognized by several awards, including NSF CAREER Award, NSF CRII Award, Mason Presidential Award for Faculty Excellence in Research, NDSS'24 Distinguished Paper Award, ACM MMSys'22 Best Student Paper Award, ACM SIGMM Best PhD Thesis Award, University at Buffalo CSE Best Dissertation Award, ACM HotMobile'18 Best Demo Award, and IEEE HealthCom'14 Best Student Paper Runner-up.
\end{IEEEbiography}

\vspace{-20pt}

\begin{IEEEbiography}[{\includegraphics[height=1.25in,clip,keepaspectratio]{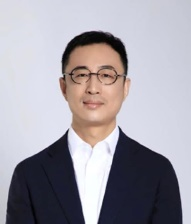}}]{Kui Ren}
(Fellow, IEEE) is the dean of College of Computer Science and Technology at Zhejiang University, where he also directs the Institute of Cyber Science and Technology. Before that, he was with State University of New York at Buffalo. He received his PhD degree in Electrical and Computer Engineering from Worcester Polytechnic Institute. Kui’s current research interests include Data Security, IoT Security, AI Security, and Privacy. He received Guohua Distinguished Scholar Award from ZJU in 2020, IEEE CISTC Technical Recognition Award in 2017, SUNY Chancellor’s Research Excellence Award in 2017, Sigma Xi Research Excellence Award in 2012 and NSF CAREER Award in 2011. Kui has published extensively in peer-reviewed journals and conferences and received the Test-of-Time Paper Award from IEEE INFOCOM and many Best Paper Awards from IEEE and ACM including MobiSys’20, ICDCS’20, Globecom’19, ASIACCS’18, ICDCS’17, etc. His h-index is 101, and his total publication citation exceeds 55,000 according to Google Scholar. Kui is a Clarivate Highly-Cited Researcher. He is a frequent reviewer for funding agencies internationally and serves on the editorial boards of many IEEE and ACM journals. He currently serves as Chair of SIGSAC of ACM China. He is a fellow of IEEE, AAAS, ACM, and CCF.
\end{IEEEbiography}

% % \newpage

 

% \begin{IEEEbiography}[{\includegraphics[height=1.25in,clip,keepaspectratio]{fig1}}]{Michael Shell}
% \end{IEEEbiography}


% \bf{If you will not include a photo:}\
% \[vspace\]
% {-33pt}

\vfill

\end{document}